\documentclass[twocolumn]{svjour3}
\usepackage{setspace}
\usepackage[table]{xcolor}

\usepackage{times}

\usepackage{epsfig}
\usepackage{graphicx}
\usepackage{amsmath}
\usepackage{amssymb}
\usepackage{fmtcount}
\usepackage{url}
\usepackage{amsmath}
\usepackage{amssymb}
\usepackage{epsfig}
\usepackage{url}
\usepackage{verbatim}
\usepackage{amsmath,amsfonts}
\usepackage[mathscr]{eucal}
\usepackage{amssymb,amsmath}
\usepackage{latexsym}
\usepackage{amsfonts}
\usepackage{amscd}
\usepackage{graphics}
\usepackage{algorithmic}
\usepackage{algorithm}

\def\bSigma{\boldsymbol\Sigma}

\newcommand {\R}{\mathbb {R}}

\newcommand {\N}{\mathcal {N}}

\DeclareMathOperator{\argmin}{argmin}

\newcommand{\dist}{{\rm dist}}

\renewcommand{\algorithmicrequire}{\textbf{Input:}}
\renewcommand{\algorithmicensure}{\textbf{Output:}}

\newcommand{\squishlist}{
 \begin{list}{$\bullet$}
  { \setlength{\itemsep}{0pt}
     \setlength{\parsep}{3pt}
     \setlength{\topsep}{3pt}
     \setlength{\partopsep}{0pt}
     \setlength{\leftmargin}{1.5em}
     \setlength{\labelwidth}{1em}
     \setlength{\labelsep}{0.5em} } }

\newcommand{\squishlisttwo}{
 \begin{list}{$\bullet$}
  { \setlength{\itemsep}{0pt}
     \setlength{\parsep}{0pt}
    \setlength{\topsep}{0pt}
    \setlength{\partopsep}{0pt}
    \setlength{\leftmargin}{2em}
    \setlength{\labelwidth}{1.5em}
    \setlength{\labelsep}{0.5em} } }

\newcommand{\squishend}{
  \end{list}  }

\newcounter{Lcount}
\newcommand{\squishlistnum}{
\begin{list}{\arabic{Lcount}. }
{ \usecounter{Lcount}
\setlength{\itemsep}{0pt}
\setlength{\parsep}{0pt}
\setlength{\topsep}{0pt}
\setlength{\partopsep}{0pt}
\setlength{\leftmargin}{2em}
\setlength{\labelwidth}{1.5em}
\setlength{\labelsep}{0.5em} } }

\newcommand{\squishendnum}{
\end{list} }

\DeclareMathOperator{\vol}{vol} 
\DeclareMathOperator{\supp}{supp} 
\DeclareMathOperator{\Sp}{span} 

\newcommand{\reals}{\mathbb R}
\newcommand{\be}{\begin{equation}}
\newcommand{\ee}{\end{equation}}

\newcommand{\di}{{\,\mathrm{d}}}

\def\bSigma{\boldsymbol\Sigma}

\def\by{\mathbf{y}}
\def\bz{\mathbf{z}}

\def\bS{\mathbf{S}}
\def\bD{\mathbf{D}}

\def\bU{\mathbf{U}}
\def\bu{\mathbf{u}}

\def\bA{\mathbf{A}}

\def\sX{\mathrm{X}}

\def \di{\mathrm{d}}

\def\bx{\mathbf{x}}

\def\eps{\varepsilon}

\begin{document}
\title{Hybrid Linear Modeling via Local Best-fit Flats
\thanks{This work was supported by NSF grants
DMS-06-12608, DMS-08-11203, DMS-09-15064 and DMS-09-56072.
Thanks to the action editor and the reviewers for the careful reading and comments; Peter Jones,
Mauro Maggioni and Amit Singer for discussions that
motivated our exploration for a multiscale SVD-based HLM algorithm; Ehsan Elhamifar and Ren\'{e} Vidal
for answering various questions regarding the SSC code and providing us an initial version before the code
was available to the public; Allen Yang for clarifying the estimation of the number of clusters in GPCA;
and the IMA for a stimulating multi-manifold modeling workshop.
}
\thanks{Corresponding Author: Gilad Lerman,
phone: (612) 624-5541,
fax: (612) 626-2017
}
}

\author{
Teng Zhang
\and
Arthur Szlam
\and
Yi Wang
\and Gilad
Lerman }

\institute{Teng Zhang \and Yi Wang \and Gilad Lerman \at
              School of Mathematics\\
              University of Minnsota\\
              \email{\{zhang620, wangx857, lerman\}@umn.edu}           %  \\
           \and
           Arthur Szlam \at
            Courant Institute of Mathematical Sciences\\
            New York University\\
              \email{aszlam@courant.nyu.edu}
}

\maketitle

\begin{abstract}
We present a simple and fast geometric method for modeling data by a union of affine subspaces.  The method begins by forming a collection of local best-fit affine subspaces, i.e., subspaces approximating the data in local neighborhoods. The correct sizes of the local neighborhoods are determined automatically by the Jones' $\beta_2$ numbers (we prove under certain geometric conditions that our method finds the optimal local neighborhoods).
The collection of subspaces is further processed by a greedy selection procedure or a spectral method to generate the final model.
We discuss applications to tracking-based motion segmentation and
clustering of faces under different illuminating conditions.
We give extensive experimental evidence demonstrating the state of
the art accuracy and speed of the suggested algorithms on these problems and also on synthetic
hybrid linear data as well as the MNIST handwritten digits data; and we demonstrate how to use our
algorithms for fast determination of the number of affine subspaces.
\end{abstract}

\noindent \textbf{Supp. webpage}:

\url{http://www.math.umn.edu/~lerman/lbf/}

\section{Introduction}
Several problems from computer vision, such as
motion segmentation and face clustering, give rise to modeling data by multiple subspaces.
This is referred to as Hybrid Linear Modeling (HLM) or alternatively as ``subspace clustering''.
In tracking-based motion segmentation, extracted feature points (tracked in all frames) are clustered according to the different moving objects. Under the affine camera model, the vectors of coordinates of feature points corresponding to a moving rigid object lie on an affine subspace of dimension at most 3 (see~\cite{Costeira98}). Thus clustering different moving objects is equivalent to clustering different affine subspaces.
Similarly, in face clustering, it has been proved that the set of all images of
a Lambertian object under a variety of lighting conditions
form a convex polyhedral cone in the image space, and this
cone can be accurately approximated by a low-dimensional
linear subspace (of dimension at most 9)~\cite{bb34497,Ho03,Basri03}. One may thus cluster certain
images of faces by HLM algorithms.

The mathematical formulation of HLM assumes a data set
$\mathrm{X}=\{\bx_i\}_{i=1}^N \subseteq \mathbb{R}^{D}$ where each $\bx_i$ lies on (or around) one of $K$ flats (i.e., affine subspaces) and requires to find the partition of $\mathrm{X}$ corresponding to the flats.  We would like to be able to do this when the data has been corrupted by additive noise and outliers\footnote{Throughout the paper outliers are corrupted data points, i.e., points generated by a distribution, which assigns sufficiently small probability for small neighborhoods around the underlying subspaces. This is different than corrupting selected entries of data points.}; in this case we may also want to determine the flats themselves.
We first assume here that all flats have the same known dimension $d$ (i.e., they are $d$-flats) and that their number $K$ is known. In Section~\ref{sec:exp} we address to some extent the cases of unknown $K$ and mixed dimensions.
%It is important to guarantee that an HLM algorithm can solve this mathematical problem and only few such guarantees exist, which are partially satisfying~\cite{spectral_theory,Arias-Castro08Surfaces,lp_recovery_part2_11}). However, it is also very important to test HLM algorithms on real data sets, which may deviate from such a mathematical model.

Several algorithms have been suggested for solving the
HLM problem (or even the more general problem of clustering manifolds),
for example the $K$-flats (KF) algorithm or any of its
variants~\cite{Tipping99mixtures,Bradley00kplanes,Tseng00nearest,Ho03,MKF_workshop09},
methods based on direct matrix factorization~\cite{Boult91factorization-basedsegmentation,Costeira98,Kanatani01,Kanatani02},
Generalized Principal Component Analysis (GPCA)~\cite{Vidal05},
Local Subspace Affinity (LSA)~\cite{Yan06LSA}, RANSAC (for HLM)~\cite{Yang06Robust},  Locally Linear Manifold Clustering (LLMC)~\cite{LLMC},
Agglomerative Lossy
Compression (ALC)~\cite{Ma07Compression}, Spectral Curvature Clustering (SCC) \cite{spectral_applied} and Sparse Subspace Clustering (SSC)~\cite{ssc09}.
Some theoretical guarantees for particular HLM algorithms appear
in~\cite{spectral_theory,Arias-Castro11Surfaces,lp_recovery_part2_11,Soltanolkotabi_Candes2011}.
We recommend a recent review on HLM by Vidal~\cite{SubspaceClustering_Vidal}.

%Despite substantial progress and the many recent algorithms there is much yet to do.
Many of the algorithms described above require an initial guess of the subspaces.   For example, the $K$-flats algorithm is an iterative method that requires an initialization, and in SCC, one needs to carefully choose collections of $d+1$ data points that lie close to each of the underlying $d$-flats.
Other algorithms require some information about the suspected deviations from the hybrid linear model;
for example  both RANSAC (for HLM) and ALC ask for a model parameter corresponding to the level of noise.

%We resolve here these particular problems as well as come up with alternative solutions for HLM by using a
Here we propose a
straightforward geometric method for the estimation of local subspaces, which is inspired by~\cite{Jones90,DS91,Lerman03}
and~\cite{Fukunaga71,MSVD09_1,MSVD09_2}.  These local subspace estimates can be used to  set the model parameters for or initialize an HLM algorithm.
The basic idea is that for a data set $X$
sampled from a hybrid linear model (perhaps with some noise),  there are many points $\bx$
such that the principal components of an appropriately sized neighborhood of $\bx$
give a good approximation to the subspace $\bx$ belongs to.  Using local subspaces to infer the global hybrid linear model was suggested in~\cite{Yan06LSA} for linear subspaces; however, there they use very small neighborhoods that are not adaptive to the structure of the data (e.g., amount of noise etc.).
An ``appropriately sized neighborhood'' needs to be larger than the noise, so
that the subspace is recognized. However, the neighborhood cannot be so large that
it contains points from multiple subspaces. The correct choice of this size is carefully quantified in Section~\ref{sec:beta}.

In addition to studying how to estimate local subspaces, we describe two complete HLM algorithms which are natural extensions of the local estimation: LBF (Local Best-fit Flat) and SLBF (spectral LBF).  On many data sets, the first obtains state of the art speed with nearly state of the art accuracy (it can also deal with very large data), and the second obtains state of the art accuracy (SLBF) with reasonable run times (it seems to be able to deal to some extent with some nonlinear structures as the ones arising in motion segmentation data). We remark that we test accuracy in various scenarios, but in particular, with intersecting subspaces and with outliers.  While in this work we only theoretically justify our choice of initializer, we are hopeful about developing a more complete theory justifying our algorithms.

In particular, we believe that such a theory can be valid in the setting suggested by Soltanolkotabi and Cand{\`e}s \cite{Soltanolkotabi_Candes2011} for analyzing the SSC algorithm, while having additional noise and restricting the fraction of outliers (or modifying our algorithms so they are even more robust to outliers). We are also interested in rigorously quantifying the limitations of our algorithms (as conjectured in Section~\ref{sec:conclusions}).

%We also show how to use the underlying idea of both algorithms, i.e., learning good local subspaces determining the structure of the data (or part of it), to effectively initialize the $K$-flats algorithms as well as speed up or modify other algorithms. While we only justify here our choice of local subspaces, we are hopeful that a more complete theory justifying some of our algorithms (e.g., SLBF and $K$-flats initialized with our strategy) can be developed. We also indicate a fundamental problem of local methods for certain type of data, in particular cropped face images.

%The LSA algorithm has already suggested using local subspaces to infer the global hybrid linear model. However, it is restricted to linear subspaces and the subspaces are studied in very small neighborhood that are not adaptive to the structure of the data (e.g., amount of noise etc.).  %%%%%%%%%this does not belong here- but it is good.  put it in the next section?

%It can be used in a stand alone manner for LBF and SLBF or as an initialization of many of the above methods.

We summarize the main contributions of this work, which is the full length version of ~\cite{LBF_cvpr10}, as follows.
\squishlist
\item We make precise the
local best-fit heuristic, using the $\beta_2$
numbers~\cite{Jones90,DS91,Lerman03}.  We give an algorithm to approximately find optimal neighborhoods in the above sense, in fact, we prove this under certain geometric conditions.
\item Using the local best-fit heuristic, we introduce the LBF and SLBF algorithms
for HLM.  At
each point of a randomly chosen subset of the data, they use the best-fit flats of the
``optimal'' neighborhoods to
build a global model with different methods
(LBF is based on energy minimization and SLBF is a spectral method).
\item
We perform extensive experiments on motion segmentation data (the Hopkins 155 benchmark of \cite{Tron07abenchmark}), face clustering (the extended Yale face database B), handwritten digits (the MNIST database), and artificial data, showing
 that both algorithms, in particular SLBF, are accurate
 on real and synthetic HLM problems, while LBF runs extremely fast
(often on the order of ten times faster than most of the previously
mentioned methods). For the cropped face data we actually indicate a fundamental problem of local methods like LBF and SLBF, though suggest a workaround that works for this particular data.

\item We demonstrate how the local best-fit heuristic can be used with other algorithms.
In particular, we give experimental evidence to show that the
$K$-flats algorithm~\cite{Ho03} is improved by initialization that is based
on the local best-fit heuristic. We also use this heuristic to estimate the main parameters of both
RANSAC (for HLM)~\cite{Yang06Robust} and ALC~\cite{Ma07Compression}.
\item We show how the combination of LBF and the elbow method can quickly determine the
number of subspaces.
\squishend

The rest of this paper is organized as follows. In
Section~\ref{sec:algorithm} we describe the
LBF and SLBF algorithms and state a theorem giving conditions
that guarantee that good neighborhoods can be found.
 Section~\ref{sec:exp} carefully tests the LBF and SLBF
algorithms (while comparing them to other common HLM algorithms) on both artificial data of synthetic hybrid linear models
and real data of motion segmentation in video sequences, face clustering and handwritten digits recognition.
It also demonstrates how to determine the number of clusters by applying the
fast algorithm of this paper together with the straightforward elbow
method. Section~\ref{sec:conclusions} concludes with a brief
discussion and mentions possibilities for future work.

\section{The local best-fit flats heuristic and the LBF and SLBF algorithms}\label{sec:algorithm}

We describe two methods, LBF and SLBF, which have at their heart an estimation of local flats capturing the global structures of the data (or part of it).
Both methods first find a set of candidate flats (the number is an input parameter for LBF). These are best-fit flats for local ``optimal'' neighborhoods (we describe an algorithm for approximately finding such neighborhoods and justify it in Section~\ref{sec:beta}).
The two algorithms process the candidates in different ways: LBF uses energy minimization and SLBF uses a spectral approach.

\subsection{Choosing the optimal neighborhood}
\label{sec:beta}

We choose the candidate flats that capture the global structure of the data by fitting them to `optimal' local neighborhoods of data points.
%Choosing the optimal neighborhood is crucial for the success of both proposed methods, and is in some sense the central problem of this paper.
For a point $\bx\in\reals^D$, we define an optimal neighborhood as the largest ball $B(\bx,r)$ (centered at $\bx$ and with radius $r$) that only contains points sampled from the same cluster as $\bx$. Indeed, neighborhoods smaller than the optimal one (around $\bx$) can mainly contain the noise around an underlying subspace (of the hybrid linear model); consequently their local best-fit flats may not match the underlying flat. On the other hand larger neighborhood than the optimal one (around $\bx$) will contain
points from more than one underlying flat, and the resulting best-fit
flat will again not match any of the underlying flats. We note that the choice of neighborhood $B(\bx,r)$ is equivalent to the choice of radius $r$, which we refer to as scale (even though it is also common to refer to $\log(r)$ or -$\log(r)$ as scale). While it
is possible to take a guess at the optimal scale as a parameter (e.g., as commonly done by fixing the number of nearest neighbors to $\bx$), we
have found that it is possible to choose the optimal scale
reasonably well automatically, while adapting it to the given point $\bx$.

We will
start at the smallest scale (i.e., smallest radius containing only $d+1$ points) and look
at larger and larger neighborhoods of a given point $\bx_0$.
At the
smallest scale, any noise may cause the local neighborhood to have higher dimension than $d$.
As we add points to the neighborhood, it becomes
better and better approximated in a scale-invariant sense (e.g., by scaling the neighborhood to have radius 1 and computing the
error of approximation by best-fit flat then) until points belonging to other flats enter the neighborhood.
To be more precise, we define the scale-invariant error for a neighborhood $\N \equiv B(\bx_0,r)$ of $\bx_0$ by the formula:
\be \label{eq:def_beta}
%\[
\beta_2(\N)=\frac{\min_{\text{$d$-flats $L$}}\sqrt{\sum_{\by\in \N}
||\by-P_{L}\by||^2/|\N|}}{\max_{\bx\in \N}||\bx-\bx_0||}, \ee
%\]
where $|\N|$ denotes the number of points in $\N$, $P_{L}$ denotes the projection onto the flat $L$ and the minimization is over all $d$-flats in $\reals^D$.
We note that the numerator is the approximation error by best-fit $\ell_2$ flat at scale $r$ and the denominator is the scale $r$.
The notion of
scale-invariant error was introduced and utilized
in~\cite{Jones90,DS91,Lerman03}.

Using this scale-invariant error we can reformulate our criterion for choosing the optimal neighborhood more precisely.
That is, we start with the smallest neighborhood containing $S$ nearest neighbors of $\bx$, increase the number of nearest neighbors by $T$ in each iteration and check the $\beta_2$ number of each neighborhood using~\eqref{eq:def_beta}.
We estimate the optimal neighborhood as the last one for which $\beta_2$ is smaller than $\beta_2$ of the previous neighborhood (that is, we search for the first local minimizer of $\beta_2(\N)$).
This procedure is summarized in Algorithm~\ref{alg:nhbd}. It is experimentally robust to outliers if $\bx$ is an inlier, since
the nearest neighbors of inliers also tend to be inliers.

\begin{algorithm}[htbp]
\caption{Neighborhood size selection for HLM by randomized local
best-fit flats} \label{alg:nhbd}
\begin{algorithmic}
\REQUIRE $\sX=\{\bx_1,\bx_2,\cdots,\bx_N\} \subseteq
\mathbb{R}^{D}$: data, $\bx$: a point in $\sX$, $S$: start size,
$T$: step size%, $\ell, m$ (optional): mean shifts parameters.
\ENSURE $\N(\bx)$: a neighborhood of $\bx$.\\
\textbf{Steps}:
\STATE
%
%\begin{enumerate}
%     $\bullet$ (Optional) Update the point $\bx$ as the center of its $\ell$-nearest neighborhood in $\sX$, while
%     repeating $m$      times\\
     $\bullet$ $k=-1$
    \REPEAT \STATE
     $\bullet$ k:=k+1\\
     $\bullet$ Let $\N_k$ be the set of the $S+kT$ nearest points in $\sX$ to
     $\bx$\\
     %$\bullet$ Set $\tilde{L}_k$ to be the best-fit flat to $\N_k$\\
     $\bullet$ Compute $\beta_2(k):=\beta_2(\N_k)$ according to~\eqref{eq:def_beta}
     \UNTIL $k>1$ and
    $\beta_2(k-1)<\min\{\beta_2(k-2),\beta_2(k)\}$\\
    $\bullet$ Output $\N(\bx) :=\N_{k-1}$
%\end{enumerate}
%
\end{algorithmic}
\end{algorithm}

\subsubsection{Theoretical justification}
The following theorem tries to justify
our strategy of estimating the optimal scale around each point by showing that in the continuous setting the first local minimizer of $\beta_2(\bx,r) := \beta_2(B(\bx,r))$ is approximately the distance from $\bx$ to the nearest cluster that does not contain $\bx$ (here the underlying model is a mixture of Lebesgue measures in strips around several subspaces and $\bx$ is an arbitrary point on one of these subspaces).
Therefore, if we choose the size of neighborhood following Algorithm~\ref{alg:nhbd} (adapted to the continuous setting), then we will approximately obtain the optimal neighborhood. It is rather standard to extend such estimates for measures to a probabilistic setting, where i.i.d.~data is sampled from the continuous distribution. The theorem will then hold with high probability for sufficiently large sample size (due to technicalities, which also require truncating the support of our continuous measure we avoid these details).
The proof of this theorem is in the Appendix.

Our theorem uses the following analog of the discrete $\beta_2$ of~\eqref{eq:def_beta} for
a measure $\mu$ and a ball $B(\bx,r)$ (see
also~\cite{Lerman03}):
\begin{align} &\nonumber \beta_2(\bx,r) = \\
&\frac{1}{r} \min_{d-\text{flats } L \subseteq \reals^D } \sqrt{\int_{B(\bx,r)} \dist(\bx,L)^2 \, \di \mu / \mu(B(\bx,r))}, \label{eq:beta_2} \end{align}
where throughout the paper $\dist(\cdot, \cdot)$ denotes the Euclidean distance, for example in this case $\dist(\bx,L):=\|\bx-P_{L}\bx\|$.
The theorem also assumes that the underlying measure is supported on a union of $K$ tubes $T(L_i,w_i):=\{\bx\in\reals^D: \|\bx-P_{L_i}\bx\|<w_i\}$, $i=1, \ldots, K$
(centered around the $d$-flats $L_1, \ldots, L_K$ respectively). %We remark that probabilistic sampling from such a measure will produce HLM data.
%For convenience we assume tubes but
%restrict to local scales.

\begin{theorem}\label{thm:main}
%For integers $D>d>0$, and any bounded set $A\subset \mathbb{R}^D$, define a function $\phi(A)$ as:
%\begin{equation}\label{eq:defi}
%\phi(A)=\min_{d-\text{flats L}}\frac{\int_A d(\bx,L)^2 \di\mu(x)}{\mu(A)\cdot diameter(A)^2}
%\end{equation}
Let $K \geq 2$, $d<D$ , $\{L_i\}_{i=1}^K$ be $K$ $d$-flats in
$\reals^D$, $\{\mu_i\}_{i=1}^K$  be $K$ Lebesgue measures on tubes $\{T(L_i,w)\}_{i=1}^{K}\\\subset\reals^D$ respectively and let
$\mu=\sum_{i=1}^{K}\mu_i$.
For fixed $1 \leq i^* \leq K$ and fixed $\bx^*\in L_{i^*}$, let
\be r_0 = r_0(\bx^*) = \dist\left(\bx^*,\cup_{i\neq i^*}T(L_i,w)\right).\ee
%
%Assume that $r_0> w$.

If $w<r_0$, then
$\beta_2(\bx^*, r)$ (as a function of $r$) is constant on $[0,w]$ and
decreases on $[w, r_0]$. If also
\begin{align}\label{eq:condition0}
\frac{w}{r_0}<\begin{cases}\min\left(0.02,\sqrt{\frac{(D+1)}{150\sqrt{2}\,(D-1)K}}\right),&\text{when $d=1$;}\\
\min\left(0.02,\sqrt{\frac{(D-d+2)}{6(50)^\frac{d}{2}(D-d)K}}\right),&\text{when $d>1$,}
\end{cases}
\end{align}
then there exists $r_0 < r^* < 1.09\,r_0$ such that
\begin{equation}
\label{eq:beta_local_min} \beta_2(\bx^*,r^*) >
\beta_2(\bx^*, r_0).
\end{equation}
That is, the first local minimum of $\beta_2(\bx^*, r)$ (as a function of $r$) occurs in $(r_0, 1.09 \, r_0)$.

\end{theorem}
The proof of this theorem indicates a weaker condition than~\eqref{eq:condition0}, which is less intuitive. It also shows that $r^*\rightarrow r_0$ as $w/r_0\rightarrow 0$
and clarifies
by example why the first local minimum of $\beta_2(\bx^*,r_0)$ is often bigger than $r_0$ (see Remark 1).

%still needs work

\subsubsection{The complexity of Algorithm~\ref{alg:nhbd}} Algorithm~\ref{alg:nhbd}
requires sorting the neighbors of $\bx$ according to their distance to $\bx$; the computational cost of this preprocessing step is $O(D\cdot N+N\cdot \log N)$. In order to obtain $\beta_2(\N_k)$, we need to obtain the top $d$ singular values of the $|\N_k|\times D$ data matrix representing the $|\N_k|$ points, which requires a complexity of $O(d\cdot D\cdot |\N_k|)$. To find $\N(x)$, we need to generate $\beta_2(\N_k)$ for any $|\N_k|=S+kT$, where $k=1,2,\cdots,(N-S)/T$, hence the complexity for obtaining $\N(x)$ is of order:
$$O(d\cdot D\cdot \sum_{k=1}^{(N-S)/T}(S+kT))\leq O(d\cdot D\cdot N^2/2T).$$
We thus note that if $T$ is in the order of $N$, e.g., $T=\max\\(N/300,2)$,  the total complexity of  Algorithm~\ref{alg:nhbd} is $O((d\cdot D+\log N)\cdot N)$.  Note that if we limit the number of scales that we search, then the two $\log$ terms above can be replaced by a constant.

\subsection{The LBF algorithm}
\label{sec:lbf_candidate}

The LBF algorithm searches for a good set of flats
from the candidates (described above) in a greedy fashion.
A measure of goodness of a $K$ tuple of flats
$G$ is chosen; here, it will be the average $l_1$
distance of each point to its nearest flat, i.e.,
\begin{equation}\label{eq:def_error_ksub}
G=G_{\sX}(\{L_1,\cdots,L_K\})
%= \sum_{\bx\in\sX}\min_{1\leq
%i \leq K}\left(\dist(\bx,\rmL_i)\right)^p \equiv
= \sum_{\bx\in \sX} \dist \left(\bx,\cup_{i=1}^{K} L_i\right).
\end{equation}
After randomly initializing $K$ flats from the list of candidates,
$p$ passes are made through the data points.  In each of the passes, we replace a random current flat with
the candidate that minimizers the value of $G$.  We then move to the
next pass, picking a random flat, etc. Algorithm~\ref{alg:candidate} sketches this procedure (where the greedy minimization of $G$
is described in step~\ref{item:greedy}).

\begin{algorithm}[htbp]
\caption{LBF: energy minimization over randomized local best-fit flats}
\label{alg:candidate}
\begin{algorithmic}
\REQUIRE $\sX=\{\bx_1,\bx_2,\cdots,\bx_N\} \subseteq
\mathbb{R}^{D}$: data, $d$: dimension of subspaces,  $C$: number of
candidate planes, $K$: number of output flats/clusters, $p$: number
of passes,  $S$ and $T$: parameters for local scale calculation.

\ENSURE  A partition of $\sX$ into $K$ disjoint clusters
$\{\sX_i\}_{i=1}^K$, each approximated
by a $d$-dimensional flat.\\
%\ENSURE $C$ candidate flats $L_1,...,L_C$.\\
\textbf{Steps}:
\STATE
\squishlistnum %\begin{enumerate}
    \item Pick $C$ random points in $\sX$
    \item \label{item:candidate} For each of the $C$ points find appropriate local scale using Algorithm~\ref{alg:nhbd}
     \item Generate a set $\mathcal{L}$ containing $C$ candidate flats $L_1,...,L_C$ from the best
    fit flats to the neighborhoods from the previous step \label{item:subspaces}
    %\item Choose $K$ flats from the candidates using Algorithm~\ref{alg:passes}; collect these in $\mathcal{L}$ \label{item:apply3}
    %\item Choose a set of $K$ flats from the candidates as follows:
    \item Pick a random subset of $K$ flats $\hat{\mathcal{L}} \subset \mathcal{L}$
    \FOR{$j=1$ to $p$}\label{item:greedy}
            %\STATE
            \item $\bullet$ Pick a random flat $L^*\in\hat{\mathcal{L}}$, and find
            $\hat{L}=\argmin_{L\in\mathcal{L}}G_{\sX}\left(\left\{\hat{\mathcal{L}}\setminus \{L^*\}\right\} \cup \{L\}\right)$
            \item $\bullet$ Update $\hat{\mathcal{L}}=\Big\{\hat{\mathcal{L}}\setminus \{L^*\}\Big\} \cup \{\hat{L}\}$
        \ENDFOR
    \item Partition $\sX$ by sending points to nearest flats
    in $\hat{\mathcal{L}}$ \label{item:partition}
\squishendnum %\end{enumerate}
\end{algorithmic}
\end{algorithm}

%\begin{algorithm}[htbp]
%\caption{Greedy $l_1$ candidate selection for HLM by randomized
%local best-fit flats} \label{alg:passes}
%\begin{algorithmic}
%%
%\REQUIRE $\sX=\{\bx_1,\bx_2,\cdots,\bx_N\} \subseteq
%\mathbb{R}^{D}$: data, $K$: number of flats, $L_1,...,L_C$:
%candidate flats, and $p$: number of passes.
%%
%\ENSURE A set of $K$ ``active'' flats $\mathcal{L}\subset \{L_1,...,L_C\}$ .\\
%%\ENSURE A partition of $\sX$ into $K$ disjoint clusters $\{\sX \}_{i=1}^K$.\\
%%\textbf{Steps}:
%%
%\textbf{Steps}:
%%
%\STATE %
%%\begin{enumerate}
%    %\item
%    Initialize $\mathcal{L}$ by randomly choosing $K$ ``active'' flats $L_{A_1},...,L_{A_K}$
%%    \begin{algorithmic}
%    \FOR{$\text{pass}=1$ to $p$}
%        \STATE
%        Pick a random flat $L_{A_l} \subset \mathcal{L}$ ($1 \leq l \leq K$)
%        \FOR{$j=1$ to $C-K$}
%            %\STATE
%            \item $\bullet$ Pick one of the ``inactive'' flats $L_j$ and form the
%            collection of flats $\tilde{\mathcal{L}}= L_j \bigcup  \mathcal{L}\setminus L_{A_l}$
%            %\STATE
%            \item $\bullet$ Set $s_j=\sum_{i=1}^N \min_{L\in
%            \tilde{\mathcal{L}}}{||x_i-P_Lx_i||}$
%        \ENDFOR
%
%        If $\min{s_j}<\sum_{i=1}^N \min_{L\in \{L_{A_1},...,L_{A_K}\}}{||x_i-P_Lx_i||}$,
%        set $L_{A_l}:= L_{\argmin{s_j}}$
%    \ENDFOR
% %   \end{algorithmic}
%%\end{enumerate}
%%
%\end{algorithmic}
%%
%\end{algorithm}
%

The simplest choice of $G$ is the sum of the squared distances of
each point in $\sX$ to its nearest flat, i.e., having the
power 2 in~\eqref{eq:def_error_ksub}.
However, in some scenarios the $l_1$ energy of~\eqref{eq:def_error_ksub} is more robust to
outliers than the mean squared error (see~\cite{lp_recovery_part1_11,lp_recovery_part2_11} for theoretical support and~\cite{MKF_workshop09} for experimental support).
%One can also imagine using spectral distances that measure the smoothness of the clusters with
%respect to some kernel, or many other global energy functionals of a
%partition.
This method also allows using energy functions, which are hard to minimize (even heuristically). Indeed, it only requires evaluating the energy on the candidate configurations. For example, when the data set requires stronger robustness to outliers, one may use the following energy:
\begin{equation*}
G'=G'_{\sX}(\{L_1,\cdots,L_K\})
%= \sum_{\bx\in\sX}\min_{1\leq
%i \leq K}\left(\dist(\bx,\rmL_i)\right)^p \equiv
= \mathrm{Median}_{\bx\in \sX} \dist \left(\bx, \cup_{i=1}^K L_i\right).
\end{equation*}

The LBF algorithm is closely related to RANSAC, since both of them use candidate subspaces to fit the data set. However Algorithm~\ref{alg:nhbd} gives LBF an advantage in choosing good candidates, while RANSAC fits a $d$-flat by arbitrarily chosen $d+1$ points.

\subsubsection{ The complexity and storage of Algorithm~\ref{alg:candidate}} For step~\ref{item:candidate} of this algorithm
we need to run Algorithm~\ref{alg:nhbd} $C$ times and thus its complexity is of order $O((d\cdot D+\log N)\cdot C\cdot N)$. Note that the  $\log N$ comes from a full sort of $N$ distances, and if we restrict to a fixed number of scales, this can be replaced by a constant.
Step~\ref{item:subspaces} of Algorithm~\ref{alg:candidate}, requires $C$ SVD decompositions for $C$ matrices of size at most $N\times D$, in order to obtain the first $d$ vectors in $\reals^D$. It thus also has a complexity at most $O(C\cdot d\cdot D \cdot N)$.

Step~\ref{item:greedy} of Algorithm~\ref{alg:candidate}
requires the evaluation of the $N\times C$ matrix representing the distances $||x_i-P_{L_j}x_i||$ between $\sX=\{x_1,x_2,\cdots,x_N\}$ and $L_1,L_1,\cdots,L_C$. This costs $O(C\cdot d\cdot D \cdot N)$ operations, since each distance from a point to a subspace costs $O(d\cdot D)$. Moreover, the $p$ passes have complexity of order $O(p\cdot (C-K)\cdot N)$. Therefore, step~\ref{item:greedy} of Algorithm~\ref{alg:candidate} has a complexity of order $O(C\cdot N \cdot (d \cdot D+p))$.
At last, Step~\ref{item:partition} of Algorithm~\ref{alg:candidate} has a complexity of order $O(K\cdot d\cdot D \cdot N)$, which comes from the construction of the $N\times K$ matrix of distances from $N$ points to $K$ subspaces.
Combining these complexities together, we have an overall complexity of $O(C\cdot N \cdot (d \cdot D+p+ \log N))$ for the LBF Algorithm; as before, if we fix the number of scales independently from $N$, the $\log$ terms can be replaced by a constant.

%For step 2 of this algorithm we need to run Algorithm 1 C
%times and thus its complexity is of order O((d  D + log N )

%In comparison, $K$-flats has a complexity of $n_s\cdot N \cdot K \cdot d \cdot D $, where $n_s$ is the number of iterations. If $n_s$ is in the order of $O(1)$ and we choose $C=70K$ and $p=5K$ as in Section~\ref{sec:exp}, LBF seems to have a slightly larger complexity (due to $p+\log N$). However, the $p+\log N$ factor competes with $D\cdot d$ and for most practical data it is not a big issue, and as shown in Sections~\ref{sec:sim} and~\ref{sec:face_seg}, both algorithms have comparable speed in experiments on artificial data and real data.

%Notice that we choose $C=70K$ and $p=5K$ in the implementation of LBF. Therefore, if $K$ is smaller than or in the same order of $d\cdot D$, then the complexities of LBF and $K$-flats are comparable.   %%%%%%%%%%%%%ADS: this might be controversial.  there is no reason the number of iterations should be comparable.

To compute the storage requirements of LBF, we note that the data set is saved in an $N \times D$ matrix, the candidate subspaces are organized in $C$ projection matrices  of size $D \times d$ and in addition the algorithm stores an $N\times C$ matrix of distances between the data points and the $C$ candidate subspaces.
Therefore, the storage of LBF is in the order of $O(D\cdot N+C\cdot D \cdot d+N\cdot C)$.

% stopped here...

\subsection{The SLBF algorithm}
\label{sec:slbf}

The SLBF algorithm (which is sketched in Algorithm~\ref{alg:lbfsc})
processes the candidate subspaces via a spectral clustering method. It first finds the neighborhoods $\{\mathcal{N}_i\}_{i=1}^{N}$ for all points $\{\bx_i\}_{i=1}^{N}$ via Algorithm \ref{alg:nhbd} and fits $d$-flats $\{L_i\}_{i=1}^{N}$ (via PCA) in these neighborhoods. It then forms the $N\times N$ matrices  $\bS$ and $\hat{\bS}$ as follows:  \begin{equation}\bS_{i,j}=\sqrt{\dist(\bx_i,L_j)\,\dist(\bx_j,L_i)},\label{eq:construct_S}\end{equation}
and
\begin{equation}\hat{\bS}_{i,j}=\exp(-\bS_{i,j}/2\sigma_j^2)+\exp(-\bS_{i,j}/2\sigma_i^2),\label{eq:construct_hatS}\end{equation}
where
\begin{equation}\sigma_j=\lambda\sqrt{\min_{\text{$d$-flats\,\,L}}\sum_{\bx\in\mathcal{N}_j}\|\bx-P_{L}\bx\|^2/|\N_j|}\label{eq:construct_sigma}\end{equation}
(we explain the choice of $\lambda$ below, when we clarify~\eqref{eq:construct_sigma}).
Finally, it applies spectral clustering with the matrix $\hat{\bS}$.
More precisely, SLBF follows the main algorithm of~\cite[Section 2]{Ng02-bis},  replacing the matrix $A$ there by $\hat{\bS}$, multiplying the unit eigenvectors of Step~3 (of~\cite[Section 2]{Ng02-bis}) by the corresponding square roots of eigenvalues and skipping step~4. We remark that the two last changes to~\cite[Section 2]{Ng02-bis} are commonly used so that the similarity matrix $\hat{\bS}$ can be considered as a Gram matrix, see e.g., Euclidean MDS~\cite{Cox01MDS} and
ISOMAP~\cite{Tenenbaum00ISOmap}.

\begin{algorithm}[htbp]
\caption{SLBF: spectral clustering based on best-fit flats} \label{alg:lbfsc}
\begin{algorithmic}
\REQUIRE $\sX=\{\bx_1,\bx_2,\cdots,\bx_N\} \subseteq
\mathbb{R}^{D}$: data, $\lambda$: a parameter (or several parameters if we use step 7, with default values $[2,2e,2e^2,\cdots,2e^6]$), other parameters used by Algorithm~\ref{alg:nhbd}.
\ENSURE A partition of $\sX$ into $K$ disjoint clusters
$\{\sX_i\}_{i=1}^K$, each approximated
by a single flat.\\
\textbf{Steps}:
\STATE
%
%\begin{enumerate}
\squishlistnum
     \item \label{item:slbf_nhbd} For each point $\bx_i$, fit a subspace ${L}_i$ by Algorithm~\ref{alg:nhbd}
     \item \label{item:slbf_S} Construct the $N\times N$ matrix $\bS$ and $\hat{\bS}$ by \eqref{eq:construct_S},  \eqref{eq:construct_hatS} and \eqref{eq:construct_sigma} \\
     \item Let $\bD$ be the $N\times N$ diagonal matrix, such that $\bD_{i,i}=\sum_{j=1}^{N}\hat{\bS}_{i,j}$\\
     \item \label{item:slbf_normalize} Normalize $\hat{\bS}$ by: $\hat{\bS}=\bD^{-\frac{1}{2}}\hat{\bS}\bD^{-\frac{1}{2}}$\\
     \item \label{item:slbf_svd} Let $\bU$ be the $N\times K$ matrix whose columns are the top $K$ eigenvectors of $\hat{\bS}$, and $\bSigma$ be the $K\times K$ matrix representing the top $K$ eigenvalues of $\hat{\bS}$\\
     \item \label{item:slbf_kmeans} Apply $K$-means to the rows of $N\times K$ matrix $\bU\,\bSigma^{1/2}$ and partition $\sX$ accordingly
     \item Repeat steps 2-6 with the default values of $\lambda$ (see input) to obtain several segmentations and choose the segmentation minimizing the error: \begin{equation}\label{eq:choose_segmentation}\sum_{i=1}^{K}\min_{\text{$d$-flat $L$}}\left(\sum_{\bx\in\sX_i}\dist^2(\bx,L_i)\right)\end{equation}
% \end{enumerate}
\squishendnum
\end{algorithmic}
\end{algorithm}

As discussed in~\cite{SubspaceClustering_Vidal},  SLBF is a ``spectral clustering-based method'',  similar to SCC, LSA and SSC. These algorithms construct an $N\times N$ affinity matrix, whose $ij$-th entry represents the similarity between points $i$ and $j$, and then apply spectral clustering using this affinity matrix. Ideally, the affinities of points from the same cluster are of order 1 and the affinities of points from different clusters are of order 0.
Indeed, for the affinity $\hat{\bS}$ of SLBF, if $\bx_i$ and $\bx_j$ are in the same cluster, then we expect that $\bx_i$ is close to $L_j$ and $\bx_j$ is close to $L_i$, which means $\bS_{i,j}$ is close to 0 and thus $\hat{\bS}_{i,j}$ is close to 1 (we assume here that $L_i$ and $L_j$ are good estimators for the underlying subspace of the cluster shared by $\bx_i$ and $\bx_j$ as suggested by Theorem~\ref{thm:main}).  Otherwise, if  $\bx_i$ and $\bx_j$ are  not in the same cluster, then we expect that $\bx_i$ is sufficiently far from $L_j$ and $\bx_j$ is sufficiently far from  $L_i$, which implies that $\hat{\bS}_{i,j}$ is close to $0$. The choice of $\sigma_j$ clearly affects this heuristic argument on the size of $\hat{\bS}_{i,j}$ . Theoretically $\sigma_j$ should be larger than the noise, such that $\hat{\bS}_{i,j}$ is close to $1$ when $\bx_i$ and $\bx_j$ are in the same cluster, but $\sigma_j$ cannot be too large so that $\hat{\bS}_{i,j}$ is close to $1$ when $\bx_i$ and $\bx_j$ are not in the same cluster. Therefore we use \eqref{eq:construct_sigma}, where $\sqrt{\min_{\text{$d$-flats\,\,L}}\sum_{\bx\in\mathcal{N}_j}\|\bx-P_{L}\bx\|^2/|\N_j|}$ is the estimated noise of the data set around the point $\bx_j$ and $\lambda$ is a parameter. Following the strategy in \cite{spectral_applied}, we choose different values of $\lambda$ (our fixed default values are $[2, 2e, 2e^2,\cdots, 2e^6]$) and consequently obtain several segmentation results (7 results when using our default values). We then choose the segmentation with the smallest error in \eqref{eq:choose_segmentation}.

We remark that SLBF is robust to outliers. Indeed, the original data points are embedded as the rows of $\bU\bSigma^\frac{1}{2}$ (see step 6 of Algorithm~\ref{alg:lbfsc}) and have magnitude smaller than 1. Therefore the subsequent application of $K$-means does not suffer from points of arbitrarily large values. To verify that the magnitude of the embedded points is smaller than 1, we note that the diagonal elements in $\hat{\bS}$ are smaller than $1$. Since $\bU\bSigma\bU^T\leq \hat{\bS}$ the diagonal elements of $\bU\bSigma\bU^T$ are also smaller than 1. Therefore, the norm of the rows of $\bU\bSigma^\frac{1}{2}$ are also smaller than 1.

Similar to SLBF, LSA~\cite{Yan06LSA} is also based on fitting local subspaces. However, LSA fits subspace by local neighborhoods of fixed number of points and is not adaptive. Moreover, the local subspaces of LSA are forced to be linear (since the affinity of LSA is based on principal angles between such subspaces) and this further restricts the applicability of LSA. There is also some similarity between the idea of SLBF and that of SCC~\cite{spectral_theory,spectral_applied}. Indeed, we may view SCC as fitting candidate subspaces based on $d+1$ data points (the iterative procedure tries to enforce the points to be from the same cluster). However, in practice they operate very differently, in particular, SCC is not based just on local information (though a local version of SCC follows from~\cite{Arias-Castro11Surfaces}). The SSC algorithm is also a spectral method, but similar to SCC its affinities are global (they are based on sparse representation of data points).

%We remark that spectral clustering-based methods, in particular, LSA, SCC, SSC and SLBF, are usually robust to outliers. One of the reason is that it embeds points within the unit ball (

\subsubsection{ Complexity and storage of the SLBF algorithm}
Step~\ref{item:slbf_nhbd} of Algorithm~\ref{alg:lbfsc} has a complexity of order $O((d\cdot D +\log N)\cdot N^2)$, since it applies Algorithm~\ref{alg:nhbd} to every point in the set $\sX$.
The most expensive calculation of steps~\ref{item:slbf_S}-\ref{item:slbf_normalize} in Algorithm~\ref{alg:lbfsc} is the construction of $\bS$, which requires a complexity of order $O(d\cdot D \cdot N^2)$. The eigenvalue decomposition in step~\ref{item:slbf_svd} has a complexity of order $O(K \cdot N^2)$ and the $K$-means algorithm in step~\ref{item:slbf_kmeans} has a complexity of order $O(n_s\cdot N \cdot K^2)$, where $n_s$ is the iterations in $K$-means.

Combining these complexities together, we have an overall complexity of order $O( (d\cdot D +\log N)\cdot N^2+n_s\cdot N \cdot K^2 )$ for SLBF.  As before, limiting to a constant number of scales replaces the $\log$ term with a constant.
%In comparison, LSA has a complexity of order $O(N^2\cdot (d^2\cdot D +\log N)+n_s\cdot K^2 \cdot N)$, where $n_s$ is the number of iteration in the $K$-means step. This complexity is slightly larger than the complexity of SLBF.

We note that SLBF stores the data set in a $D \times N$ matrix, the candidate subspaces in $N$ $D \times d$ matrices (recall that in SLBF every data point is assigned a subspace and thus $C=N$) and it also uses the $N \times N$ matrix $\bS$. Therefore, the storage of SLBF is in the order of $O(N\cdot D \cdot d+N^2)$.

%\begin{theorem}
%Let the total variation of $\hat{\bS}$ be:
%\[\TV(\hat{\bS}):=\sum_{1\leq k\leq K}\sum_{i\in\sI_k}\|\bu_i-\bc_i\|,\]
%where $\bu_i$ is the $i$-th row in step 6 of Algorithm~\ref{alg:lbfsc} and $\bc_i$ is the center for the $i$-th cluster $\sI_i$ in step 6.
%
%Assume  $\hat{\bS}_0$ corresponds to the clean case: \[\hat{\bS}_0\equiv \left( \begin{array}{cccc}
%\frac{1}{N_1} & 0 & \cdots &0 \\
%0 & \frac{1}{N_2} & \cdots &0 \\
%\vdots & \vdots & \ddots & \vdots\\
%0 & 0 & \cdots &\frac{1}{N_K}
%\end{array} \right),\]
%then $\TV(\hat{\bS})\leq 2\|\hat{\bS}-\hat{\bS}_0\|_*$.
%\end{theorem}
%\begin{proof}
%We have
%\begin{align*}
%\TV(\hat{\bS})=\|\bU-\hat{\bS}_0\bU\|_F^2
%\end{align*}
%Let $\bA=\bU\,\Sigma^{1/2}$ and $\bA_0=\bU_0\,\bSigma_0^{1/2}$, where $\bU_0$ is the $N\times N$ matrix whose columns are all the $N$ eigenvectors of $\hat{\bS}$, and $\Sigma_0$ be the $N\times N$ matrix representing all $N$ eigenvalues of $\hat{\bS}$. Then
%\begin{align*}
%&\|\bU_0(\bI-\hat{\bS}_0)\|_F^2=\tr((\bI-\hat{\bS}_0)\bU_0^T\bU_0(\bI-\hat{\bS}_0))
%=\left\langle\bU_0^T\bU_0,(\bI-\hat{\bS}_0)^2\right\rangle\\
%=&\left\langle\hat{\bS},\bI-\hat{\bS}_0\right\rangle
%=\left\langle\hat{\bS}-\hat{\bS}_0,\bI-\hat{\bS}_0\right\rangle,
%\end{align*}
%and
%\begin{align*}
%\|(\bU-\bU_0)(\bI-\hat{\bS}_0)\|_F^2
%\end{align*}
%\end{proof}
\subsection{Adaptation of the proposed algorithms to motion segmentation data}
\label{sec:eng}
Note that the first minimum in the Theorem~\ref{thm:main} excludes the left
endpoint, and thus $k=0$ is excluded in Algorithm~\ref{alg:nhbd}).  In our experiments, we noticed that on data without too
much noise, it is useful to allow the first scale to count as a
local minimum and allow $k=0$ in Algorithm~\ref{alg:nhbd}). We refer to the implementation of LBF and SLBF with those two techniques tailored for motion segmentation data as LBF-MS and SLBF-MS.

\section{Experimental results}
\label{sec:exp}

\addtocounter{footnote}{1}
\begin{table*}[htbp]

\centering \caption{\small {Mean percentage of misclassified points
in artificial data for linear-subspace cases or affine-subspace case.
}\label{tab:error}}
{\scriptsize
\begin{tabular}{|r|r||r|r||r|r||r|r||r|r||r|r||r|r||r|r||}
 \hline
%    &\multicolumn{12}{c||}{Linear}
%    &\multicolumn{2}{c|}{Affine}\\
   \cline{2-14}
   \multicolumn{2}{|c||}{}
    &\multicolumn{2}{c||}{}
    &\multicolumn{2}{c||}{}
    &\multicolumn{2}{c||}{}
    &\multicolumn{2}{c||}{}
    %&\multicolumn{2}{c||}{$(1,2,3)$}
    &\multicolumn{2}{c||}{$(4,5,6)$}
    &\multicolumn{2}{c||}{$(1,5)$}\\
    \multicolumn{2}{|c||}{\raisebox{1.5ex}[0pt]{\normalsize{Linear}}}
    &\multicolumn{2} {c||}{\raisebox{1.5ex}[0pt]{$2^2 \in\mathbb{R}^4$}}
    &\multicolumn{2}{c||}{\raisebox{1.5ex}[0pt]{$4^2 \in\mathbb{R}^6$}}
    &\multicolumn{2}{c||}{\raisebox{1.5ex}[0pt]{$2^4 \in\mathbb{R}^4$}}
    &\multicolumn{2}{c||}{\raisebox{1.5ex}[0pt]{$10^2 \in\mathbb{R}^{15}$}}
    %&\multicolumn{2}{c||}{{$\in\mathbb{R}^{5}$}}
    &\multicolumn{2}{c||}{$\in\mathbb{R}^{10}$}
    &\multicolumn{2}{c||}{$\in\mathbb{R}^{6}$}\\
    \cline{2-14}

    \cline{1-14}

    \multicolumn{2}{|c||}{  Outl.  \%   }    & 0 & 30 & 0 & 30 & 0 & 30 & 0 & 30 & 0 & 30&0 & 30\\
  \hline\hline
 \rowcolor[gray]{.8}& $e\%$ &  2.6&6.9&\textbf{0.0}&2.6&\textbf{0.1}&22.4&0.5&3.8&1.8&28.2&N/A&34.6\\
  \rowcolor[gray]{.8}\raisebox{1.5ex}[0pt]{LSCC}& $t(s)$  &1.1&0.8&1.0&1.8&1.5&2.0&13.3&5.7&5.1&8.4&N/A&1.9\\\hline
& $e\%$ &  2.7&1\textbf{0.0}&\textbf{0.0}&4.1&\textbf{0.1}&36.7&0.7&31.9&1.4&19.8&N/A&32.9\\
\raisebox{1.5ex}[0pt]{LSCC-MS}& $t(s)$ &  1.1&0.5&1.1&1.4&1.7&1.5&5.1&5.6&4.0&4.6&N/A&2.0\\\hline
 \rowcolor[gray]{.8}& $e\%$ &18.4&19.6&0.1&12.7&0.4&21.0&0.1&9.9&5.9&6.6&27.4&35.4\\
  \rowcolor[gray]{.8}\raisebox{1.5ex}[0pt]{LSA}& $t(s)$ &6.8&16.0&7.1&20.8&23.8&54.4&11.7&31.5&20.1&54.4&6.6&13.8\\\hline
& $e\%$ &\textbf{2.5}&15.8&2.5&18.4&\textbf{0.1}&34.3&\textbf{0.0}&33.8&\textbf{1.0}&30.6&20.2&23.5\\
\raisebox{1.5ex}[0pt]{KF}& $t(s)$ &0.5&0.6&0.2&0.8&0.7&1.8&0.4&1.0&\textbf{0.7}&2.8&0.3&0.5\\\hline
 \rowcolor[gray]{.8}& $e\%$ &\textbf{2.5}&14.2&\textbf{0.0}&17.7&\textbf{0.1}&34.2&\textbf{0.0}&38.8&1.6&34.7&23.4&24.0\\
  \rowcolor[gray]{.8}\raisebox{1.5ex}[0pt]{MoPPCA}& $t(s)$ &0.3&0.5&0.2&0.7&0.7&2.0&\textbf{0.2}&1.1&1.1&3.3&0.5&0.5\\\hline
& $e\%$ &6.0&\textbf{2.5}&\textbf{0.0}&\textbf{2.0}&\textbf{0.1}&\textbf{6.3}&\textbf{0.0}&14.6&14.6&N/A&5.9&N/A\\
\raisebox{1.5ex}[0pt]{GPCA}& $t(s)$ &2.1&38.0&1.9&85.2&10.8&136.2&11.2&546.0&73.8&N/A&0.7&N/A\\\hline
%LBF-MS&2.6&3.0&2.4&2.4&6.4&43.9&1.4&1.9&2.8&2.5\\

 \rowcolor[gray]{.8}& $e\%$ &2.8&3.7&\textbf{0.0}&2.3&\textbf{0.1}&11.5&\textbf{0.0}&\textbf{1.9}&1.5&\textbf{1.5}&18.8&14.1\\
  \rowcolor[gray]{.8}\raisebox{1.5ex}[0pt]{LBF}& $t(s)$ &0.6&0.5&0.5&0.5&1.8&2.7&0.6&0.8&1.1&1.4&0.5&0.5\\\hline
& $e\%$ &2.7&3.0&\textbf{0.0}&2.6&\textbf{0.1}&11.7&\textbf{0.0}&2.2&1.3&\textbf{1.5}&19.5&13.7\\
\raisebox{1.5ex}[0pt]{LBF-MS}& $t(s)$ &0.6&0.5&0.4&0.5&1.7&2.6&0.4&\textbf{0.6}&0.9&\textbf{1.3}&\textbf{0.4}&0.4\\\hline
\rowcolor[gray]{.8}& $e\%$ &5.2&6.3&0.1&7.0&\textbf{0.1}&23.9&\textbf{0.0}&6.2&2.0&2.4&11.1&13.5\\
\rowcolor[gray]{.8}\raisebox{1.5ex}[0pt]{SLBF}& $t(s)$ &11.2&20.7&9.4&21.7&65.0&174.9&9.5&23.3&23.2&64.2&9.3&15.3\\\hline
& $e\%$ &7.8&11.7&0.1&6.6&0.2&46.6&\textbf{0.0}&4.8&1.9&2.6&19.7&22.1\\
\raisebox{1.5ex}[0pt]{SLBF-MS}& $t(s)$ &12.0&24.0&8.8&24.4&68.1&202.0&8.4&23.5&22.0&72.4&9.8&16.3\\\hline
 \rowcolor[gray]{.8}& $e\%$  &2.7&2.6&2.9&2.1&8.0&9.4&0.5&5.8&1.7&1.5
&N/A&31.6\\
\rowcolor[gray]{.8}\raisebox{1.5ex}[0pt]{RANSAC (oracle)} & $t(s)$ &\textbf{0.1}&\textbf{0.1}&\textbf{0.1}&\textbf{0.2}&\textbf{0.1}&\textbf{0.2}&5.9&6.7&1.5&7.1
&N/A&\textbf{0.2}\\\hline
& $e\%$ &3.2&2.6&2.1&2.4&7.7&9.8&0.4&6.7&1.8&\textbf{1.5}&N/A&30.6\\
\raisebox{1.5ex}[0pt]{RANSAC ($\epsilon$ from LBF)}& $t(s)$ &\textbf{0.1}&\textbf{0.1}&\textbf{0.1}&\textbf{0.2}&\textbf{0.1}&\textbf{0.2}&5.9&6.7&1.5&7.0&N/A&0.3\\\hline
 \rowcolor[gray]{.8}& $e\%$  &4.1 &   3.4&   0.1&   16.3&   \textbf{0.1}&   30.1&   50.0 &  50.0&    5.4&   36.1&\textbf{0.3}&0.4\\
\rowcolor[gray]{.8}\raisebox{1.5ex}[0pt]{ALC (oracle)}& $t(s)$ &    7.3   &      23.2&         7.7        &33.6     &    28.4     & 136.3       &  13.9     &   172.6 &        23.0       &180.1&7.8&17.3\\\hline
& $e\%$ &4.5 &   5.7 &   0.1 &  10.0 &   \textbf{0.1} &  14.0 &  50.0 &  50.0 & 2.5 &  1.8&0.4&\textbf{0.3}\\
\raisebox{1.5ex}[0pt]{ALC ($\epsilon$ from LBF)}& $t(s)$ &8.0&         28.0&         8.1&         37.9&         29.6   &   121.9      &   16.6      &  152.4     &    24.0  &      151.6&8.3&18.1\\\hline

 %SSC1& 24.1  & 34.1  & 29.6  & 41.1  & 49.4  & 52.9   &13.9  & 44.0   &23.0  & 51.3\\
\rowcolor[gray]{.8}& $e\%$ &  19.5  & 34.3  & 0.2  & 43.5 & 0.4  & 52.8  & 47.0  & 44.9
 &11.5   &54.0&9.4&15.9\\
 \rowcolor[gray]{.8}\raisebox{1.5ex}[0pt]{SSC}& $t(s)$ & 114.8        &236.2      &  97.6        &247.9       & 227.7    &  591.3       & 106.0       & 276.6        &185.5        &437.9&94.1&142.1\\\hline

%   \rowcolor[gray]{.8} SSC3& 21.7  & 31.4  & 27.3  & 40.7 &  47.3  & 50.4   & 15.6  & 43.9  & 23.0 &  51.9\\

  \hline
\end{tabular}
%\end{table}
%\begin{table}[htbp]
%\centering \caption{\small {Mean percentage of misclassified points
%in simulation for affine-subspaces cases}}\label{tab:error2}

\begin{tabular}{|r|r||r|r||r|r||r|r||r|r||r|r||r|r||r|r||}
 \hline
%    &\multicolumn{12}{c||}{Linear}
%    &\multicolumn{2}{c|}{Affine}\\
   \cline{2-14}
   \multicolumn{2}{|c||}{}
    &\multicolumn{2}{c||}{}
    &\multicolumn{2}{c||}{}
    &\multicolumn{2}{c||}{}
    &\multicolumn{2}{c||}{}
    %&\multicolumn{2}{c||}{$(1,2,3)$}
    &\multicolumn{2}{c||}{$(4,5,6)$}
    &\multicolumn{2}{c||}{$(1,5)$}\\
    \multicolumn{2}{|c||}{\raisebox{1.5ex}[0pt]{\normalsize{Linear}}}
    &\multicolumn{2} {c||}{\raisebox{1.5ex}[0pt]{$2^2 \in\mathbb{R}^4$}}
    &\multicolumn{2}{c||}{\raisebox{1.5ex}[0pt]{$4^2 \in\mathbb{R}^6$}}
    &\multicolumn{2}{c||}{\raisebox{1.5ex}[0pt]{$2^4 \in\mathbb{R}^4$}}
    &\multicolumn{2}{c||}{\raisebox{1.5ex}[0pt]{$10^2 \in\mathbb{R}^{15}$}}
    %&\multicolumn{2}{c||}{{$\in\mathbb{R}^{5}$}}
    &\multicolumn{2}{c||}{$\in\mathbb{R}^{10}$}
    &\multicolumn{2}{c||}{$\in\mathbb{R}^{6}$}\\
    \cline{2-14}

    \cline{1-14}

    \multicolumn{2}{|c||}{  Outl.  \%   }      & 0 & 30 & 0 & 30 & 0 & 30 & 0 & 30 & 0 & 30&0&30\\
  \hline\hline
 \rowcolor[gray]{.8}& $e\%$& \textbf{0.0}&0.6&\textbf{0.0}&\textbf{0.0}&\textbf{0.0}&0.5&\textbf{0.0}&0.7&\textbf{0.0}&5.8&N/A&N/A\\
  \rowcolor[gray]{.8}\raisebox{1.5ex}[0pt]{SCC}& $t(s)$&1.2&1.0&1.1&2.0&1.4&2.5&6.1&13.7&5.6&6.0&N/A&N/A\\\hline
& $e\%$&  \textbf{0.0}&2.2&\textbf{0.0}&0.5&\textbf{0.0}&5.8&\textbf{0.0}&\textbf{0.0}&\textbf{0.0}&3.1&N/A&N/A\\
\raisebox{1.5ex}[0pt]{SCC-MS}& $t(s)$&  1.2&0.7&1.2&1.6&1.7&2.2&5.4&6.0&4.6&4.8&N/A&N/A\\\hline
 \rowcolor[gray]{.8}& $e\%$ &0.1&11.0&\textbf{0.0}&4.7&0.4&41.7&\textbf{0.0}&\textbf{0.0}&\textbf{0.0}&1.1&37.5&37.9\\
 \rowcolor[gray]{.8} \raisebox{1.5ex}[0pt]{LSA}& $t(s)$&6.7&16.1&7.1&20.8&22.2&54.0&11.7&32.2&38.3&54.0&6.6&13.9\\\hline
& $e\%$&0.2&15.1&0.1&26.0&0.3&37.1&\textbf{0.0}&24.9&\textbf{0.0}&23.5&24.8&27.1\\
\raisebox{1.5ex}[0pt]{KF}& $t(s)$&0.8&0.6&0.4&0.7&1.0&1.4&0.6&1.7&\textbf{1.0}&1.4&0.5&\textbf{0.5}\\\hline
 \rowcolor[gray]{.8}& $e\%$&0.2&23.7&0.1&38.3&0.5&39.8&\textbf{0.0}&45.2&\textbf{0.0}&46.8&30.8&30.4\\
 \rowcolor[gray]{.8} \raisebox{1.5ex}[0pt]{MoPPCA}& $t(s)$&0.9&0.5&0.5&0.6&1.1&1.4&0.9&1.9&1.9&2.0&0.5&\textbf{0.5}\\\hline
& $e\%$&0.2&18.4&0.2&22.2&0.4&38.1&\textbf{0.0}&27.9&0.3&N/A&N/A&N/A\\
\raisebox{1.5ex}[0pt]{GPCA}& $t(s)$&1.8&43.7&3.3&104.0&8.3&209.3&11.8&501.1&69.1&N/A&N/A&N/A\\\hline
 \rowcolor[gray]{.8}& $e\%$&\textbf{0.0}&2.0&\textbf{0.0}&0.7&\textbf{0.0}&4.5&\textbf{0.0}&0.3&\textbf{0.0}&\textbf{0.0}&4.7&11.2\\
 \rowcolor[gray]{.8}\raisebox{1.5ex}[0pt]{LBF}& $t(s)$&0.7&0.6&0.5&0.6&1.9&2.8&0.6&0.8&1.2&1.5&\textbf{0.4}&0.5\\\hline
& $e\%$&\textbf{0.0}&2.7&\textbf{0.0}&1.5&\textbf{0.0}&5.2&\textbf{0.0}&0.5&\textbf{0.0}&\textbf{0.0}&3.9&10.5\\
\raisebox{1.5ex}[0pt]{LBF-MS}& $t(s)$&0.6&0.5&0.4&\textbf{0.5}&1.7&2.7&\textbf{0.4}&\textbf{0.6}&\textbf{1.0}&\textbf{1.3}&\textbf{0.4}&\textbf{0.4}\\\hline
\rowcolor[gray]{.8}& $e\%$&\textbf{0.0}&1.0&\textbf{0.0}&\textbf{0.0}&\textbf{0.0}&\textbf{0.1}&\textbf{0.0}&\textbf{0.0}&\textbf{0.0}&\textbf{0.0}&\textbf{0.0}&\textbf{0.0}\\
\rowcolor[gray]{.8}\raisebox{1.5ex}[0pt]{SLBF}& $t(s)$&9.3&19.1&5.8&19.0&37.7&143.1&6.3&19.4&35.1&61.4&5.9&14.8\\\hline
& $e\%$&\textbf{0.0}&0.1&\textbf{0.0}&\textbf{0.0}&\textbf{0.0}&\textbf{0.1}&\textbf{0.0}&\textbf{0.0}&\textbf{0.0}&\textbf{0.0}&\textbf{0.0}&\textbf{0.0}\\
\raisebox{1.5ex}[0pt]{SLBF-MS}& $t(s)$&8.8&21.7&5.6&21.9&38.0&175.5&5.9&21.1&40.1&66.7&5.9&14.3\\\hline
%LBF-MS&0.1&1.1&0.1&1.3&0.4&32.4&0.0&0.2&0.0&0.7&&\\
% \rowcolor[gray]{.8}  LBF&0.2&2.1&0.1&1.8&0.5&3.7&0.0&0.5&0.0&0.0&&\\
%LBF-MS&0.4&2.0&0.1&2.6&0.7&6.0&0.0&0.3&0.0&0.0&&\\
% \rowcolor[gray]{.8}WLBF&0.1&0.1&0.1&0.1&0.4&1.0&0.0&0.5&0.0&0.3&&\\
% WLBF-MS&0.5&0.0&0.5&0.0&0.6&0.1&1.0&0.0&3.3&0.0&&\\
%\rowcolor[gray]{.8} %SLBF&0.0&0.0&0.0&0.0&0.1&0.1&0.0&0.0&0.0&0.0&&\\
% SLBF-MS&0.0&0.0&0.0&0.0&0.1&0.1&0.0&0.0&0.0&0.0&&\\

 \rowcolor[gray]{.8}& $e\%$&13.8&11.6&9.8&9.6&30.9&27.0&1.9&8.3&1.2&3.4&N/A&23.6\\
  \rowcolor[gray]{.8} \raisebox{1.5ex}[0pt]{RANSAC (oracle)}& $t(s)$&\textbf{0.1}&\textbf{0.2}&\textbf{0.4}&1.8&\textbf{0.4}&\textbf{0.8}&6.4&6.8&3.7&7.4&N/A&\textbf{0.5}\\\hline
& $e\%$&13.6&11.6&11.6&10.4&29.9&28.5&1.4&9.6&1.2&2.4&N/A&23.1\\
 \raisebox{1.5ex}[0pt]{ RANSAC ($\epsilon$ from LBF)}& $t(s)$&\textbf{0.1}&\textbf{0.2}&\textbf{0.4}&1.9&\textbf{0.4}&\textbf{0.8}&6.4&6.7&3.7&7.4&N/A&\textbf{0.5}\\\hline
   \rowcolor[gray]{.8}& $e\%$& \textbf{0.0}&     \textbf{0.0}&     \textbf{0.0}&     \textbf{0.0}&   \textbf{0.0} &  25.1 &    \textbf{0.0}&   40.0&     \textbf{0.0}&   65.0&\textbf{0.0}&\textbf{0.0}\\
   \rowcolor[gray]{.8}
  \raisebox{1.5ex}[0pt]{ALC (oracle)}& $t(s)$&    17.6     &   25.2   &      16.6   &      39.1&         64.2      &119.3 &        20.0&         43.0  &       39.7      &   92.7&18.3&36.8\\\hline
& $e\%$& \textbf{0.0}&     0.4&     \textbf{0.0}&     \textbf{0.0}&     \textbf{0.0}&     0.3&     \textbf{0.0}&     \textbf{0.0}&     \textbf{0.0}&     \textbf{0.0}&\textbf{0.0}&\textbf{0.0}\\
  \raisebox{1.5ex}[0pt]{ ALC ($\epsilon$ from LBF)}& $t(s)$&18.7   &      26.8         &17.2      &  29.8         &65.2   &     113.6  &       24.4   &      55.5    &     47.9  &       85.2&18.8&38.9\\\hline
 % \rowcolor[gray]{.8}
 % \rowcolor[gray]{.8} SSC1&1.1   & 2.5    & 0.1     &0.1  & 10.4 &   9.7 &    0.0&     0.0&     0.0    & 0.0&&\\
\rowcolor[gray]{.8}& $e\%$&\textbf{0.0}   & 1.9    & \textbf{0.0}     &0.1  & 0.1 &   6.4 &    \textbf{0.0}&     \textbf{0.0}&     \textbf{0.0}    & \textbf{0.0}&\textbf{0.0}&\textbf{0.0}\\
  \rowcolor[gray]{.8} \raisebox{1.5ex}[0pt]{SSC}& $t(s)$& 135.9 &       226.8    &    176.0  &     134.7 &     283.8   &      592.4  &      187.0  &      311.9   &     338.6  &      504.1&127.1&183.9\\\hline

  \hline
\end{tabular}
}
\end{table*}

In this section, we conduct experiments on artificial and real data
sets to verify the effectiveness of the proposed algorithm in
comparison to other HLM algorithms.  We will see that in many situations, the methods we propose are fast and accurate; however, in Section \ref{sec:face_seg} we will show a failure mode of our method, and discuss how this can be corrected.
%\subsection{The definition of misclassification rate in this section}

We measure the accuracy of those algorithms by the rate of
misclassified points with outliers excluded, that is
%\begin{align*}
%&\text{error}\%= 100\% \times\\
%&\frac{\text{\# of misclassified points}-\text{\# of misclassified
%outliers}} {\text{\# of points}- \text{\# of outliers}}.
\begin{equation}
\text{error}\%= \frac{\text{\# of misclassified inliers}}{\text{\#
of total inliers}} \times 100\%\,.
%\end{align*}
\label{eq:misclassifcation}\end{equation}

In all the experiments below, the number $C$ in
Algorithm~\ref{alg:candidate} is $70 \cdot K$, where $K$ is the number of subspaces,
the number $p$ in Algorithm~\ref{alg:candidate} is $5 \cdot K$, and the numbers $S$ and $T$ in Algorithm~\ref{alg:nhbd}
are  $2 \cdot d$ and 2 respectively, where $d$ is the dimension of the subspace.
According to our experience, LBF and SLBF are very robust to changes in parameters, but unsurprisingly, there is a general trade off between
accuracy (higher $C$, higher $p$, smaller $T$), and run time (lower
$C$, lower $p$, larger $T$).  We have chosen these parameters for a
balance between run time and accuracy. Nevertheless, we have insisted to use the same parameters for all data sets and experiments, even though
particular parameters could obtain even better or near perfect results for some of the data sets.
The experiments in Sections~\ref{sec:sim} and~\ref{sec:Hopkins155} run on a computer with Intel Core 2 CPU at 2.66GHz and 2 GB memory, and experiments in Sections~\ref{sec:face_seg} and~\ref{sec:MNIST_seg} run on a machine with Intel Core 2 Quad Q6600 at 2.4GHz and 8 GB memory.

\subsection{Clustering results on artificial data}
\label{sec:sim}

%\footnotetext[\value{footnote2}]{For some cases the RANSAC algorithm report error, then it is recorded as N/A.}

We compare our algorithms with the following algorithms: Mixtures of
PPCA (MoPPCA)~\cite{Tipping99mixtures}, $K$-flats (KF)~\cite{Ho03},
Local Subspace Analysis (LSA)~\cite{Yan06LSA}, Spectral Curvature
Clustering (SCC)~\cite{spectral_applied}, Random Sample Consensus
(RANSAC) for HLM~\cite{Yang06Robust}, Agglomerative Lossy
Compression (ALC)~\cite{Ma07Compression} and GPCA with voting/robust GPCA (GPCA)~\cite{Ma07,RobustGPCA}. Throughout the rest of the paper, we use the Matlab codes of the GPCA, MoPPCA and
KF algorithms from http://perception.\\csl.uiuc.edu/gpca, the LSA algorithm from http://www.\\vision.jhu.edu/db, the SCC
algorithm from http://www.math.\\umn.edu/$\sim$lerman/scc, the ALC algorithm from http://\\perception.csl.uiuc.edu/coding/motion/, the RANSAC algorithm from http://www.vision.jhu.edu/code/ and the SSC algorithm from http://www.cis.jhu.edu/$\sim$ehsan/ssc.htm. %Three different options are available in the spectral clustering step of SSC, though the code and paper~\cite{ssc09} do not specify which one to choose;  we thus report all of them as SSC1, SSC2 and SSC3.

For the SCC algorithm, we also try a slightly modified version tailored for motion segmentation as in step~\ref{item:slbf_kmeans} of Algorithm~\ref{alg:lbfsc}, which we refer to as SCC-MS (SCC for motion segmentation): Following the notation
of~[Algorithm~2]\cite{spectral_applied} we let the matrix $\bU$ be the $N \times K$ matrix whose columns are the top  $K$ left
singular vectors of $\bA_C^*$ and also denote by $\bSigma$ the diagonal $K \times K$ matrix whose elements are the top  $K$ left
singular values of $\bA_C^*$.  Then the $K$-means step of SCC-MS is applied directly to the rows of the
$N\times K$ matrix $\bU\,\bSigma^{1/2}$ (as opposed to applying it to $U$ (or its row-wise normalization by 1) in SCC).

The MoPPCA algorithm is always initialized with a random guess of
the membership of the data points. The LSCC algorithm is initialized
by randomly picking $100\times K$ $(d+1)$-tuples (following
~\cite{spectral_applied}) and KF
is initialized with a random guess. Since algorithms like KF tend to
converge to local minimum, we use 10 restarts for MoPPCA, 30
restarts for KF, and recorded the misclassification rate of the one
with the smallest $\ell_2$ error for both of these algorithms. The
number of restarts was restricted by the running time and accuracy. In SSC algorithm, we set the value $\lambda$ to be $0.01$, as suggested in the code.

The RANSAC for HLM and ALC algorithms~\cite{Yang06Robust,Ma07Compression} depend on a user supplied inlier threshold.
RANSAC (oracle) and ALC (oracle) use the oracle inlier bound given by the true noise variance of the model and thus clearly have
an advantage over the other algorithms listed. RANSAC ($\epsilon$ from LBF) and ALC ($\epsilon$ from LBF) estimate the inlier threshold by the local best-fit flats heuristic of this paper. That is, they fit best-fit neighborhoods for all $N$ points using the latter heuristic and estimate the least error of approximation by $d$-flats in these $N$ neighborhoods. The inlier bound $\epsilon$ is then the average of these errors. If the number of clusters resulting from ALC ($\epsilon$ from LBF or oracle) is larger than $K$, then we choose the $K$ largest clusters and identify the points in the rest of clusters as outliers.
For some cases the RANSAC algorithm breaks down and we then report it as N/A. The reason for this is that RANSAC (for HLM)~\cite{Yang06Robust} is very sensitive to the estimate of $\eps$ and an overestimate can result in removal of points belonging to more than one subspace, so that the algorithm may exhaust all points before detecting $K$ subspaces.
%We remark that due to outliers we could not effectively use the voting procedure for ALC (described later).

We remark that GPCA cannot naturally deal with outliers, therefore we use robust GPCA with Multivariate Trimming~\cite{RobustGPCA} and the parameters `angleTolerance' and `boundarythreshold' are set to be 0.3 and 0.4 respectively.
%For ALC algorithm, Rao et al.~\cite{ALC_motion} suggested to label a point as an outlier if it belongs to a group with
%less than five samples. We use a similar strategy: we consider the $K$ largest clusters as inliers and label the points in other smaller clusters as outliers.
%We remark that we have also tried versions of ALC explained in later sections, but did not get good results. We decided not to report them, since ALC tries to estimate the number of clusters and thus the comparison is unfair with the other algorithms that require this number.

\begin{table*}[htbp!]

\centering \caption{\label{tab:comp1}  \small The mean and median
percentage of misclassified points for two-motions and three-motions
in Hopkins 155 database. } \vspace{.1in}
{\scriptsize
\begin{tabular}{|l||r|r||r|r||r|r||r|r|}
  \hline

    &\multicolumn{2}{c||}{Checker}
    &\multicolumn{2}{c||}{Traffic}
    &\multicolumn{2}{c||}{Articulated}
    &\multicolumn{2}{c|}{All}\\
    \cline{2-9}
%$e_{\%}$
%    &Mean &Median &Mean &Median &Mean &Median &Mean &Median\\
%  \hline\hline
    \raisebox{1.5ex}[0pt]{\normalsize{2-motion}} &Mean &Median &Mean &Median &Mean &Median &Mean &Median\\
  \hline\hline
 % CCS&16.37&10.64&5.27&0.00&17.58&7.07&12.16&0.00\\

  GPCA     & 6.09&1.03&1.41&\textbf{0.00}&2.88&\textbf{0.00}&4.59&0.38   \\
%KF&5.33&0.04&2.36&0.00&3.83&1.11&4.43&0.00\\
%KF 4$K$&5.81&0.17&3.55&0.02&4.97&1.15&5.15&0.06\\

%KF 5&11.35&5.47&4.57&1.43&12.47&5.54&9.70&3.65\\
%KF(R)&15.37&6.96&15.93&8.61&12.73&6.63&15.27&7.29\\

%LLMC $4K$&4.65&0.11&3.65&0.33&5.23&1.30&4.44&0.24\\

 \rowcolor[gray]{.8} LLMC 5&4.37&\textbf{0.00}&0.84&\textbf{0.00}&6.16&1.37&3.62&\textbf{0.00}\\
LSA 4$K$&  2.57&0.27&5.43&1.48&4.10&1.22&3.45&0.59\\
%LSA 5&8.84&3.43&2.15&1.00&4.66&1.28&6.73&1.99\\
 \rowcolor[gray]{.8} LBF(4$K$,3)&3.65&\textbf{0.00}&3.89&\textbf{0.00}&4.43&0.15&3.78&\textbf{0.00}\\
LBF-MS(4$K$,3)&2.90&\textbf{0.00}
&1.64&\textbf{0.00}&2.51&0.06&2.54&\textbf{0.00}\\
\rowcolor[gray]{.8}SLBF(2$F$,3)&1.59&\textbf{0.00}&\textbf{0.20}&\textbf{0.00}&\textbf{0.80}&\textbf{0.00}&1.16&\textbf{0.00}\\
 SLBF-MS(2$F$,3)&\textbf{1.28}&\textbf{0.00}&0.21&\textbf{0.00}&0.94&\textbf{0.00}&\textbf{0.98}&\textbf{0.00}\\

% \rowcolor[gray]{.8} LBF(4$K$,3)&3.31&\textbf{0.00}&3.29&0.00&4.31&0.12&3.40&0.00\\
%LBF-MS(4$K$,3)&3.05&0.00&0.78&0.00&1.73&0.03&2.34&0.00\\
%\rowcolor[gray]{.8}WLBF(4$K$,3)&2.56&0.00&2.01&0.00&3.93&0.00&2.54&0.00\\
% WLBF-MS(4$K$,3)&1.94&0.00&0.85&0.00&2.66&0.17&1.72&0.00\\
%\rowcolor[gray]{.8}SLBF(2$F$,3)&1.32&0.00&0.18&0.00&2.84&0.00&1.16&0.00\\

%SLBF-MS(2$F$,3)&0.62&0.00&1.21&0.00&1.04&0.00&0.81&0.00\\

  \rowcolor[gray]{.8}SCC(4$K$,3)&2.42&\textbf{0.00}&4.44&\textbf{0.00}&9.51&2.04&3.60&\textbf{0.00}\\
 SCC-MS(4$K$,3)&2.00&\textbf{0.00}&0.35&\textbf{0.00}&4.11&1.12&1.77&\textbf{0.00}\\
 % \rowcolor[gray]{.8}SSC-N1&1.29 &    \textbf{0.00}  &  0.24 &    \textbf{0.00}&    0.97 &    \textbf{0.00} &   0.99 &    \textbf{0.00}\\
\rowcolor[gray]{.8}SSC-N(4$K$,3)&1.29&     \textbf{0.00} &   0.29 &    \textbf{0.00} &   0.97 &   \textbf{0.00} &  1.00    & \textbf{0.00}\\
 % \rowcolor[gray]{.8}SSC-N3&1.39&     \textbf{0.00} &   0.40 &    0.00 &   1.37 &   0.00 &  1.13    & 0.0\\
  MSL&4.46&\textbf{0.00}&2.23&\textbf{0.00}&7.23&\textbf{0.00}&4.14&\textbf{0.00}\\
 \rowcolor[gray]{.8}RANSAC&6.52&1.75&2.55&0.21&7.25&2.64&5.56&1.18\\

%LSCC 5&2.45&0.332&3.46&0.199&9.52&3.9&3.36&0.398\\
%LSCC $4K$&1.39&0.00&2.91&0.199&12.4&1.37&2.79&0.00\\
%LSCC&1.83&0.00&7.13&0.199&10.3&1.37&3.97&0.00\\
%\textbf{MKF}&3.70&0.00&0.90&0.00&6.80&0.00&3.26&0.00\\
%MKF $4K$&4.51&0.01&1.59&0.00&6.08&0.92&3.90&0.00\\

%MKF 5&9.37&4.10&3.47&0.00&10.68&5.84&7.97&2.39\\
%MKF(R)&29.06&31.34&16.78&12.49&25.55&27.54&25.57&28.31\\

  \hline

\end{tabular}

\begin{tabular}{|l||r|r||r|r||r|r||r|r|}
  \hline

    &\multicolumn{2}{c||}{Checker}
    &\multicolumn{2}{c||}{Traffic}
    &\multicolumn{2}{c||}{Articulated}
    &\multicolumn{2}{c|}{All}\\
    \cline{2-9}
%$e_{\%}$
%    &Mean &Median &Mean &Median &Mean &Median &Mean &Median\\
%  \hline\hline
    \raisebox{1.5ex}[0pt]{\normalsize{3-motion}}  &Mean &Median &Mean &Median &Mean &Median &Mean &Median\\
  \hline\hline
 % CCS& 28.63&33.21&3.02&0.18&44.89&44.89&26.18&31.74\\
GPCA&  31.95&32.93&19.83&19.55&16.85&28.66&28.66&28.26\\
%KF&15.61&11.26&5.63&0.57&13.55&13.55&13.50&6.53\\

%KF 4$K$&16.12&11.37&7.06&0.75&16.66&16.66&14.34&7.11\\

%KF 5&26.95&31.88&8.09&5.67&17.65&17.65&22.65&25.08\\
%KF(R)&21.83&24.52&8.70&5.00&15.85&15.85&18.86&17.81\\
 \rowcolor[gray]{.8} LLMC 4$K$&12.01&9.22&7.79&5.47&9.38&9.38&11.02&6.81\\

LLMC 5&10.70&9.21&2.91&\textbf{0.00}&5.60&5.60&8.85&3.19\\
 \rowcolor[gray]{.8} LSA 4$K$&5.80&1.77&25.07&23.79&7.25&7.25&9.73&2.33\\
LSA 5&30.37&31.98&27.02&34.01&23.11&23.11&29.28&31.63\\

 \rowcolor[gray]{.8} LBF(4$K$,3)&8.50&1.26&16.31&13.52&20.75&20.75&10.77&1.70\\
LBF-MS(4$K$,3)&6.97&1.15&7.06&0.62&21.38&21.38&7.81&0.98\\
\rowcolor[gray]{.8}SLBF(2$F$,3)&4.57&0.94&0.38&\textbf{0.00}&2.66&2.66&3.63&0.64\\
 SLBF-MS(2$F$,3)&3.33&0.39&\textbf{0.24}&\textbf{0.00}&\textbf{2.13}&\textbf{2.13}&2.64&\textbf{0.22}\\

% \rowcolor[gray]{.8} LBF(4$K$,3)&8.42&1.29&14.80&9.21&20.45&20.45&10.38&1.63\\
%LBF-MS(4$K$,3)&6.87&1.47&1.40&0.00&24.10&24.10&6.76&0.89\\
%\rowcolor[gray]{.8}WLBF(4$K$,3)&5.30&0.92&8.34&1.20&21.73&21.73&6.85&0.84\\

% WLBF-MS(4$K$,3)&4.48&0.86&1.65&0.00&18.33&18.33&4.70&0.60\\
%\rowcolor[gray]{.8}SLBF(2$F$,3)&4.47&1.10&0.38&0.00&1.60&1.60&3.49&0.89\\

%SLBF-MS(2$F$,3)&2.74&0.64&0.00&0.00&1.60&1.60&2.12&0.30\\

 \rowcolor[gray]{.8} SCC(4$K$,3)&7.80&1.04&8.05&2.37&7.07&7.07&7.81&0.67\\

SCC-MS(4$K$,3)&6.28&0.80&4.09&0.58&7.22&7.22&5.89&0.68\\

%\rowcolor[gray]{.8}SSC-N1&3.23 &   0.29  &  0.53  &   0.00 &  1.60 &  1.60 &  2.60&    0.22\\
\rowcolor[gray]{.8}SSC-N(4$K$,3)&\textbf{3.22} &  \textbf{ 0.29}  &  0.53  &   \textbf{0.00} &  \textbf{2.13}&\textbf{2.13} &  \textbf{2.62}&    \textbf{0.22}\\
 % \rowcolor[gray]{.8}SSC-N3&3.33 &   0.51 &  7.06 &   0.91 &  2.13 &  2.13&   4.01 &   0.56\\

 MSL&10.38&4.61&1.80&\textbf{0.00}&2.71&2.71&8.23&1.76\\
 \rowcolor[gray]{.8}RANSAC&25.78&26.01&12.83&11.45&21.38&21.38&22.94&22.03\\

  \hline
\end{tabular}
}
\end{table*}

\begin{table*}[htbp!]

\centering \caption{\label{tab:std}  \small The standard deviation to the mean and median
percentage of misclassified points for two-motions and three-motions
in Hopkins 155 database. } \vspace{.1in}
{\scriptsize
\begin{tabular}{|l||r|r||r|r||r|r||r|r|}
  \hline

    &\multicolumn{2}{c||}{Checker}
    &\multicolumn{2}{c||}{Traffic}
    &\multicolumn{2}{c||}{Articulated}
    &\multicolumn{2}{c|}{All}\\
    \cline{2-9}
%$e_{\%}$
%    &Mean &Median &Mean &Median &Mean &Median &Mean &Median\\
%  \hline\hline
    \raisebox{1.5ex}[0pt]{\normalsize{2-motion}} &Mean &Median &Mean &Median &Mean &Median &Mean &Median\\
  \hline\hline
\rowcolor[gray]{.8}LBF(4$K$,3)&0.71&\textbf{0.00}&1.22&\textbf{0.00}&1.04&0.66&0.50&\textbf{0.00}\\
LBF-MS(4$K$,3)&0.53&\textbf{0.00}&1.06&\textbf{0.00}&1.14&0.28&0.47&\textbf{0.00}\\
\rowcolor[gray]{.8}WLBF(4$K$,3)&0.53&\textbf{0.00}&0.98&\textbf{0.00}&1.35&\textbf{0.00}&0.47&\textbf{0.00}\\
% WLBF-MS(4$K$,3)&0.46&\textbf{0.00}&0.57&0.00&0.92&0.44&0.36&0.00\\
%\rowcolor[gray]{.8}SLBF(4$K$,3)&0.00&0.00&0.00&0.00&0.00&0.00&0.00&0.00\\
 SLBF-MS(4$K$,3)&\textbf{0.00}&\textbf{0.00}&\textbf{0.00}&\textbf{0.00}&\textbf{0.00}&\textbf{0.00}&\textbf{0.00}&\textbf{0.00}\\

  \rowcolor[gray]{.8}SCC(4$K$,3)&0.27&\textbf{0.00}&1.51&\textbf{0.00}&2.34&1.52&0.38&\textbf{0.00}\\
 SCC-MS(4$K$,3)&0.33&\textbf{0.00}&0.25&\textbf{0.00}&1.03&0.46&0.25&\textbf{0.00}\\
  %\rowcolor[gray]{.8}SSC-N1&\textbf{0.00}&0.00&0.00&0.00&0.00&0.00&0.00&0.00\\

\rowcolor[gray]{.8}SSC-N(4$K$,3)&\textbf{0.00}&\textbf{0.00}&\textbf{0.00}&\textbf{0.00}&\textbf{0.00}&\textbf{0.00}&\textbf{0.00}&\textbf{0.00}\\
%\rowcolor[gray]{.8}SSC-N3&0.00&0.00&0.00&0.00&0.00&0.00&0.00&0.00\\

  \hline

\end{tabular}

\begin{tabular}{|l||r|r||r|r||r|r||r|r|}

  \hline

    &\multicolumn{2}{c||}{Checker}
    &\multicolumn{2}{c||}{Traffic}
    &\multicolumn{2}{c||}{Articulated}
    &\multicolumn{2}{c|}{All}\\
    \cline{2-9}
%$e_{\%}$
%    &Mean &Median &Mean &Median &Mean &Median &Mean &Median\\
%  \hline\hline
    \raisebox{1.5ex}[0pt]{\normalsize{3-motion}}  &Mean &Median &Mean &Median &Mean &Median &Mean &Median\\
  \hline\hline
\rowcolor[gray]{.8}LBF(4$K$,3)&1.52&0.58&3.71&9.69&7.37&7.37&1.43&0.65\\
LBF-MS(4$K$,3)&1.48&0.45&3.81&2.35&6.59&6.59&1.42&0.40\\
% \rowcolor[gray]{.8}WLBF(4$K$,3)&1.30&0.38&3.82&3.05&5.96&5.96&1.25&0.39\\

%WLBF-MS(4$K$,3)&1.23&0.38&1.36&0.00&2.88&2.88&1.04&0.34\\
 \rowcolor[gray]{.8}SLBF(4$K$,3)&\textbf{0.00}&\textbf{0.00}&\textbf{0.00}&\textbf{0.00}&\textbf{0.00}&\textbf{0.00}&\textbf{0.00}&\textbf{0.00}\\

SLBF-MS(4$K$,3)&\textbf{0.00}&\textbf{0.00}&\textbf{0.00}&\textbf{0.00}&\textbf{0.00}&\textbf{0.00}&\textbf{0.00}&\textbf{0.00}\\

\rowcolor[gray]{.8} SCC(4$K$,3)&1.20&0.53&5.70&7.00&1.77&1.77&1.43&0.49\\

SCC-MS(4$K$,3)&0.94&0.50&3.25&0.54&2.54&2.54&0.92&0.33\\
%  \rowcolor[gray]{.8}SSC-N1&0.00&0.00&0.00&0.00&0.00&0.00&0.00&0.00\\

\rowcolor[gray]{.8}SSC-N(4$K$,3)&\textbf{0.00}&\textbf{0.00}&\textbf{0.00}&\textbf{0.00}&\textbf{0.00}&\textbf{0.00}&\textbf{0.00}&\textbf{0.00}\\
%\rowcolor[gray]{.8}SSC-N3&0.00&0.00&0.00&0.00&0.00&0.00&0.00&0.00\\

  \hline
\end{tabular}
}
\end{table*}

The artificial data represents various instances of $K$ linear
subspaces in $\mathbb{R}^D$. If their dimensions are fixed and equal
$d$, we follow~\cite{spectral_applied} and refer to the setting as
$d^K\in \mathbb{R}^D$. If they are mixed, then we follow~\cite{Ma07}
and refer to the setting as $(d_1, \ldots, d_K) \in \mathbb{R}^D$.
Fixing $K$ and $d$ (or $d_1, \ldots, d_K$), we randomly generate 100
different instances of corresponding hybrid linear models according
to the code in http://perceptio\\n.csl.uiuc.edu/gpca. More precisely,
for each of the 100 experiments, $K$ linear subspaces of the
corresponding dimensions in $\reals^D$ are randomly generated.
The random variables sampled within each subspace are sums of two other variables. One of them is sampled from a uniform distribution in a $d$-dimensional ball of radius $1$ of that subspace (centered at the origin for the case of linear subspaces). The other is sampled from a $D$-dimensional multivariate normal distribution with mean $\mathbf{0}$ and covariance matrix $0.05^2\cdot \mathbf{I}_{D\times D}$. Then,
for each subspace 250 samples are generated according to the distribution just described. Next, the data is further corrupted
with 5\% or 30\% uniformly distributed outliers in a cube of sidelength determined by the maximal distance of the former 250
samples to the origin (using the same code).

Since most algorithms (SCC, LSA, MoPPCA, LBF, SLBF, RANSAC, SSC) do not support mixed
dimensions natively, we assume each subspace has the maximum
dimension in the experiment. GPCA and ALC support mixed dimensions natively, and GPCA is the only algorithm for which we specify the dimensions for each subspace in mixed-dimension case (note that the knowledge of dimensions are unnecessary in ALC algorithm).

The mean (over 100 instances) misclassification rates and the mean  running
time of the various
algorithms are recorded in Table~\ref{tab:error}.
From Table \ref{tab:error} we can see that our algorithms, i.e., LBF, LBF-MS, SLBF, SLBF-MS, perform
well in various artificial instances of hybrid linear modeling (with
both linear subspaces and affine subspaces), and their advantage is
especially obvious with many outliers and affine subspaces.
%This corresponds to what we has proposed:
%$\ell_1$ error is more robust to outliers. However, it did not do it
%as well in mixed dimensions with large difference of dimension.
The robustness to outliers is a result of our use of both $\ell_1$
loss function (see e.g., \cite{lp_recovery_part1_11,lp_recovery_part2_11}) and random sampling.
The SLBF and SLBF-MS are better at the affine cases because of  their use of spectral clustering. Also
unlike many other methods, the proposed methods natively support
affine subspace models (the particular data has non-intersecting subspaces, which makes advantageous to some other
algorithms, e.g., SSC).
The results of RANSAC ($\epsilon$ from LBF) and ALC ($\epsilon$ from LBF) show that the local best-fit heuristic can be effectively used to estimate the main parameter of RANSAC and ALC, i.e., to estimate the local noise.
Table \ref{tab:error} also shows that the running time of LBF/LBF-MS is less than the running time of most other algorithms,
especially GPCA, LSA, RANSAC, ALC and SSC.   The difference is large enough that
we can also use the proposed algorithm as an initialization for the
others.  LBF and LBF-MS algorithms are slower than a single run of $K$-flats, but
it usually takes many restarts of $K$-flats to get a decent result.
Notice that the choice of $C$ and $p$ in our algorithm function in a
similar manner to the number of restarts in KF. SLBF and SLBF-MS cost more time when $N$ is large,
because of the construction of the $N\times N$ matrix in spectral clustering, but it still
has a comparable speed to LSA and is faster than SSC, which are two spectral-clustering based algorithms.

\subsection{Clustering results on motion segmentation data}\label{sec:Hopkins155}
\begin{figure}
\begin{center}
\includegraphics[width=.4\textwidth,height=.3\textwidth]{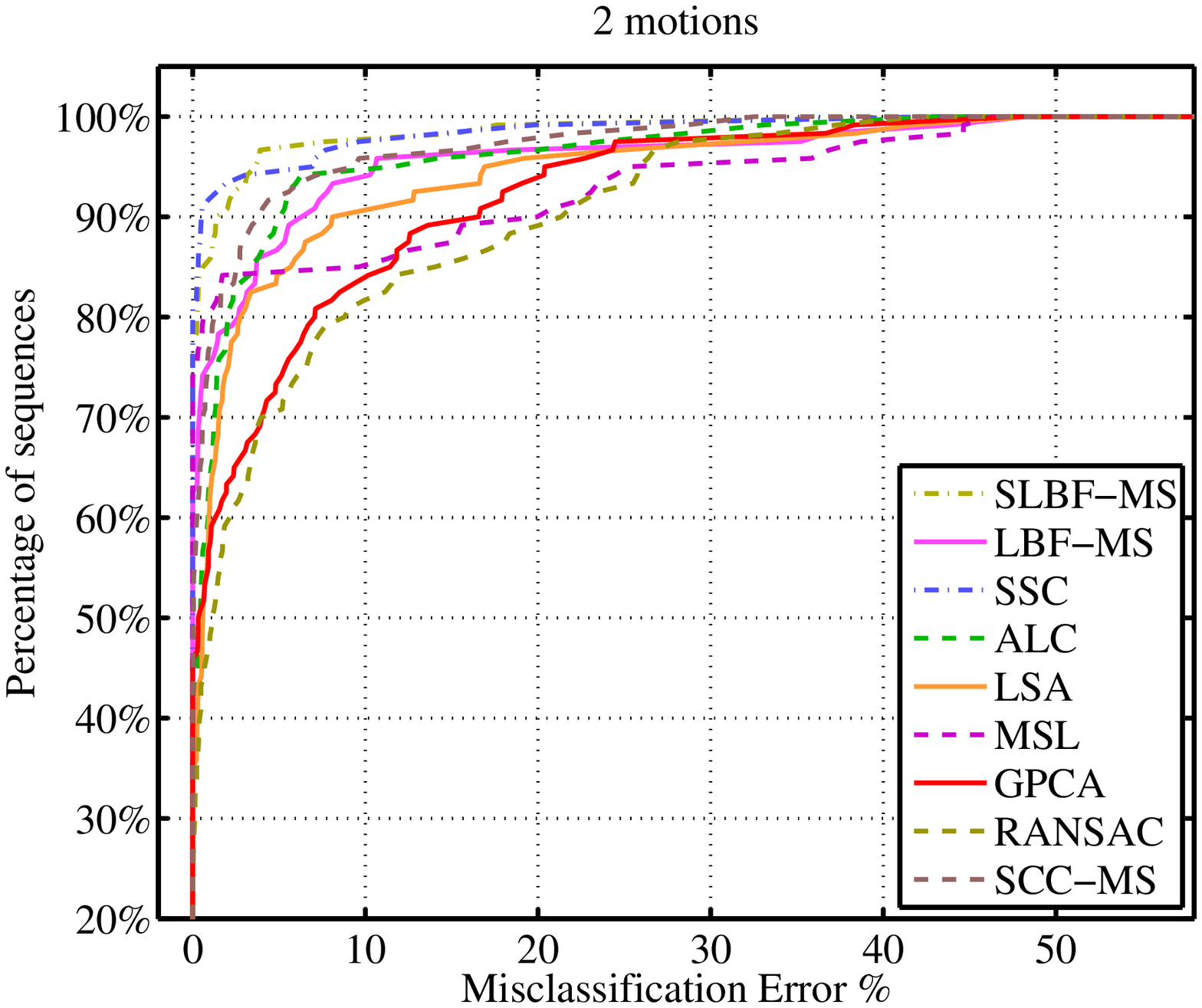}

\includegraphics[width=.4\textwidth,height=.3\textwidth]{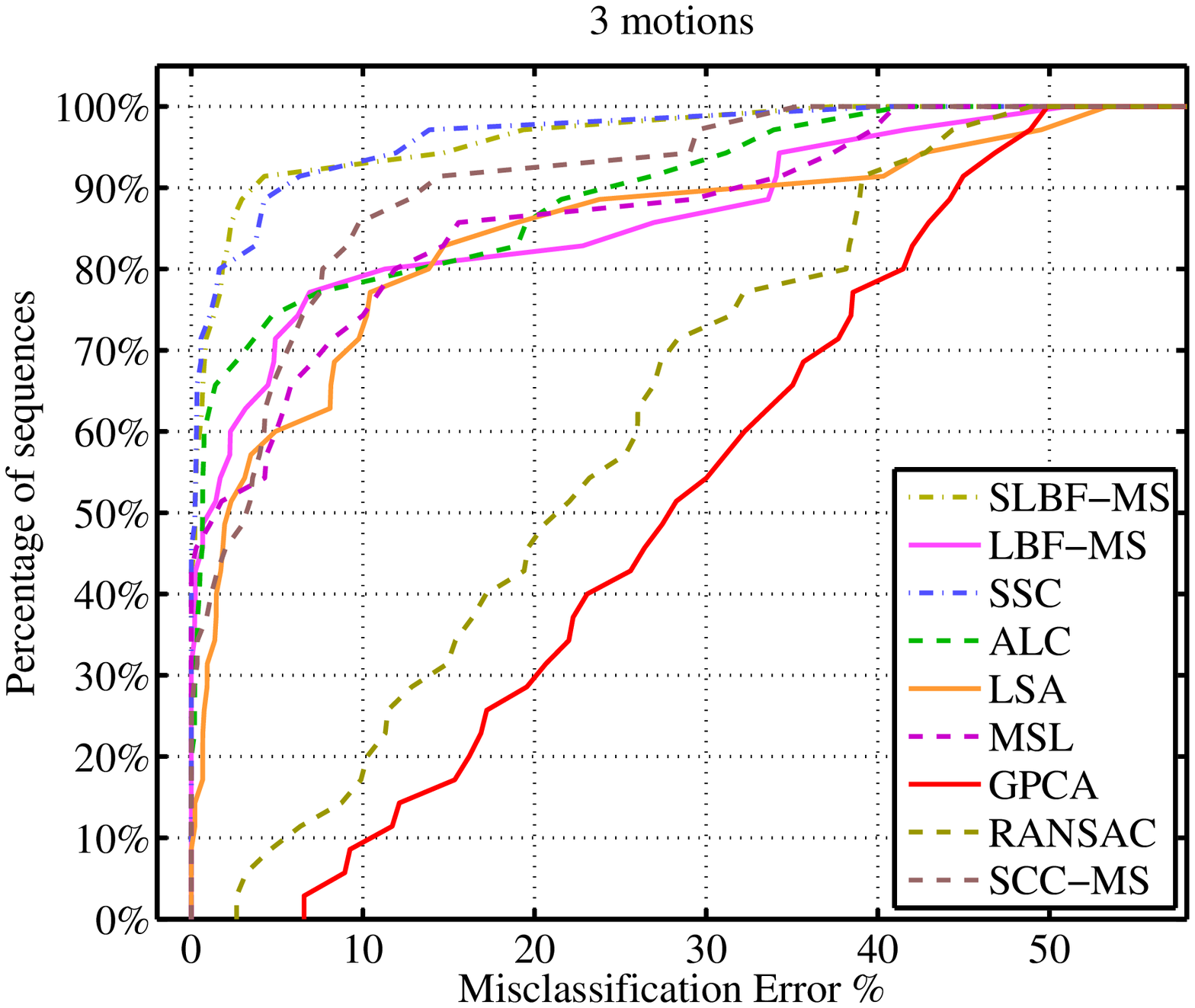}
\caption{\it The misclassification rate of some algorithms for the Hopkins 155 database. The $y$-axis represent the percentage of data sets that have misclassification rates (under corresponding algorithms) lower than that of $x$-axis. \label{fig:hop}}
\end{center}
\end{figure}
%\begin{figure}
%\begin{center}
%
%\caption{\it Data set \#3 from Section~\ref{sec:ikf}.  The color
%value represents the number of neighbors chosen at that point. Note
%that the algorithm chooses smaller neighborhoods for points closer
%to the intersection of the planes. \label{fig:hop2}}
%\end{center}
%\end{figure}

\begin{figure}
\begin{center}
\includegraphics[width=.13\textwidth,height=.1\textwidth]{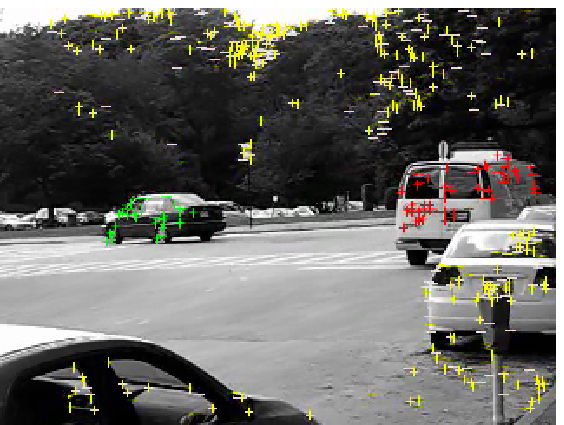}
\includegraphics[width=.13\textwidth,height=.1\textwidth]{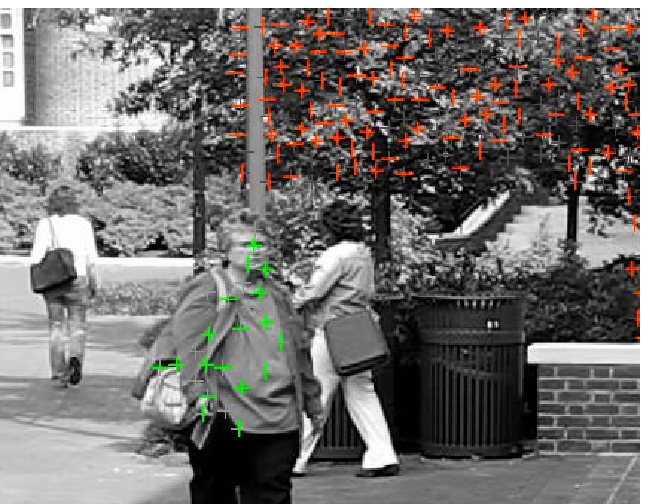}
\includegraphics[width=.13\textwidth,height=.1\textwidth]{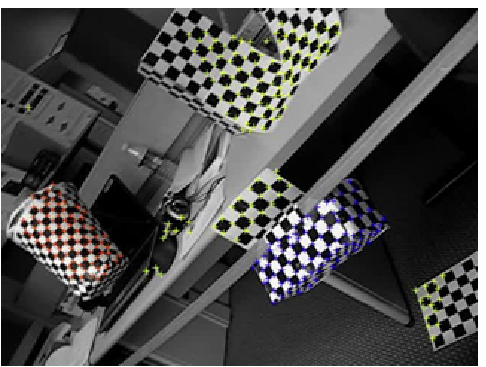}\\
\includegraphics[width=.13\textwidth,height=.1\textwidth]{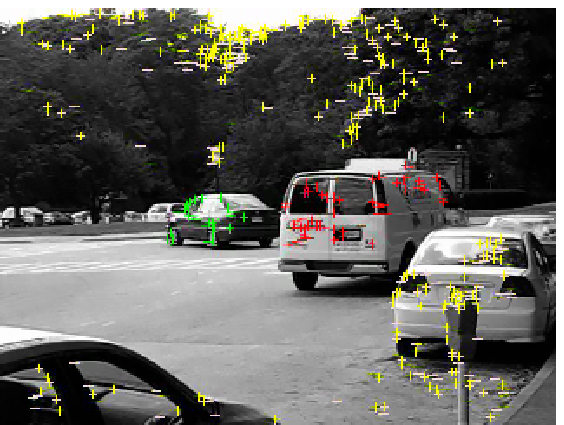}
\includegraphics[width=.13\textwidth,height=.1\textwidth]{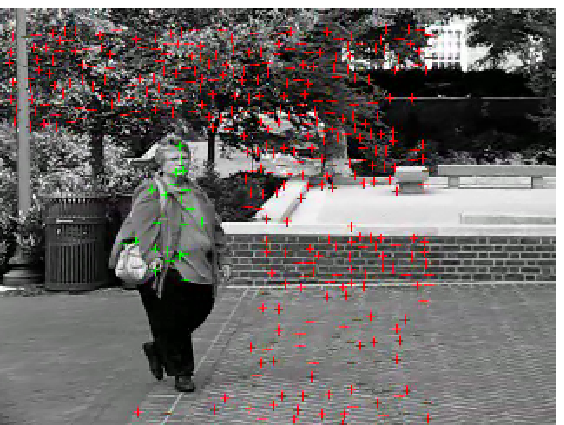}
\includegraphics[width=.13\textwidth,height=.1\textwidth]{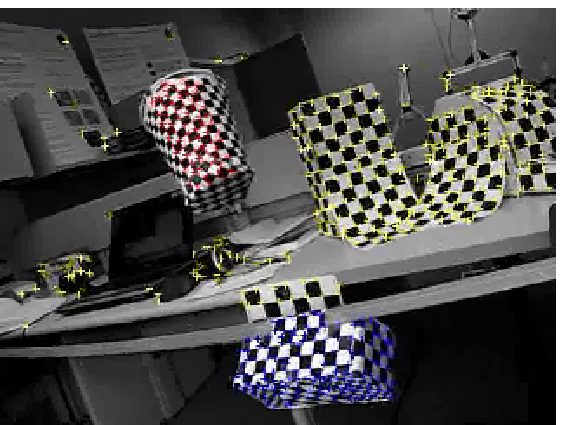}
\caption{\it Frames of the traffic, articulated and checker (from left to right) videos in Hopkins 155 database. \label{fig:hopkins}}
\end{center}
\end{figure}

We test the proposed algorithms on the Hopkins 155 database of motion
segmentation, which is available at
http://www.vis\\ion.jhu.edu/data/hopkins155. This data set contains 155
video sequences along with the coordinates of certain features
extracted and tracked for each sequence in all its frames. The main
task is to cluster the feature vectors (across all frames) according
to the different moving objects and background in each video. It consists of three types of videos: checker, traffic and articulated (see Figure~\ref{fig:hopkins} for demonstration of frames of such videos).

More formally, for a given video sequence, we denote the number of
frames by $F$. In each sequence, we have either one or two
independently moving objects, and the background can also move due
to the motion of the camera. We let $K$ be the number of moving
objects plus the background, so that $K$ is 2 or 3 (and distinguish
accordingly between two-motions and three-motions). For each
sequence, there are also $N$ feature points
$\by_1,\by_2,\cdots,\by_N \in \mathbb{R}^3$ that are detected on the
objects and the background. Let $\bz_{ij}\in \mathbb{R}^2$ be the
coordinates of the feature point $\by_j$ in the $i^{\rm th}$ image frame
for every $1\leq i\leq F$ and $1\leq j \leq N$. Then
$\bz_j=[\bz_{1j},\bz_{2j},\cdots,\bz_{Fj}] \in \mathbb{R}^{2F}$ is
the trajectory of the $j^{th}$ feature point across the $F$ frames.
The actual task of motion segmentation is to separate these
trajectory vectors $\bz_1,\bz_2,\cdots,\bz_N$ into $K$ clusters
representing the $K$ underlying motions.

It has been shown~\cite{Costeira98} that under the affine camera model,
 the trajectory vectors corresponding
to different moving objects and the background across the $F$ image
frames live in distinct affine subspaces of dimension at most three
in $\mathbb{R}^{2F}$. Following this theory, we implement our
algorithm with $d=3$.

%The methods that we compare with are the Mixtures
%of PPCA algorithm (MoPPCA) (\cite{Tipping99mixtures}
%1999), the $K$-Subspaces algorithm (KS) (\cite{Ho03}), the Local Subspace Analysis (LSA) (\cite{Yan06LSA}), the Spectral Curvature Clustering (SCC)(\cite{spectral_applied}) and the GPCA algorithm with voting (GPCA) (Ma et al.
%2008).
\begin{table*}[htbp]
\centering \caption{\small {Average total computation times for all 155 sequences.}}\label{tab:Hopkins_time}
\begin{tabular}{|c|c|c|c|c|c|c|}
  \hline
RANSAC&LBF-MS (4$K$,3)&LBF (4$K$,3)&  SCC-MS(4$K$,3)& SLBF-MS(2$F$,3) &SLBF(2$F$,3)&SSC-N(4$K$,3) \\
\hline
60s & 73s  & 91s&196s&28min &31min &427min \\
  \hline
\end{tabular}
\end{table*}
%
%\setlength{\abovecaptionskip}{4pt}   % 0.5cm as an example
%\setlength{\belowcaptionskip}{4pt}
%\begin{table*}[htbp]
%\centering \caption{\small {The percentage of incorrectness ($e\%$)
%and the average computation time $t$ of the three methods SOD (LBF),
%ALC and GPCA.}}\label{tab:SOD} {\scriptsize
%\begin{tabular}{|c|c|@{}c@{}|@{}c@{}|@{}c@{}|@{}c@{}|@{}c@{}|@{}c@{}|@{}c@{}|@{}c@{}|@{}c@{}|@{}c@{}|@{}c@{}|}
%\hline
%
%&&\multicolumn{7}{c|}{no minimum angle}&\multicolumn{4}{c|}{minimum
%angle
%$=\pi/8$} \\
%\hline
%
%&&$1^6\in\mathbb{R}^5$ &$2^4\in\mathbb{R}^5$ &$3^3\in\mathbb{R}^5$
%&$10^2\in\mathbb{R}^{15}$ &$1^6\in\mathbb{R}^3$
%&$2^4\in\mathbb{R}^3$ &$3^3\in\mathbb{R}^4$ &$1^6\in\mathbb{R}^3$
%&$2^4\in\mathbb{R}^3$ &$3^3\in\mathbb{R}^4$
%&$10^2\in\mathbb{R}^{15}$ \\
%
%
%\hline
%
%&e\%&17&3&2&0&55&29&19&3&5&5&0\\
%
%\raisebox{1.5ex}[0pt]{SOD (LBF)}&t&3.51&4.07&3.37&7.31&3.13&3.77&3.85&3.09&3.45&3.32&6.78\\
%\hline
%
%ALC&e\%&1&0&0&16&34&31&1&0&10&1&13\\
%
%$\epsilon=0.05$&t&23.74&43.44&59.14&1370.92&20.49&37.49&53.59&20.22&37.41&54.11&1354.11\\
%\hline
%
%&e\%&88&100&100&100&27&100&100&13&100&100&100\\
%
%\raisebox{1.5ex}[0pt]{GPCA}&t&0.03&0.09&0.12&1.30&0.06&0.09&0.12&0.04&0.09&0.12&1.30\\
%
%\hline
%\end{tabular}
%}
%\end{table*}
%

We compare our algorithm with the following ones:
%Connected Component
%Search (CCS)~\cite{LLMC},
improved GPCA for motion segmentation (GPCA)~\cite{gpca_motion_08},
$K$-flats (KF)~\cite{Ho03} (implemented for linear subspaces), Local
Linear Manifold Clustering (LLMC)~\cite{LLMC}, Local Subspace
Analysis (LSA)~\cite{Yan06LSA},
%Linear Spectral Curvature Clustering
%(LSCC)~\cite{spectral_applied},
Multi Stage Learning (MSL)~\cite{Sugaya04},  Spectral Curvature Clustering
(SCC)~\cite{spectral_applied} and SCC-MS (see description earlier), Sparse Subspace Clustering (SSC)~\cite{ssc09}, and RANSAC for HLM~\cite{Yang06Robust}.

For GPCA (improved for motion segmentation), LLMC, LSA, MSL and RANSAC (for HLM), we copy the results from
http://www.vision.jhu.edu/data/hopkins155 (they are \\based on
experiments reported in~\cite{Tron07abenchmark} and~\cite{LLMC}). We perform our own experiments for SCC, SCC-MS, SSC-N (SSC-B is not reported since it did not perform as well as SSC-N), LBF, LBF-MS, SLBF, SLBF-MS, we perform the experiments on our own and
record the mean misclassification rate and the median
misclassification rate for each algorithm for any fixed $K$ (two or
three-motions) and for the different type of motions (``checker",
``traffic" and ``articulated").
%\subsection{Discussion of Results}
Each experiment (testing the latter set of algorithms) was repeated 500 times. The average misclassification rates, standard deviation and running time are recorded in Tables \ref{tab:face_seg} and \ref{tab:face_std} and demonstrated in Figure~\ref{fig:hop}.

Our misclassification rates for SCC are different than \cite{SCC_hopkins} and~\cite{lauer09_iccv} and our misclassification rates
for SSC are different than~\cite{ssc09} (the difference between our and their results differ more than twice the standard deviations of misclassification rates obtained here). This can be explained by possible evolutions of the codes since then (at least for SSC).
We remark though that the misclassification rates of SCC-MS here are even slightly better than the misclassification rates of SCC in~\cite{SCC_hopkins}.

From Table \ref{tab:comp1} and Figure~\ref{fig:hop} we can see that our
algorithms work well for the Hopkins database. Of all the methods tested,
SLBF-MS and SSC-N are the most accurate algorithms. Besides SLBF/SLBF-MS and SSC-N, only SCC-MS is better than LBF-MS. However, From Table \ref{tab:Hopkins_time}, LBF-MS ran more than 100 times faster than SSC-N and SLBF-MS is also more than 10 times faster than SSC. In many of the
cases, the $\ell_1$ energy (as
well as the $\ell_2$ energy) was lower for the labels obtained by
LBF than the true labels. We thus suspect that the reason SLBF/SLBF-MS works better than LBF/LBF-MS is that good
clustering of the Hopkins data requires additional type of
information (e.g., spectral information) to be combined with
subspace clustering (i.e., hybrid linear modeling).

By adapting the parameters of SLBF-MS (or alternatively, SLBF, LBF, LBF-MS), we can further improve its misclassification rates on Hopkins 155 (e.g., total $0.81\%$ for two-motions and $2.12\%$ for three-motions by SLBF-MS).
However, we have fixed in advance all parameters and insisted using the same parameters on
all four kinds of data (see the third paragraph of Section~\ref{sec:exp}).

From Table \ref{tab:std} we can see that SLBF-MS, SLBF and SSC-N have negligible randomness. Indeed, their only randomness come from the $K$-means step, but this randomness is effectively reduced because of  the restarting strategy. LBF and LBF-MS are more random, but still have comparable standard deviations with other good algorithms on Hopkins 155 database such as SCC/SCC-MS.
%\section{Segmentation}
\subsection{Clustering results on the extended Yale face database B} \label{sec:face_seg}
We test LBF, LBF-MS, SLBF and SLBF-MS and compare them with ALC, $K$-flats, and SSC on the extended Yale face database B~\cite{KCLee05}, which is available on {http://vision.ucsd.edu/\\~leekc/ExtYaleDatabase/ExtYaleB.html}.  We will see that this data set shows a failure mode of our algorithms; and we will show how we can engineer a workaround.

We use subsets of the extended Yale face database B consisting of face images of $K=2,3,\cdots, 10$ persons under 64 varying lighting conditions. Our objective is to cluster these images according to the persons. In implementation, for any fixed $K$ we repeat each algorithm on 100 randomly chosen subsets of $K$ persons.  The HLM model is applicable to this database, because the images of a face under variable lighting lies in a three-dimensional
linear subspace if shadow is not considered~\cite{KCLee05}, and a nine-dimensional subspace with shadow considered~\cite{Basri03}. In our experiments, we found that the images of a person in this database lie roughly in a 5-dimensional subspace, and therefore we first reduce the dimension of the data to $5K$ (recall that $K$ is the number of persons).
We do not include the GPCA algorithm since it is slow and does not work well on this database. We also do not include SCC and RANSAC since the code provided returns errors for some examples. The setting of ALC (voting with $K$) follows~\cite[sec.~(2.2)]{ALC_motion} exactly: it chooses $\eps$ from 101 values in the range $10^{-5}-10^3$ (see the code in http://perception.csl.uiuc.edu/co\\ding/motion/\#Software).

We can see from  the first row of Table~\ref{tab:face_seg} that LBF does a poor job discriminating the linear clusters in this data set.
%This is because the majority of the energy in this data (in the sense of the singular values of the data matrix) is concentrated in its first few principal components, which contain very little discriminative information about the faces.
The failure occurs because of a combination of two factors: the first is the relatively sparse sampling of the data, with only 64 points per
$5$-dimensional cluster, and the second, the relative nearness of the underlying subspaces to each other.
In particular, almost any neighborhood of any given point (even very small neighborhood) has points from the other affine clusters and consequently there is no ``optimal'' scale.     For example, in the 128 face images from persons 1 and 2, more than a fifth of the points are closer to the subspace spanned by the first 5 principal components of the points in the other cluster than to their second nearest neighbors, and more than two thirds of the points are closer to the other subspace than to their 4th nearest neighbors.
%We illustrate this in an artificial example, shown in Figure~\ref{fig:example_yaleface}.
In some sense, this {\it is} a single $5$-dimensional set, rather than two $5$-dimensional sets.   For example, the average distance of a point to
the $5$-dimensional best fit subspace by the points in the same cluster is $2.7\times 10^3$, and the average distance to the $5$-dimensional best fit subspace of the whole data set is $3.3\times 10^3$, whereas the average norm of a point in the data set is $1.1\times 10^4$.  Thus one loses little in terms of relative fitting error by considering the set as spanned by a single subspace.

However, most points are actually closer to the subspace spanned by the same face than to the subspace spanned by the other face, if only by a little, and a global method such as SSC is still able to find and discriminate between the two affine clusters.
The problem of data having large variance in directions irrelevant to a classification task is not unusual.  A standard method of dealing with this situation is to ``whiten'' the data; i.e. reduce the value of the large singular values.  A very crude whitening is obtained by simply removing the first few principal components.
If we exclude the first two principal components after reducing the dimension to $5K$ for LBF/SLBF algorithms,
% to improve the separation between clusters and avoid the failure of the nearest neighborhood search strategy in Algorithm~\ref{alg:nhbd}.  Without this preprocessing, the neighborhood of a image usually contains images of a different person, as shown in the left figure of Figure~\ref{fig:yaleface0}, where $36\%$ of the $4$-NN neighborhoods are images from the other person. The reason is that the two clusters have close top principal components: the angle between the first components is $0.31$ (i.e., about 18 degree), and the angle between the second components is $0.39$ (i.e., about 23 degree). An clearer artificial example is shown in Figure~\ref{fig:example_yaleface}: with close principal components, the nearest neighborhood includes points from the other cluster. We also note that we only use this preprocessing to LBF/SLBF since it does not improve the performance of other algorithms.
we see in Table~\ref{tab:face_seg} that the results are greatly improved and become competitve\footnote{Removing principal components harms the performance of the other algorithms.}.  With more sophisticated whitening, the results can be further improved.

\setlength{\abovecaptionskip}{4pt}   % 0.5cm as an example
\setlength{\belowcaptionskip}{4pt}
\begin{table*}[htbp]
\centering \caption{\small {Mean percentage of misclassified points and mean running time on clustering the extended Yale
face database B. }}\label{tab:face_seg}
\scriptsize

\begin{tabular}{|c|c||c|c|c|c|c|c|c|c|c||}
\hline

\multicolumn{2}{|c||}{$K$} & 2&3&4&5&6&7&8&9&10 \\ \hline \hline
\rowcolor[gray]{.8}& $e\%$ & 32.49 & 54.42 & 57.45 & 56.00 & 56.24 & 56.94 & 59.53 & 59.66 & 60.74\\
 \rowcolor[gray]{.8}\raisebox{1.5ex}[0pt]{LBF(without whitening)}& $t(s)$ & 0.24 & 0.48 & 0.82 & 1.26 & 1.93 & 2.97 & 4.18 & 5.81 & 8.05 \\ \hline

& $e\%$   & 18.27 & 36.22 & 48.24 & 50.18 & 49.99 & 50.68 & 53.08 & 54.06 & 54.73 \\
\raisebox{1.5ex}[0pt]{LBF-MS(without whitening)}& $t(s)$  & 0.12 & 0.21 & 0.36 & 0.57 & 0.89 & 1.41 & 2.06 & 2.98 & 4.13  \\ \hline

% \rowcolor[gray]{.8}& $e\%$ & 13.49 &14.89 &19.09 &24.14 &31.74 &34.77 &36.82 &38.34 &40.28\\
% \rowcolor[gray]{.8}\raisebox{1.5ex}[0pt]{LBF(M1)}& $t(s)$ & 0.27 & 0.55 & 0.95 & 1.48 & 2.20 & 3.45 & 4.75 & 6.62 & 8.81  \\ \hline
%
%& $e\%$  & 7.82 & 10.28 & 11.84 & 16.23 & 22.50 & 24.05 & 27.17 & 31.57 & 32.63 \\
%\raisebox{1.5ex}[0pt]{LBF-MS(M1)}& $t(s)$ & 0.12 & 0.22 & 0.37 & 0.58 & 0.89 & 1.41 & 2.05 & 2.94 & 3.99  \\ \hline

\rowcolor[gray]{.8}& $e\%$ & 7.94 & 8.33 & 12.89 & 17.83 & 27.40 & 31.89 & 35.04 & 38.53 & 38.95\\
 \rowcolor[gray]{.8}\raisebox{1.5ex}[0pt]{LBF}& $t(s)$  & 0.24 & 0.50 & 0.87 & 1.38 & 2.09 & 3.28 & 4.58 & 6.38 & 8.57  \\ \hline

& $e\%$  & 8.40 & 9.51 & 12.18 & 15.57 & 19.18 & 22.88 & 27.20 & 30.39 & 33.17 \\
\raisebox{1.5ex}[0pt]{LBF-MS}& $t(s)$  & \textbf{ 0.12 } & \textbf{ 0.22 } & \textbf{ 0.37 } & \textbf{ 0.58 } & \textbf{ 0.89 } & \textbf{ 1.41 } & \textbf{ 2.07 } & \textbf{ 2.94 } & \textbf{ 4.02}  \\ \hline

% \rowcolor[gray]{.8}& $e\%$  & 12.46 & 25.43 & 31.98 & 35.48 & 40.20 & 41.41 & 43.48 & 43.67 & 46.34 \\
% \rowcolor[gray]{.8}  \raisebox{1.5ex}[0pt]{SLBF(M1)}& $t(s)$  & 4.57 & 12.42 & 24.84 & 44.86 & 72.86 & 109.64 & 158.76 & 210.74 & 284.36  \\ \hline
%
%& $e\%$ & 9.48 & 13.86 & 17.06 & 21.29 & 24.34 & 30.10 & 31.17 & 33.12 & 35.50\\
%\raisebox{1.5ex}[0pt]{SLBF-MS(M1)}& $t(s)$  & 4.28 & 12.28 & 24.61 & 42.39 & 65.92 & 97.01 & 143.76 & 190.42 & 261.77 \\ \hline

\rowcolor[gray]{.8}& $e\%$   & 11.12 & 14.78 & 20.42 & 26.52 & 32.96 & 36.91 & 40.49 & 42.99  & 46.63    \\
 \rowcolor[gray]{.8}  \raisebox{1.5ex}[0pt]{SLBF}& $t(s)$  & 4.17 & 12.72 & 25.70 & 44.89 & 72.99 & 111.58 & 165.47 & 226.56 & 310.30  \\ \hline

& $e\%$ & 9.12 & 12.48 & 18.61 & 25.27 & 30.50 & 33.97 & 36.22 & 38.66 & 41.44 \\
\raisebox{1.5ex}[0pt]{SLBF-MS}& $t(s)$  & 3.84 & 12.20 & 23.88 & 41.24 & 64.10 & 95.73 & 142.09 & 192.34 & 262.40  \\ \hline

 \rowcolor[gray]{.8}ALC & $e\%$  & \textbf{3.46} & \textbf{6.08} & 14.59 & 29.59 & 67.06 &69.04 & 76.00 & 73.94 & 77.16\\
 \rowcolor[gray]{.8}(voting with $K$) & $t(s)$  & 42.99 & 122.29 & 258.20 & 451.07 & 699.52 & 1090.96 & 1625.10 & 2384.69 & 3343.93 \\ \hline

ALC & $e\%$  & 10.43 & 15.23 & 32.20 & 42.15  & 58.10 & 62.54 & 70.84 & 81.14 & 84.25\\
($\epsilon$ from LBF) & $t(s)$   & 0.95 & 2.49 & 5.54 & 11.54 & 24.38 & 45.27 & 78.05 & 132.35
& 211.15 \\ \hline

%\rowcolor[gray]{.8}ALC(9$K$) & $e\%$   & 5.35 & 18.77 & 64.29 & 58.59  & 51.62 \\
% \rowcolor[gray]{.8}(voting with $K$) & $t(s)$  & 55.51 & 149.31 & 300.39 & 543.04 & 920.03 \\ \hline
%
%ALC(9$K$) & $e\%$   & 8.18 & 11.52 & 42.88 & 59.42 & 76.38  \\
%($\epsilon$ from LBF) & $t(s)$   & 1.26 & 3.75 & 10.52 & 25.76 & 58.28  \\ \hline

\rowcolor[gray]{.8}& $e\%$ & 5.39 & 11.82 & 29.39 & 41.96 & 49.56 & 54.51 & 55.49 & 57.24 & 58.94 \\
\rowcolor[gray]{.8}\raisebox{1.5ex}[0pt]{SCC}&$t(s)$  & 1.62 & 3.85 & 9.52 & 15.37 & 22.71 & 32.45 & 54.54 & 56.91 & 75.92 \\ \hline

& $e\%$  & 4.51 & 15.05 & 36.00 & 51.68 & 59.66 & 64.15 & 68.71 & 71.18 & 74.01\\
\raisebox{1.5ex}[0pt]{SCC-MS} & $t(s)$  & 1.62 & 4.20 & 9.28 & 14.49 & 22.08 & 31.71 & 54.21 & 56.99 & 73.10 \\ \hline

%\rowcolor[gray]{.8}& $e\%$ & 6.48 & 20.89 & 39.10 & 51.13 & 54.19 & 57.70 & 58.93 & 60.59 & 60.78 \\
%\rowcolor[gray]{.8}\raisebox{1.5ex}[0pt]{SCC(9$K$)}&$t(s)$   & 1.69 & 4.31 & 8.97 & 15.13 & 23.21 & 34.90 & 56.09 & 57.13 & 77.22 \\ \hline
%
%& $e\%$ & 9.75 & 27.34 & 47.11 & 56.06 & 62.36 & 66.32& 69.51 & 72.76 & 75.82 \\
%\raisebox{1.5ex}[0pt]{SCC-MS(9$K$)} & $t(s)$  & 1.78 & 4.18 & 8.95 & 14.64 & 22.57 & 33.58 & 56.45 & 58.05 & 79.27 \\ \hline

\rowcolor[gray]{.8}& $e\%$ & 6.45 & 8.10 & \textbf{ 10.04 } & \textbf{ 10.32 } & \textbf{ 11.02 } & \textbf{11.85 } & \textbf{ 12.47 } & \textbf{ 13.41 } & \textbf{ 15.44 }\\
\rowcolor[gray]{.8}\raisebox{1.5ex}[0pt]{SSC}&$t(s)$ & 28.36 & 46.45 & 67.11 & 92.75 & 128.46 & 182.65 & 259.66 & 340.12 & 612.21 \\ \hline

& $e\%$  & 7.20 & 12.12 & 19.06 & 26.77 & 32.59 & 35.18 & 38.58 & 42.00 & 44.40\\
\raisebox{1.5ex}[0pt]{K-flats} & $t(s)$  & 0.16 & 0.37 & 0.76 & 1.29 & 2.14 & 3.25 & 5.18 & 6.91 & 9.60 \\ \hline

%\rowcolor[gray]{.8}& $e\%$  & 4.17 & 4.38 & 6.84 & 7.72 & 8.45 & 10.05 & 10.62 & 12.39 & 14.51 \\
%\rowcolor[gray]{.8}\raisebox{1.5ex}[0pt]{SSC(9$K$)}&$t(s)$ & 31.74 & 62.43 & 95.53 & 136.46 & 188.54 & 262.36 & 363.41 & 490.76 & 833.17 \\ \hline

%& $e\%$  & 6.65 & 14.90 & 22.09 & 30.53 & 34.03 & 39.01 & 40.95 & 43.02 & 45.78 \\
%\raisebox{1.5ex}[0pt]{K-flats(9$K$)} & $t(s)$  & 0.23 & 0.64 & 1.45 & 2.73 & 4.76 & 7.40 & 11.20 & 13.98 & 18.26 \\ \hline
%

%\rowcolor[gray]{.8}& $e\%$ & 0.0 & 49.5 & 0.0 & 26.6 & 9.9 & 25.2 & 28.5 & 30.6 & 19.8 \\
%\rowcolor[gray]{.8}\raisebox{1.5ex}[0pt]{GPCA}&$t(s)$& 1.42 & 2.72 & 4.91 & 8.08 & 11.71 & 33.11 & 99.49 & 286.36 & 1122.50 \\ \hline

\end{tabular}
\end{table*}

\setlength{\abovecaptionskip}{4pt}   % 0.5cm as an example
\setlength{\belowcaptionskip}{4pt}
\begin{table*}[htbp]
\centering \caption{\small {The standard deviation($\%$) to the mean percentage of misclassified points on the extended Yale
face database B.}}\label{tab:face_std}
\scriptsize

\begin{tabular}{|c||c|c|c|c|c|c|c|c|c||}
\hline

{Real $K$} & 2&3&4&5&6&7&8&9&10 \\ \hline \hline

\rowcolor[gray]{.8}{LBF(without whitening)}  & 20.46 & 14.73 & 4.87 & 3.89 & 5.54 & 4.99 & 4.63 & 3.95 & 3.22 \\ \hline

LBF-MS(without whitening)    & 18.23 & 18.77 & 13.40 & 6.74 & 4.52 & 5.51 & 5.31 & 4.90 & 4.14 \\ \hline

\rowcolor[gray]{.8}{LBF}  & 5.27 & 3.73 & 7.97 & 9.86 & 11.21 & 10.38 & 8.27 & 6.52 & 6.20 \\ \hline

LBF-MS   & 4.25 & \textbf{3.08} & 5.33 & 6.24 & 7.73 & 8.02 & 8.29 & 8.05 & 7.25 \\ \hline

 \rowcolor[gray]{.8}SLBF  & 4.76 & 5.37 & 5.08 & 5.25 & 5.48 & 5.42 & 4.57 & 4.74 & 3.79 \\ \hline

 SLBF-MS   & 4.77 & 5.37 & 5.84 & 4.91 & 3.75 & 3.76 & \textbf{2.87} & \textbf{3.01} & \textbf{3.22}  \\ \hline

 \rowcolor[gray]{.8}ALC(voting with $K$) & \textbf{2.21} & 6.93 & 13.87 & 14.89 & 16.84& 24.71 & 18.05 & 21.62 & 16.98 \\ \hline

ALC($\epsilon$ from LBF)& 13.14 & 12.96 & 14.91 & 16.40 & 15.22& 12.22 & 10.89 & 6.76 & 6.10  \\ \hline

 \rowcolor[gray]{.8}SCC  & 5.21 & 11.71 & 14.65 & 10.60 & 6.68 & 5.14 & 4.67 & 4.32 & 5.03 \\ \hline

SCC-MS & 2.84 & 13.66 & 14.66 & 10.41 & 8.29 & 6.72& 5.61 & 5.93 & 5.46 \\ \hline

\rowcolor[gray]{.8}SSC  & 4.57 & 3.76 & \textbf{4.52} & \textbf{3.82} & \textbf{3.59} & \textbf{2.87} & 3.18 & 3.45 & 5.21\\ \hline

K-flats & 4.67 & 6.86 & 8.53 & 8.89 & 7.29 & 6.41 & 6.67 & 4.84 & 5.43\\ \hline

\end{tabular}

\end{table*}

\subsection{Clustering results on MNIST data set} \label{sec:MNIST_seg}
Finally, we work on the MNIST data set (available at {http://ya\\nn.lecun.com/exdb/mnist/}).
This data set consists of several thousand $28\times 28$ images of the digits $0$ through $9$.  We work on some subsets of the data which contain 2 or 3 digits and choose 200 images for each digit at random. We apply PCA to reduce the dimension to $D=5$ for GPCA and to both $D=10$ and $D=50$ for the rest of algorithms. The choice of both $D=10$ and $D=50$ provide richer testing opportunities, this is however unavailable for GPCA, which cannot handle $D = 50$ and often get stuck with $D=10$. We process the data the same way as in Section~\ref{sec:face_seg}. We run each experiment 500 times, using $d=3$ and the correct number of clusters,  and record the misclassification rates, the standard deviation and running time in Tables \ref{tab:MNIST_seg},  \ref{tab:MNIST_std}, \ref{tab:MNIST_seg2} and  \ref{tab:MNIST_std2}.

\setlength{\abovecaptionskip}{4pt}   % 0.5cm as an example
\setlength{\belowcaptionskip}{4pt}
\begin{table*}[htbp]
\centering \caption{\small {Mean percentage of misclassified points and mean running time on clustering MNIST data set (D=5 for GPCA, D=10 for other algorithms). }}\label{tab:MNIST_seg}
\scriptsize

\begin{tabular}{|c|c||c|c|c|c|c|c|c||}
\hline
\multicolumn{2}{|c||}{subsets} & [1 2] & [1 3] & [1 7] & [4 7] & [2 4 8] & [3 6 8] & [1 2 3] \\ \hline
\multicolumn{2}{|c||}{$K$} & 2&2&2&2&3&3&3 \\ \hline \hline

 \rowcolor[gray]{.8}& $e\%$ & 8.0 & 8.5 & 12.9 & 25.5 & 28.8 & 28.1 & 20.2 \\
 \rowcolor[gray]{.8}\raisebox{1.5ex}[0pt]{LBF}& $t(s)$ & 0.4 & 0.4 & 0.3 & 0.4 & 0.7 & 0.7 & 0.7   \\ \hline

& $e\%$ & 9.7 & 7.8 & 8.8 & 24.0 & 40.2 & 33.5 & 21.5 \\
\raisebox{1.5ex}[0pt]{LBF-MS}& $t(s)$ & \textbf{0.2} &\textbf{0.2} & \textbf{0.2} & \textbf{0.2} & \textbf{0.5} & \textbf{0.4} & \textbf{0.4}   \\ \hline

 \rowcolor[gray]{.8}& $e\%$ & 0.5 & \textbf{1.0} & \textbf{2.0} & \textbf{3.0} & \textbf{3.8} & \textbf{19.7} & \textbf{17.3} \\
 \rowcolor[gray]{.8}  \raisebox{1.5ex}[0pt]{SLBF}& $t(s)$ & 13.9 & 13.7 & 13.5 & 14.5 & 41.9 & 41.0 & 42.7  \\ \hline

& $e\%$ & 0.5 & \textbf{1.0} & \textbf{2.0} & \textbf{3.0} & \textbf{3.8} & 19.7 & 17.3  \\
\raisebox{1.5ex}[0pt]{SLBF-MS}& $t(s)$ & 12.8 & 13.7 & 13.0 & 14.6 & 38.6 & 46.3 & 39.0  \\ \hline

 \rowcolor[gray]{.8}ALC & $e\%$ & \textbf{0.2} &   2.2 &   3.5 &  48.5 &   4.2 &  42.7&   45.3 \\
 \rowcolor[gray]{.8}(voting with $K$) & $t(s)$ & 830.5 & 823.3 & 813.3 & 753.2 & 1789.5 & 1871.8 & 1987.7  \\ \hline

ALC & $e\%$ &  20.3&   32.0&   51.8&   27.5&    4.0&   30.3&   14.5 \\
($\epsilon$ from LBF) & $t(s)$ & 23.2&        22.5&        21.6&        23.0   &     55.6     &   54.7        &54.0  \\ \hline

\rowcolor[gray]{.8}& $e\%$ & 7.0 & 6.4 & 11.4 & 23.4 & 22.8 & 26.7 & 39.2 \\
\rowcolor[gray]{.8}\raisebox{1.5ex}[0pt]{SCC}&$t(s)$& 1.2 & 1.5 & 1.4 & 1.3 & 2.5 & 2.7 & 2.3 \\ \hline

& $e\%$ & 6.3 & 7.9 & 10.5 & 23.2 & 23.3 & 26.9 & 32.8 \\
\raisebox{1.5ex}[0pt]{SCC-MS} & $t(s)$ & 0.9 & 0.8 & 1.1 & 1.0 & 1.9 & 1.9 & 1.5 \\ \hline

\rowcolor[gray]{.8}& $e\%$ & 22.3 & 30.8 & 32.5 & 47.0 & 48.2 & 33.8 & 31.0 \\
\rowcolor[gray]{.8}\raisebox{1.5ex}[0pt]{GPCA}&$t(s)$& 8.7 & 9.2 & 9.4 & 10.8 & 24.9 & 24.5 & 22.5 \\ \hline

& $e\%$ & 11.1 & 6.8 & 6.3 & 29.1 & 43.9 & 40.7 & 29.2 \\
\raisebox{1.5ex}[0pt]{K-flats} & $t(s)$ & 0.4 & 0.4 & 0.4 & 0.4 & 0.9 & 0.8 & 0.6  \\ \hline

%\rowcolor[gray]{.8}& $e\%$ &   5.3 &   3.7  &  9.8 &  19.2 &  41.8 &  25.2 &  50.0 \\
%\rowcolor[gray]{.8}\raisebox{1.5ex}[0pt]{SSC1}&$t(s)$&    220.55 &      196.64 &      200.75 &      203.23 &      322.62  &     333.03 &      338.21\\ \hline

\rowcolor[gray]{.8}& $e\%$ &  4.5 &   3.5 &   9.0  & 21.0  & 19.5 &  24.5  & 49.3 \\
\rowcolor[gray]{.8}\raisebox{1.5ex}[0pt]{SSC} & $t(s)$ & 220.6 &      196.6 &      200.8 &      203.2 &      322.6  &     333.0 &      338.2\\ \hline

%\rowcolor[gray]{.8}& $e\%$ &  4.5 &  3.7 &   8.8 &  11.7  & 18.5  & 23.0 &  47.5 \\
%\rowcolor[gray]{.8}\raisebox{1.5ex}[0pt]{SSC3}&$t(s)$& 220.55 &      196.64 &      200.75 &      203.23 &      322.62  &     333.03 &      338.2\\ \hline
\end{tabular}

\end{table*}

\begin{table*}[htbp]
\centering \caption{\small {Mean percentage of misclassified points and mean running time on clustering MNIST data set (D=50). }}\label{tab:MNIST_seg2}
\scriptsize

\begin{tabular}{|c|c||c|c|c|c|c|c|c||}
\hline
\multicolumn{2}{|c||}{subsets} & [1 2] & [1 3] & [1 7] & [4 7] & [2 4 8] & [3 6 8] & [1 2 3] \\ \hline
\multicolumn{2}{|c||}{$K$} & 2&2&2&2&3&3&3 \\ \hline \hline

 \rowcolor[gray]{.8}& $e\%$ & 20.5 & 13.1 & 18.2 & 30.2 & 26.3 & 24.1 & \textbf{19.2} \\
 \rowcolor[gray]{.8}\raisebox{1.5ex}[0pt]{LBF}& $t(s)$ & 2.8 & 2.8 & 2.6 & 3.1 & 5.2 & 5.1 & 4.7   \\ \hline

& $e\%$ & 12.5 & 16.9 & 10.7 & 19.1 & 23.5 & 27.3 & 24.3 \\
\raisebox{1.5ex}[0pt]{LBF-MS}& $t(s)$ & 1.3 & 1.3 & 1.3 & 1.3 & 2.3 & 2.3 & 2.3   \\ \hline

 \rowcolor[gray]{.8}& $e\%$ & 8.3 & 4.3 & \textbf{2.3} & 13.8 & 4.3 & \textbf{17.5} & 21.7 \\
 \rowcolor[gray]{.8}  \raisebox{1.5ex}[0pt]{SLBF}& $t(s)$ & 15.1 & 15.0 & 14.6 & 16.8 & 37.5 & 38.5 & 39.5  \\ \hline

& $e\%$ & 5.5 & \textbf{3.3} & 5.0 & \textbf{5.5} & \textbf{3.2} & 18.5 & 21.7  \\
\raisebox{1.5ex}[0pt]{SLBF-MS}& $t(s)$ & 11.8 & 12.3 & 12.3& 12.5 & 34.3 & 36.9 & 34.4  \\ \hline

 \rowcolor[gray]{.8}ALC & $e\%$ &  47.0 &        46.0&         48.8&        100.0 &       100.0 &       100.0 &        65.3 \\
 \rowcolor[gray]{.8}(voting with $K$) & $t(s)$ &  1469.2 &      1445.6 &      1489.2 &       679.0&       1530.1 &      1528.5&       3032.4 \\ \hline

ALC & $e\%$ &    50.5 &  50.8 &  50.5&   99.8 &  99.8 &  99.8 &  67.0 \\
($\epsilon$ from LBF) & $t(s)$ &  93.0  &       93.6 &        91.0 &         9.4 &        18.2 &        17.9 &       163.5 \\ \hline

\rowcolor[gray]{.8}& $e\%$ & 5.8 & 4.9 & 5.3 & 17.1 & 23.0 & 29.7 & 33.6 \\
\rowcolor[gray]{.8}\raisebox{1.5ex}[0pt]{SCC}&$t(s)$& \textbf{0.9} & \textbf{1.0} & \textbf{1.1} & \textbf{0.9} & \textbf{1.6} & 2.0 & \textbf{1.7} \\ \hline

& $e\%$ & \textbf{5.1} & 5.4 & 5.1 & 26.2 & 28.6 & 41.7 & 33.0 \\
\raisebox{1.5ex}[0pt]{SCC-MS} & $t(s)$ & \textbf{0.9}& \textbf{1.0} & 1.2 & 1.0 & 1.8 & \textbf{1.9} & 2.0 \\ \hline

\rowcolor[gray]{.8}& $e\%$ & N/A & N/A & N/A & N/A & N/A & N/A & N/A \\
\rowcolor[gray]{.8}\raisebox{1.5ex}[0pt]{GPCA}&$t(s)$& N/A & N/A & N/A & N/A & N/A & N/A & N/A \\ \hline

& $e\%$ & 10.9 & 14.9 & 13.5 & 30.4 & 45.3 & 41.6 & 26.9 \\
\raisebox{1.5ex}[0pt]{K-flats} & $t(s)$ & 2.8 & 2.9 & 2.9 & 3.1 & 6.2 & 5.6 & 5.1  \\ \hline

%\rowcolor[gray]{.8}& $e\%$ &  17.0  &  2.2 &   3.2 &  20.7 &  11.5 &  17.7 &  46.5 \\
%\rowcolor[gray]{.8}\raisebox{1.5ex}[0pt]{SSC1}&$t(s)$&    411.78  &     402.73 &      395.07 &      395.96 &      760.94 &      763.09 &      776.98\\ \hline

\rowcolor[gray]{.8}& $e\%$ &   16.8  &  2.0 &   3.2 &  20.0 &  11.3 &  17.8 &  45.5 \\
\rowcolor[gray]{.8}\raisebox{1.5ex}[0pt]{SSC} & $t(s)$ &   411.8  &     402.7 &      395.1 &      396.0 &      760.9 &      763.1 &      777.0\\ \hline

%\rowcolor[gray]{.8}& $e\%$ &15.0  &  2.0 &   2.5 &  16.5 &  11.0 &  16.3 &  46.5 \\
%\rowcolor[gray]{.8}\raisebox{1.5ex}[0pt]{SSC3}&$t(s)$&   411.78  &     402.73 &      395.07 &      395.96 &      760.94 &      763.09 &      776.98\\ \hline

\end{tabular}

\end{table*}

\setlength{\abovecaptionskip}{4pt}   % 0.5cm as an example
\setlength{\belowcaptionskip}{4pt}
\begin{table*}[htbp]
\centering \caption{\small {The standard deviation to the mean percentage of misclassified points on clustering MNIST data set (D=5 for GPCA, D=10 for other algorithms). }}\label{tab:MNIST_std}
\scriptsize

\begin{tabular}{|c||c|c|c|c|c|c|c||}
\hline
{subsets} & [1 2] & [1 3] & [1 7] & [4 7] & [2 4 8] & [3 6 8] & [1 2 3] \\ \hline
{$K$} & 2&2&2&2&3&3&3 \\ \hline \hline

 \rowcolor[gray]{.8}LBF  & 3.5 & 4.1 & 10.0 & 11.4 & 11.6 & 8.3 & 9.5 \\ \hline

LBF-MS  & 5.9 & 3.8 & 10.0 & 10.0 & 10.3 & 7.2 & 7.8 \\ \hline

 \rowcolor[gray]{.8}SLBF  &  \textbf{0.0} & \textbf{0.0} & \textbf{0.0} & \textbf{0.0} & 0.0 & \textbf{0.0} & \textbf{0.0} \\ \hline

 SLBF-MS  &  \textbf{0.0} & \textbf{0.0} & \textbf{0.0} & \textbf{0.0} & \textbf{0.0} & \textbf{0.0} & \textbf{0.0} \\ \hline

 \rowcolor[gray]{.8}ALC(voting with $K$)  & \textbf{0.0} & 0.0 & \textbf{0.0} & \textbf{0.0} & \textbf{0.0} & \textbf{0.0} & \textbf{0.0}  \\ \hline

ALC($\epsilon$ from LBF)  & \textbf{0.0} & \textbf{0.0} & \textbf{0.0} & \textbf{0.0} & \textbf{0.0} & \textbf{0.0} & \textbf{0.0} \\ \hline

 \rowcolor[gray]{.8}SCC  & 2.3 & 2.7 & 4.6 & 9.9 & 9.4 & 7.5 & 11.9 \\ \hline

 SCC-MS  & 2.0 & 3.7 & 5.2 & 10.2 & 8.3 & 8.5 & 9.2 \\ \hline

\rowcolor[gray]{.8} GPCA  & \textbf{0.0} & 0.0 & \textbf{0.0} & \textbf{0.0} & \textbf{0.0} & \textbf{0.0} & \textbf{0.0} \\ \hline

K-flats  & 7.6 & 8.5 & 7.8 & 5.7 & 7.4 & 7.5 & 5.9  \\ \hline

%\rowcolor[gray]{.8} SSC1  & \textbf{0.0} & 0.0 & 0.0 & 0.0 & 0.0 & 0.0 & 0.0 \\ \hline
\rowcolor[gray]{.8}SSC  & \textbf{0.0} & \textbf{0.0} & \textbf{0.0} & \textbf{0.0} & \textbf{0.0} & \textbf{0.0} & \textbf{0.0} \\ \hline
%\rowcolor[gray]{.8} SSC3  & 0.0 & 0.0 & 0.0 & 0.0 & 0.0 & 0.0 & 0.0 \\ \hline

\end{tabular}

\end{table*}
\setlength{\abovecaptionskip}{4pt}   % 0.5cm as an example
\setlength{\belowcaptionskip}{4pt}
\begin{table*}[htbp]
\centering \caption{\small {The standard deviation of the mean percentage of misclassified points on clustering MNIST data set (D=50). }}\label{tab:MNIST_std2}
\scriptsize

\begin{tabular}{|c||c|c|c|c|c|c|c||}
\hline
{subsets} & [1 2] & [1 3] & [1 7] & [4 7] & [2 4 8] & [3 6 8] & [1 2 3] \\ \hline
{$K$} & 2&2&2&2&3&3&3 \\ \hline \hline

 \rowcolor[gray]{.8}LBF  & 5.6 & 8.0 & 8.3 & 10.6 & 11.0 & 6.0 & 6.0 \\ \hline

LBF-MS  & 8.7 & 10.5 & 11.4 & 11.2 & 12.3 & 8.9 & 9.1 \\ \hline

 \rowcolor[gray]{.8}SLBF  &   \textbf{0.0} & 0.0 & \textbf{0.0} & \textbf{0.0} & \textbf{0.0} & \textbf{0.0} & \textbf{0.0} \\ \hline

 SLBF-MS  &  \textbf{0.0} & 0.0 & \textbf{0.0} & \textbf{0.0} & \textbf{0.0} & \textbf{0.0} & \textbf{0.0} \\ \hline

 \rowcolor[gray]{.8}ALC(voting with $K$)  &  \textbf{0.0} & 0.0 & \textbf{0.0} & \textbf{0.0} & \textbf{0.0} & \textbf{0.0} & \textbf{0.0} \\ \hline

ALC($\epsilon$ from LBF)  &  \textbf{0.0} & 0.0 & \textbf{0.0} & \textbf{0.0} & \textbf{0.0} & \textbf{0.0} & \textbf{0.0} \\ \hline

 \rowcolor[gray]{.8}SCC  & 0.6 & 1.0 & 0.9 & 10.3 & 3.7 & 4.3 & 13.9 \\ \hline

 SCC-MS  & 0.4 & 0.7 & 0.9 & 15.5 & 5.4 & 4.5 & 5.8 \\ \hline

\rowcolor[gray]{.8} GPCA  & N/A & N/A & N/A & N/A & N/A & N/A & N/A \\ \hline

K-flats  & 7.2 & 11.3 & 11.1 & 7.5 & 7.3 & 8.1 & 7.7  \\ \hline

%\rowcolor[gray]{.8} SSC1  & 0.0 & 0.0 & 0.0 & 0.0 & 0.0 & 0.0 & 0.0 \\ \hline
\rowcolor[gray]{.8}SSC  &  \textbf{0.0} & 0.0 & \textbf{0.0} & \textbf{0.0} & \textbf{0.0} & \textbf{0.0} & \textbf{0.0}\\ \hline
%\rowcolor[gray]{.8} SSC3  & 0.0 & 0.0 & 0.0 & 0.0 & 0.0 & 0.0 & 0.0 \\ \hline

\end{tabular}

\end{table*}
 From Table \ref{tab:MNIST_seg} and \ref{tab:MNIST_seg2}, SLBF and SLBF-MS are the best algorithms among all the methods in terms of misclassification rates, although these misclassification rates are larger when $K=3$. SCC, SCC-MS, SSC, LBF and LBF-MS are also good algorithms for this data set. ALC is almost as good as SLBF and SLBF-MS when $K=2$, but it fails when $K=3$.  LBF, LBF-MS and $K$-flats are the fastest algorithms in MNIST data set.

\subsection{Automatic determination of the number of flats}

%\subsection{Number of clusters via the elbow method }
We explain how to use the elbow method to determine the number of affine clusters in any HLM algorithm, in particular LBF and SLBF.
Fixing an arbitrary HLM algorithm with the correct input of number of clusters $K$, let $F_j,\,j=1,\dots,K$ be the $K$ flats returned by that algorithm
and $W_K$ be the sum of squared distances of all data points to the flat, among these $K$ flats, corresponding to their clusters.
That is,
\begin{equation}
W_K = \sum_{j=1}^K \sum_{ {\bx} \in C_j} \dist^2({\bx}, F_j).
\end{equation}
We note that $W_K$ decreases as $K$ increases.

A classical method for determining the number of clusters is to find the ``elbow'', or the $K$ past which adding more clusters does not significantly decrease the error.  We search for the elbow by finding the maximum of the Second Order Difference (SOD) of the logarithm of $W_K$~\cite{SOD}:
\begin{equation}
\mathrm{SOD}(\ln W_K) = \ln W_{K-1} + \ln W_{K+1} - 2\ln W_K.
\label{eq:SODlnWk}
\end{equation}
The optimal $K$ is thus found by
\begin{equation}
K_{\mathrm{SOD}} = \arg \max_K \mathrm{SOD}(\ln W_K),  \label{eq:optlnK}.
\end{equation}
where $K=2,\dots,K_{\max}$.

In the following sections, we compare SOD (LBF), i.e., SOD applying LBF, SOD (LBF-MS), SOD (SLBF), SOD (SLBF-MS), SOD (SCC), SOD (SCC-MS) and SOD(SSC) with ALC~\cite{Ma07Compression} and part of GPCA~\cite{Vidal05} on a number of artificial data sets and real data sets. These experiments  run on a machine with Intel Core 2 Quad Q6600 at 2.4GHz and 8 GB memory.

\subsubsection{Finding the number of clusters on artificial data}
We test SOD with LBF and SLBF on artificial data (where the number of clusters is not provided to the user)
and compare them with some other methods
(three variations of ALC, number of clusters by GPCA and SOD with SSC and SCC).
The artificial data sets are
generated by the Matlab code borrowed from the GPCA \cite{Vidal05}
package on {http://perception.csl.\\uiuc.edu/gpca}. For each subspace
$100d$ initial data points are uniformly sampled in a unit cube in this
subspace centered around the origin and then corrupted with Gaussian
noise in $\reals^D$ of standard deviation 0.05.
For the last four experiments, we restrict the angle between subspaces
to be at least $\pi/8$ for separation. The dimension $d$ is given and we
let $K_{\max}=10$ in SOD.

In ALC (voting), we try 101 different values
from $10^{-5}$ to $10^3$ for $\epsilon$ (as in~\cite{ALC_motion}) and
choose the estimated $K$ by majority. In ALC ($\epsilon$ from LBF), we
choose the average noise in the neighborhood using the local best-fit
heuristic as the distortion rate $\epsilon$. In ALC (oracle), we input
the true noise level ($\epsilon=0.05$) as the distortion rate. For GPCA,
we use the original idea of~\cite[eqs.~(26), (28)]{Vidal05} to
find the number of clusters (see our implementation in
the supplemental webpage). We project the data onto a $d+1$-dimensional
subspace by PCA and let the tolerance of rank detection be $0.05$ (chosen by trying
 different values and picking the one obtaining the lowest error). This algorithm is independent of
 other parts of the GPCA algorithm and is thus extremely fast and can perform in high ambient dimensions. We
even tried other ideas of~\cite[eqs.~(3.28), (3.29)]{Ma07} (for
the same given dimension $d$), while applying them to several HLM algorithms (even though they were originally presented for GPCA).
Nevertheless, they did not work well and we thus did not report them.
Each experiment is repeated $100$ times (except for SOD(SSC), which is repeated $10$ times due to its low speed) and the error rates of finding the number of clusters $K$ and the computation time (in seconds) are recorded in Table~\ref{tab:SOD}.

\begin{table*}[htbp]
\centering \caption{\small {The mean percentage of incorrectness ($e\%$)
for finding the number of clusters $K$ and the computation time in seconds $t(s)$ on artificial data.}}\label{tab:SOD}
\scriptsize
\setlength{\tabcolsep}{0pt}

\begin{tabular}{|c|c||c|c|c|c|c|c|c||c|c|c|c||}
\hline

\multicolumn{2}{|c||}{}&\multicolumn{7}{c||}{no minimum angle}&\multicolumn{4}{c||}{minimum
angle
$=\pi/8$} \\
\hline

\multicolumn{2}{|c||}{}&$1^6\in\mathbb{R}^5$ &$2^4\in\mathbb{R}^5$ &$3^3\in\mathbb{R}^5$
 &$1^6\in\mathbb{R}^3$ &$2^4\in\mathbb{R}^3$ &$3^3\in\mathbb{R}^4$&$10^2\in\mathbb{R}^{15}$ &$1^6\in\mathbb{R}^3$
&$2^4\in\mathbb{R}^3$ &$3^3\in\mathbb{R}^4$&$10^2\in\mathbb{R}^{15}$
 \\

\hline

\hline
 \rowcolor[gray]{.8}SOD &e\%&22&2&\textbf{0}&58&32&12&\textbf{0}&2&6&5&\textbf{0}\\
 \rowcolor[gray]{.8}(LBF)&$t(s)$&10.43&13.76&14.83&9.84&13.08&14.49&34.16&9.95&13.22&14.47&34.04\\
\hline

SOD &e\%&13&1&3&67&33&9&\textbf{0}&3&8&6&\textbf{0}\\
(LBF-MS)&$t(s)$&8.70&11.90&12.92&8.37&11.54&12.84&27.56&8.42&11.60&12.84&27.69 \\
\hline

 \rowcolor[gray]{.8}SOD &e\%&75&10&5&\textbf{0}&90&95&55&\textbf{0}&85&90&55\\
 \rowcolor[gray]{.8}(SLBF)&$t(s)$& 1097.19 & 2148.06 & 2895.85 & 1076.24 & 1774.74 & 2629.26 &16124.50 & 1224.96 & 2387.70 & 2830.83 & 16510.13\\
\hline

SOD &e\%&90&95&70&\textbf{0}&90&85&85&0&75&80&80\\
(SLBF-MS)&$t(s)$& 908.76 & 2094.68 & 3141.77 & 927.25 & 1740.03 & 2695.59 & 15754.05 & 990.88 & 2302.66 & 3010.64 & 16493.95 \\
\hline

 \rowcolor[gray]{.8}ALC&$e\%(K)$&24&12&11&32&30&17&100&5&9&9&100 \\
 \rowcolor[gray]{.8}(voting)&$t(s)$&2094.75&2700.07&3530.26&1207.54&2346.69&3628.24&119584.04&1184.08&2354.19&3956.05&117353.17\\
\hline

ALC&$e\%(K)$& \textbf{1} & \textbf{0} & 1 & 20 & \textbf{20} & 3 & 58 & \textbf{0} & \textbf{3} & \textbf{1} & 63           \\
($\epsilon$ from LBF) & $t(s)$& 23.72 & 43.50 & 57.50 & 19.76 & 36.67 & 53.25 & 1516.02 & 19.81 & 36.60 & 53.01 & 1770.77 \\ \hline

 \rowcolor[gray]{.8}ALC&$e\%(K)$&\textbf{1}&\textbf{0}&\textbf{0}&34&31&\textbf{1}&16&\textbf{0}&10&\textbf{1}&13\\
 \rowcolor[gray]{.8}(oracle)&$t(s)$&23.74&43.44&59.14&20.49&37.49&53.59&1370.92&20.22&37.41&54.11&1354.11\\
\hline

&e\%&88&100&100&27&100&100&100&13&100&100&100\\
\raisebox{1.5ex}[0pt]{GPCA}&$t(s)$&\textbf{0.03}&\textbf{0.09}&\textbf{0.12}&\textbf{0.06}&\textbf{0.09}&\textbf{0.12
}&\textbf{1.30}&\textbf{0.04}&\textbf{0.09}&\textbf{0.12}&\textbf{1.30}\\
\hline

\rowcolor[gray]{.8}SOD&$e\%(K)$&35&21&1&63&39&17&\textbf{0}&9&32&11&1\\
 \rowcolor[gray]{.8}(SCC)&$t(s)$&32.09&61.26&95.79&25.83&59.41&76.13&475.45&26.74&41.95&61.53&466.79\\
\hline

SOD&$e\%(K)$& 71&32&2&80&50&12&\textbf{0}&46&33&3&\textbf{0}          \\
(SCC-MS) & $t(s)$&31.78&67.77&111.15&22.29&55.25&74.07&475.50&24.53&51.98&75.03& 471.31 \\ \hline

\rowcolor[gray]{.8}SOD&$e\%(K)$&10&80&70&100&70&70& 100  &50&80&80&100\\
 \rowcolor[gray]{.8}(SSC)&$t(s)$&39.88&2634.80&3039.55&1708.37&2447.01&2925.27&14918.10&1452.43&2101.84&2641.68&14227.32\\
\hline

\end{tabular}

\end{table*}
As in Table \ref{tab:SOD}, ALC (oracle) and ALC ($\epsilon$ from LBF) work the best for low dimensions ($d=1,2,3$), but in real problems this choice (the noise level) for $\epsilon$ is usually unknown. The local best-fit flat heuristic provides a good estimation for the distortion rate and helps ALC reduce its running time. ALC (voting) is not as good as SOD (LBF) for artificial data. All options of ALC suffer from the computation complexity, especially ALC (voting). SOD (LBF) and SOD (LBF-MS) get reasonable outputs and have obvious advantage of computing time. GPCA is very fast, but does not work well.

\subsubsection{Finding the number of clusters on the extended Yale face database B}
We use the extended Yale face database B as in Section~\ref{sec:face_seg} for testing the above algorithms for detecting the number of clusters.
The ambient dimension is reduced to $D=5K$ by PCA for all of the methods and the intrinsic dimension of subspaces is set as $d=5$ (see Section~\ref{sec:face_seg}). For SOD with different clustering algorithms, we let $K_{\max}=6$, $8$, $8$, $10$ and $10$ respectively for $2$ to $6$ clusters. For GPCA, we let tolerance be $0.05$ which does not affect the performance in this experiment. Each experiment is repeated $500$ times (except for SOD(SSC), which is repeated $10$ times due to its low speed). Following Section~\ref{sec:face_seg}, we apply LBF, LBF-MS, SLBF and SLBF-MS with whitening. The error rates of finding the correct number of clusters and the computation time are recorded in Table \ref{tab:face_SOD}.

\setlength{\abovecaptionskip}{4pt}   % 0.5cm as an example
\setlength{\belowcaptionskip}{4pt}
\begin{table*}[htbp]
\centering \caption{\small {The mean percentage of incorrectness ($e\%$)
for finding the correct number of clusters $K$ and the computation time in seconds $t(s)$ on the extended Yale
face database B.}}\label{tab:face_SOD}
\scriptsize

\begin{tabular}{|c|c||c|c|c|c|c||}
\hline

\multicolumn{2}{|c||}{Real $K$} & 2&3&4&5&6\\ \hline \hline

% \rowcolor[gray]{.8}SOD & $e\%(K)$  & 58 & 59 & 64 & 71 & 63 & 83 & 91 & 84 & 91 \\
% \rowcolor[gray]{.8}(LBF,M1) & $t(s)$  & 1.42 & 3.95 & 6.19 & 12.04 & 16.61 & 31.25 & 51.29 & 80.79 & 122.88 \\ \hline
%
%SOD & $e\%(K)$  & 60 & 64 & 75 & 58 & 66 & 77 & 77 & 91 & 93 \\
%(LBF-MS,M1) & $t(s)$  & 0.67 & 1.65 & 2.49 & 4.89 & 6.83 & 13.04 & 22.27 & 36.12 & 55.33 \\ \hline

\rowcolor[gray]{.8}SOD & $e\%(K)$  & 62 & 61 & 69 & 78 & 84  \\
 \rowcolor[gray]{.8}(LBF) & $t(s)$  & 1.30 & 3.60 & 5.69 & 11.30 & 15.84  \\ \hline

SOD & $e\%(K)$   & 65 & 75 & 78 & 81 & 80  \\
(LBF-MS) & $t(s)$   & 0.67 & 1.65 & 2.49 & 4.90 & 6.83   \\ \hline

% \rowcolor[gray]{.8}SOD & $e\%(K)$  & 34 & 86 & 96 & 96 \\
% \rowcolor[gray]{.8}(SLBF,M1) & $t(s)$ & 109.58 & 289.01 & 329.14 & 701.27 \\ \hline
%
% SOD & $e\%(K)$  & 26 & 46 & 40 & 54  \\
% (SLBF-MS,M1) & $t(s)$ & 100.80 & 261.31 & 300.66 & 645.52 \\ \hline

  \rowcolor[gray]{.8}SOD & $e\%(K)$   & 24 & 60 & 70 & 86 &98\\
 \rowcolor[gray]{.8}(SLBF) & $t(s)$  & 115.97 & 303.02 & 338.35 & 729.74 &811.40\\ \hline

 SOD & $e\%(K)$  & \textbf{20} & 60 & 76 & 92&96  \\
 (SLBF-MS) & $t(s)$  & 106.87 & 272.88 & 306.22 & 649.50& 721.42 \\ \hline

 \rowcolor[gray]{.8}ALC  & $e\%(K)$  & 100 & 100 & 100 & 100 & 100  \\
 \rowcolor[gray]{.8}(voting) & $t(s)$  & 42.99 & 122.29 & 258.20 & 451.07 & 699.52 \\ \hline

ALC  & $e\%(K)$  & 42 & 36 & 76 & 86 & 100  \\
($\epsilon$ from LBF) & $t(s)$  & 0.95 & 2.49 & 5.54 & 11.54 & 24.38  \\ \hline

%\rowcolor[gray]{.8}ALC(9$K$) & $e\%(K)$  & 100 & 100 & 100 & 100 & 100  \\
% \rowcolor[gray]{.8}(voting) & $t(s)$   & 55.51 & 149.31 & 300.39 & 543.04 & 920.03 \\ \hline
%
%ALC(9$K$) & $e\%(K)$   & 28 & 16 & 92 & 100 & 100  \\
%($\epsilon$ from LBF) & $t(s)$  & 0.95 & 2.49 & 5.54 & 11.54 & 24.38  \\ \hline

  \rowcolor[gray]{.8}& $e\%(K)$ & 100 & 100  & 100 & 100 & 100 \\
 \rowcolor[gray]{.8}\raisebox{1.5ex}[0pt]{GPCA}&$t(s)$& \textbf{ 0.07 }& \textbf{ 0.13 }& \textbf{ 0.52 }& \textbf{ 0.71 }& \textbf{ 1.02 } \\ \hline

% SOD&$e\%(K)$  & 4 & 36 & 70 & 80 & 88   \\
% (SCC)&$t(s)$  & 23.77 & 62.34 & 109.49 & 185.63 & 131.95 \\
%\hline

%\rowcolor[gray]{.8}SOD&$e\%(K)$ & 4 & 50        \\
%\rowcolor[gray]{.8}(SCC-MS) & $t(s)$ & 32.91 & 81.66 \\ \hline

% SOD&$e\%(K)$   & 8 & 48 & 64 & 88 & 80 & 90 & 92  \\
% (SCC(9$K$))&$t(s)$  & 24.27 & 62.38 & 110.72 & 188.29 & 138.75 & 242.73 & 777.66 \\
%\hline
%
%\rowcolor[gray]{.8}SOD&$e\%(K)$ & 12 & 73         \\
%\rowcolor[gray]{.8}(SCC-MS(9$K$)) & $t(s)$ & 33.80 & 82.34 \\ \hline

SOD&$e\%(K)$ & 100 & \textbf{ 8 }& \textbf{ 12 }& \textbf{ 28 }& \textbf{ 38 }\\
(SSC)&$t(s)$ & 172.50 & 389.66 & 567.39 & 1015.99 & 1336.57\\

% \rowcolor[gray]{.8}SOD&$e\%(K)$& 100 & 18 & 18 & 26 & 50 \\
% \rowcolor[gray]{.8}(SSC(9$K$))&$t(s)$ & 178.18 & 413.47 & 623.92 & 1172.43 & 1649.51 \\
\hline
\end{tabular}

\end{table*}

We see from Table \ref{tab:face_SOD} that SOD only performs well with SSC with $K$ smaller than 4. We note that this is due to the difficulty of the database. Indeed for a simpler database such as  Yale Face database B~\cite{GeBeKr01} of uncropped faces, SOD (SLBF), SOD (SLBF-MS), ALC ($\epsilon$ from LBF) and ALC (voting) have perfect detection for $K\leq 10$ (whitening is not applied then).%; %SOD (SLBF) and SOD (SLBF-MS) are faster than ALC (voting) on data with small sizes, but slower with large sizes; LBF provides a good estimation for the distortion rate and makes ALC ($\epsilon$ from LBF) much faster than ALC (voting); SOD (LBF) and SOD (SLBF) have nearly perfect detection and obvious advantage of efficiency; GPCA does not work well on finding the number of clusters.

\subsubsection{Finding the number of clusters on MNIST data set}
We preprocess MNIST data set exactly the same way as we did in Section \ref{sec:MNIST_seg}. The ambient dimension is reduced to both $D=10$ and $D=50$ by PCA for all of the methods including GPCA and 3 is given as the intrinsic subspace dimension. For SOD with different clustering algorithms, we let $K_{\max}=6, \:\mathrm{and}\:8$ respectively for 2 and 3 clusters. For GPCA, we let the tolerance be 0.05 which does not affect the performance in this experiment. Each experiment is repeated $500$ times (except for SOD(SSC), which is repeated $10$ times due to its low speed). The error rates of finding the correct number of clusters and the computation time are recorded in Tables \ref{tab:MNIST_SOD} and \ref{tab:MNIST_SOD2}.

\setlength{\abovecaptionskip}{4pt}   % 0.5cm as an example
\setlength{\belowcaptionskip}{4pt}
\begin{table*}[htbp]
\centering \caption{\small {The mean percentage of incorrectness ($e\%$)
for finding the correct number of clusters $K$ and the computation time in seconds $t(s)$ on MNIST data set (D=10). }}\label{tab:MNIST_SOD}
\scriptsize

\begin{tabular}{|c|c||c|c|c|c|c|c|c||}
\hline
\multicolumn{2}{|c||}{subsets} & [1 2] & [1 3] & [1 7] & [4 7] & [2 4 8] & [3 6 8] & [1 2 3] \\ \hline
\multicolumn{2}{|c||}{$K$} & 2&2&2&2&3&3&3 \\ \hline \hline

 \rowcolor[gray]{.8}SOD & $e\%$ & 16.8 & 3.8 & 50.8 & 50.4 & 75.6 & 70.0 & 54.8 \\
 \rowcolor[gray]{.8}(LBF)& $t(s)$ & 3.5 & 3.2 & 3.0 & 3.3 & 7.7 & 7.5 & 7.3  \\ \hline

SOD& $e\%$ & 9.6 & 6.6 & 33.4 & 68.2 & 80.4 & 76.6 & 44.2 \\
(LBF-MS)& $t(s)$ & 1.9 & 1.9 & 1.9 & 1.8 & 4.6 & 4.6 & 4.7  \\ \hline

 \rowcolor[gray]{.8}SOD & $e\%$ & \textbf{0.0} & \textbf{0.0} & \textbf{0.0} & \textbf{0.0} & \textbf{0.0} & 100.0 & \textbf{0.0} \\
 \rowcolor[gray]{.8}  (SLBF)& $t(s)$ & 173.9 & 164.6 & 160.3 & 248.6 & 710.1 & 610.9 & 548.5  \\ \hline

SOD & $e\%$ & \textbf{0.0} & \textbf{0.0} & \textbf{0.0} & \textbf{0.0} & \textbf{0.0} & 100.0 & \textbf{0.0}  \\
(SLBF-MS)& $t(s)$ & 164.6 & 159.9 & 150.1 & 228.5 & 676.6 & 586.4 & 556.2  \\ \hline

 \rowcolor[gray]{.8}ALC & $e\%$ & 100.0 & 100.0 & 100.0 & 100.0 & 100.0 & 100.0 & 100.0 \\
 \rowcolor[gray]{.8}(voting) & $t(s)$ & 830.4 & 823.2 & 813.2 & 753.2 & 1789.5 & 1871.8 & 1987.5   \\ \hline

ALC & $e\%$ & 100.0 & 100.0 & 100.0 & 100.0 & 100.0 & \textbf{0.0} & 100.0 \\
($\epsilon$ from LBF) & $t(s)$ & 23.2&        22.5&        21.5&        22.9   &     55.6     &   54.7       &54.0   \\ \hline

\rowcolor[gray]{.8}& $e\%$ & 100.0 & 100.0 & 100.0 & 100.0 & 100.0 & 100.0 & 100.0 \\
\rowcolor[gray]{.8}\raisebox{1.5ex}[0pt]{GPCA}&$t(s)$&\textbf{ 1.0 }&\textbf{ 1.0 }&\textbf{ 1.0 }&\textbf{ 1.1 }&\textbf{ 2.8 }&\textbf{ 2.8 }&\textbf{ 2.7} \\ \hline

SOD&$e\%(K)$&3.8&7.8&66.4&81.8&64.4&47.6&82.6\\
 (SCC)&$t(s)$&14.5&13.3&14.7&16.9&37.5&34.1&35.0\\
\hline

\rowcolor[gray]{.8}SOD&$e\%(K)$& 2.4&16.4&53.0&77.4&70.4&49.6&77.8         \\
\rowcolor[gray]{.8}(SCC-MS) & $t(s)$&13.7&13.8&13.5&16.4&38.0&35.6&29.4 \\ \hline

% SOD& $e\%(K)$ & 0.0 & 0.0 & 0.0 & 100.0 & 100.0 & 100.0 & 100.0  \\
%(SSC1) & $t(s)$ &233.62 &      210.33 &      213.34 &      218.46 &      380.06 &      386.45 &      390.54 \\ \hline

 SOD& $e\%(K)$ & \textbf{0.0} & \textbf{0.0}  & \textbf{0.0} & 100.0 & \textbf{0.0} & 100.0 & 100.0  \\
(SSC)&$t(s)$& 233.6 &      210.3 &      213.3 &      218.4 &      380.0 &      386.4&      390.5 \\ \hline

%SOD & $e\%(K)$ & 0.0 & 0.0  & 0.0 & 0.0 & 0.0 & 100.0 & 100.0  \\
%(SSC3) & $t(s)$ & 233.62 &      210.33 &      213.34 &      218.46 &      380.06 &      386.45 &      390.54  \\ \hline

\end{tabular}

\end{table*}
\begin{table*}[htbp]
\centering \caption{\small {The mean percentage of incorrectness ($e\%$)
for finding the correct number of clusters $K$ and the computation time in seconds $t(s)$ on MNIST data set (D=50).}}\label{tab:MNIST_SOD2}
\scriptsize

\begin{tabular}{|c|c||c|c|c|c|c|c|c||}
\hline
\multicolumn{2}{|c||}{subsets} & [1 2] & [1 3] & [1 7] & [4 7] & [2 4 8] & [3 6 8] & [1 2 3] \\ \hline
\multicolumn{2}{|c||}{$K$} & 2&2&2&2&3&3&3 \\ \hline \hline

 \rowcolor[gray]{.8}SOD & $e\%$ & 45.0 & 35.0 & 54.0 & 79.0 & 72.0 & 67.0 & 60.0 \\
 \rowcolor[gray]{.8}(LBF)& $t(s)$ & 22.9 & 23.5 & 22.2 & 24.9 & 56.2 & 54.6 & 51.1 \\\hline

SOD& $e\%$ & 32.0 & 22.0 & 38.0 & 66.0 & 44.0 & 82.0 & 58.0 \\
(LBF-MS)& $t(s)$ & \textbf{12.2} & 12.2 & 12.2 & 12.2 & 29.3 & 29.4 & 29.4  \\ \hline

 \rowcolor[gray]{.8}SOD & $e\%$ & \textbf{0.0} & \textbf{0.0} & \textbf{0.0} & \textbf{0.0} & \textbf{0.0} & 100.0 & 100.0 \\
 \rowcolor[gray]{.8}  (SLBF)& $t(s)$ & 204.2 & 198.1 & 207.8 & 295.8 & 864.5 & 766.5 & 706.1  \\ \hline

SOD & $e\%$ & \textbf{0.0} & \textbf{0.0} & 100.0 & \textbf{0.0} & \textbf{0.0} & 100.0 & 100.0  \\
(SLBF-MS)& $t(s)$ & 213.7 & 201.7 & 176.6 & 259.9 & 748.1 & 640.0 & 681.1  \\ \hline

 \rowcolor[gray]{.8}ALC & $e\%$ & 100.0 & 100.0 & 100.0& 100.0 & 100.00 & 100.0 & 100.0 \\
 \rowcolor[gray]{.8}(voting) & $t(s)$ &  1469.2 &      1445.6 &      1489.2 &       679.0&       1530.1 &      1528.5&       3032.4   \\ \hline

ALC & $e\%$ & 100.0 & 100.0 & 100.0 & 100.0 & 100.0 & 100.0 & 100.0 \\
($\epsilon$ from LBF) & $t(s)$ & 93.0  &       93.6 &        91.0 &         \textbf{9.4} &        \textbf{18.2} &        \textbf{17.9} &       163.5    \\ \hline

\rowcolor[gray]{.8}& $e\%$ & N/A & N/A & N/A & N/A & N/A & N/A & N/A \\
\rowcolor[gray]{.8}\raisebox{1.5ex}[0pt]{GPCA}&$t(s)$& N/A & N/A & N/A & N/A & N/A & N/A & N/A \\ \hline

SOD&$e\%(K)$& \textbf{0.0}&    4.0&    1.0&   50.5 &  78.8 &  30.3&   83.8\\
 (SCC)&$t(s)$& 14.9&   \textbf{10.6}&   \textbf{11.6}&   11.6 &  24.7 &  26.2 &  \textbf{25.4}\\
\hline

\rowcolor[gray]{.8}SOD&$e\%(K)$& \textbf{0.0}&     \textbf{0.0}&     \textbf{0.0}&   42.4&   89.9&   97.0 &  93.9         \\
\rowcolor[gray]{.8}(SCC-MS) & $t(s)$&12.6 &  13.0 &  14.7 &  13.9&   34.0 &  36.8 &  30.7 \\ \hline

% SOD& $e\%(K)$ &0.0 & 0.0  & 0.0 & 0.0 & 0.0 & 100.0 & 100.0   \\
%(SSC1) & $t(s)$ &426.45 &      417.69 &      409.31 &      413.50  &     823.82 &      821.28 &      836.87 \\ \hline

  SOD& $e\%(K)$ & \textbf{0.0} & \textbf{0.0}  & \textbf{0.0} & \textbf{0.0} & \textbf{0.0} & 100.0 & 100.0  \\
 (SSC)&$t(s)$& 426.4 &      417.6 &      409.3 &      413.5  &     823.8 &      821.2 &      836.8 \\ \hline

%SOD & $e\%(K)$ & 0.0 & 0.0  & 0.0 & 100.0 & 0.0 & 100.0 & 100.0  \\
%(SSC3) & $t(s)$ & 426.45 &      417.69 &      409.31 &      413.50  &     823.82 &      821.28 &      836.87 \\ \hline

\end{tabular}

\end{table*}

%\subsection{Conclusion about finding the number of clusters}

For all the methods, determining the number $K$ of clusters becomes very difficult when the real $K$ is larger than $3$. For real $K \leq 3$, we see from Table \ref{tab:MNIST_SOD} that when we project data to $10$-dimensional space, ALC and GPCA fail in most cases, except for ALC ($\epsilon$ from LBF) on digits [3 6 8]. SOD (SLBF), SOD (SLBF-MS) and SOD (SSC) outperform all others although they are not very efficient.

\subsection{Initializing $K$-flats with the local best-fit heuristic}
\label{sec:ikf}
Here we demonstrate that our choice of neighborhoods in Algorithm~\ref{alg:nhbd}
can be used to get a more robust initialization of $K$-flats.  We work
with geometric farthest insertion.  For fixed neighborhood
sizes, say of $m$ neighbors, this goes as follows:  we pick a random
point $\bx_0$ and then find the best-fit flat $F_0$ for the $m$
point neighborhood of $\bx_0$.  Then we find the point $\bx_1$ in
our data farthest from $F_0$, find the best-fit flat $F_1$ of the
$m$ neighborhood of $\bx_1$, and then choose the point $\bx_2$
farthest from $F_0$ and $F_1$ to continue.  We stop when we have $K$
flats; we use these as an initialization for $K$-flats.

\begin{figure}
\begin{center}
\includegraphics[width=.4\textwidth,height=.3\textwidth]{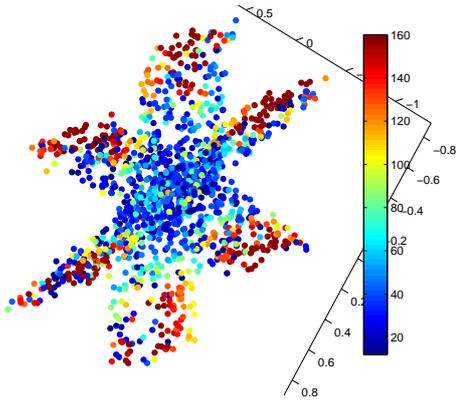}
\caption{\it Color map of neighborhood size obtained by the local best-fit flat heuristic.  The color
value represents the number of neighbors chosen at that point. Note
that the algorithm chooses smaller neighborhoods for points closer
to the intersection of the planes. \label{fig:nonei}}
\end{center}
\end{figure}

\begin{figure*}[htbp]
\begin{center}
\begin{minipage}{0.31\textwidth} \centering
\includegraphics[width=1\textwidth,height=.8\textwidth]{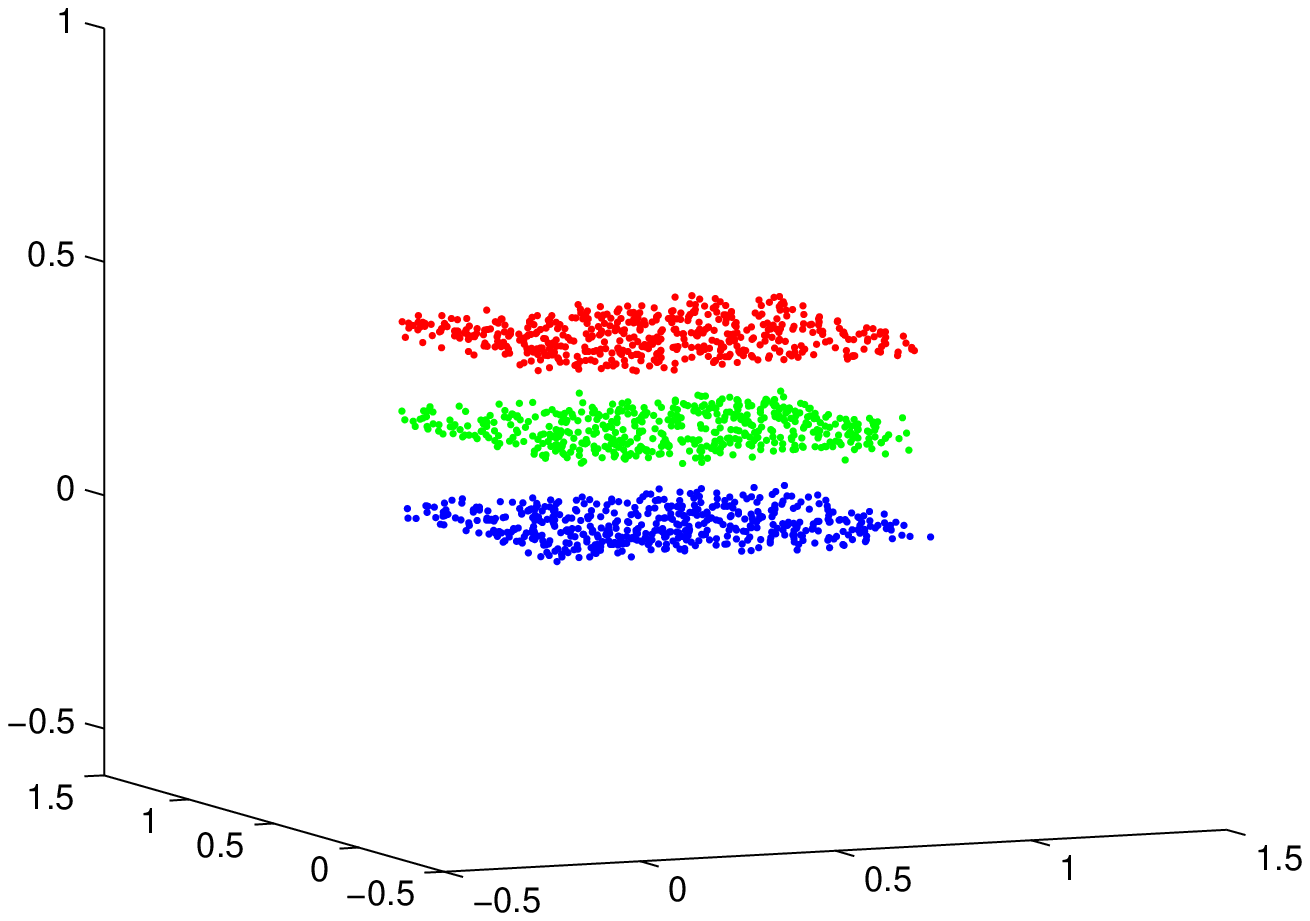}
\end{minipage}
\begin{minipage}{0.31\textwidth} \centering
\includegraphics[width=1\textwidth,height=.8\textwidth]{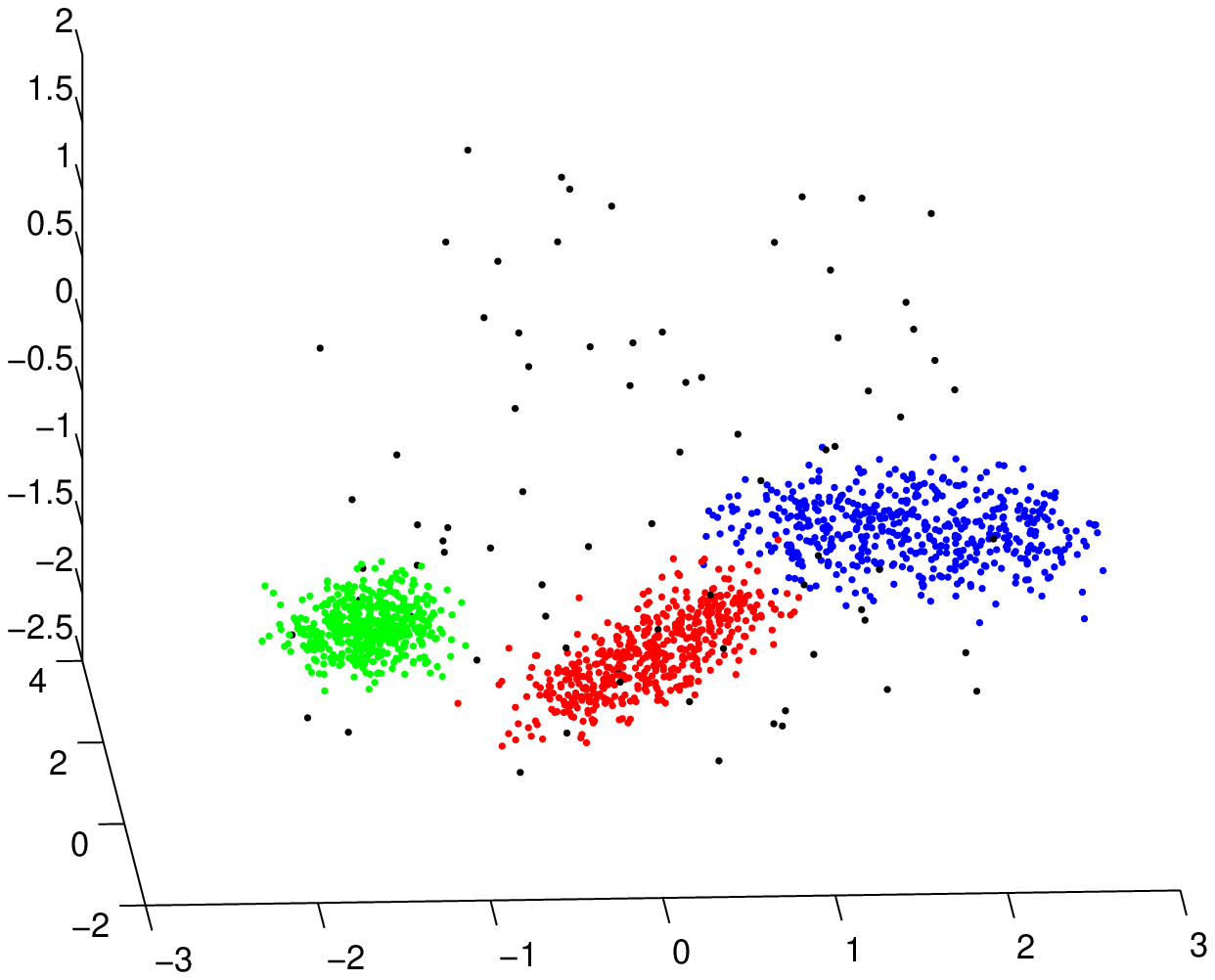}
\end{minipage}
\begin{minipage}{0.31\textwidth} \centering
\includegraphics[width=1\textwidth,height=.8\textwidth]{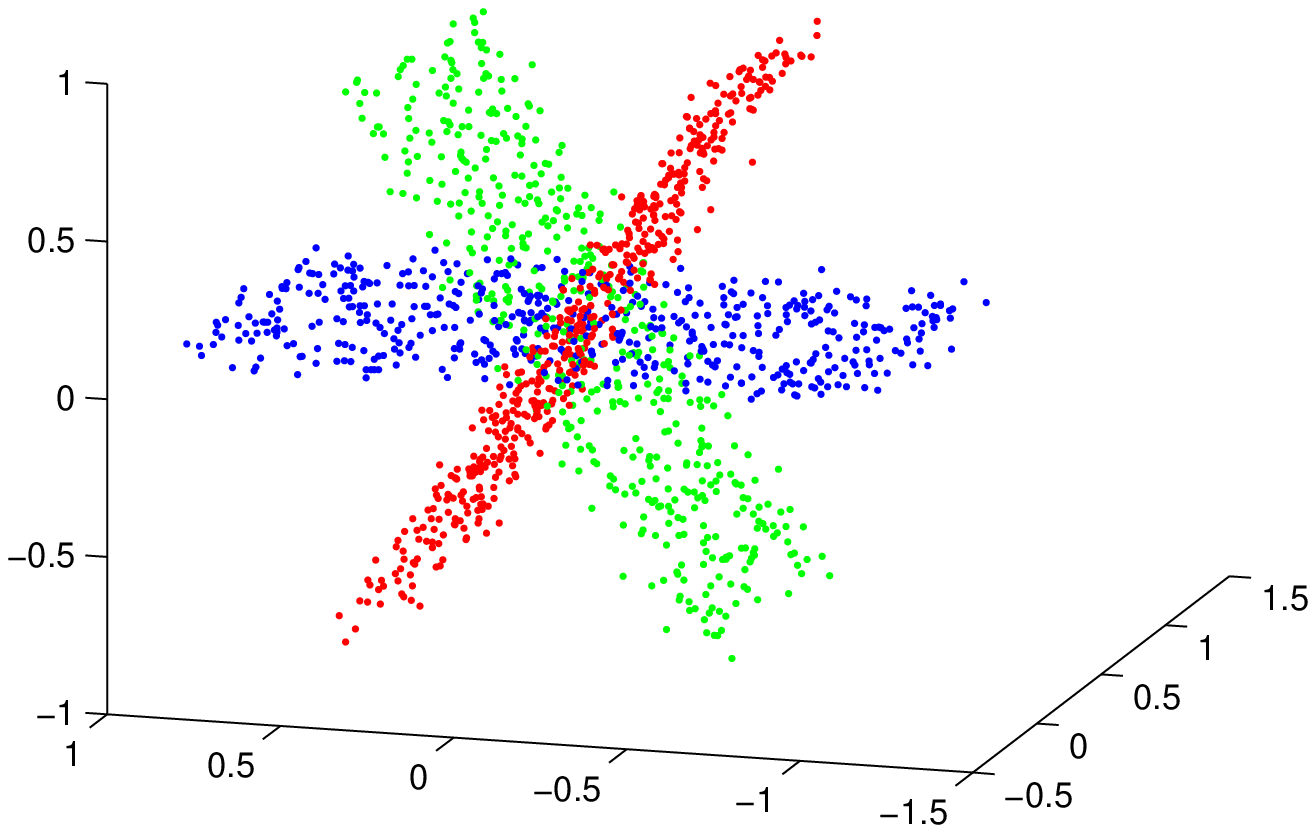}
\end{minipage}
\begin{minipage}{0.31\textwidth} \centering
\includegraphics[width=1\textwidth,height=.8\textwidth]{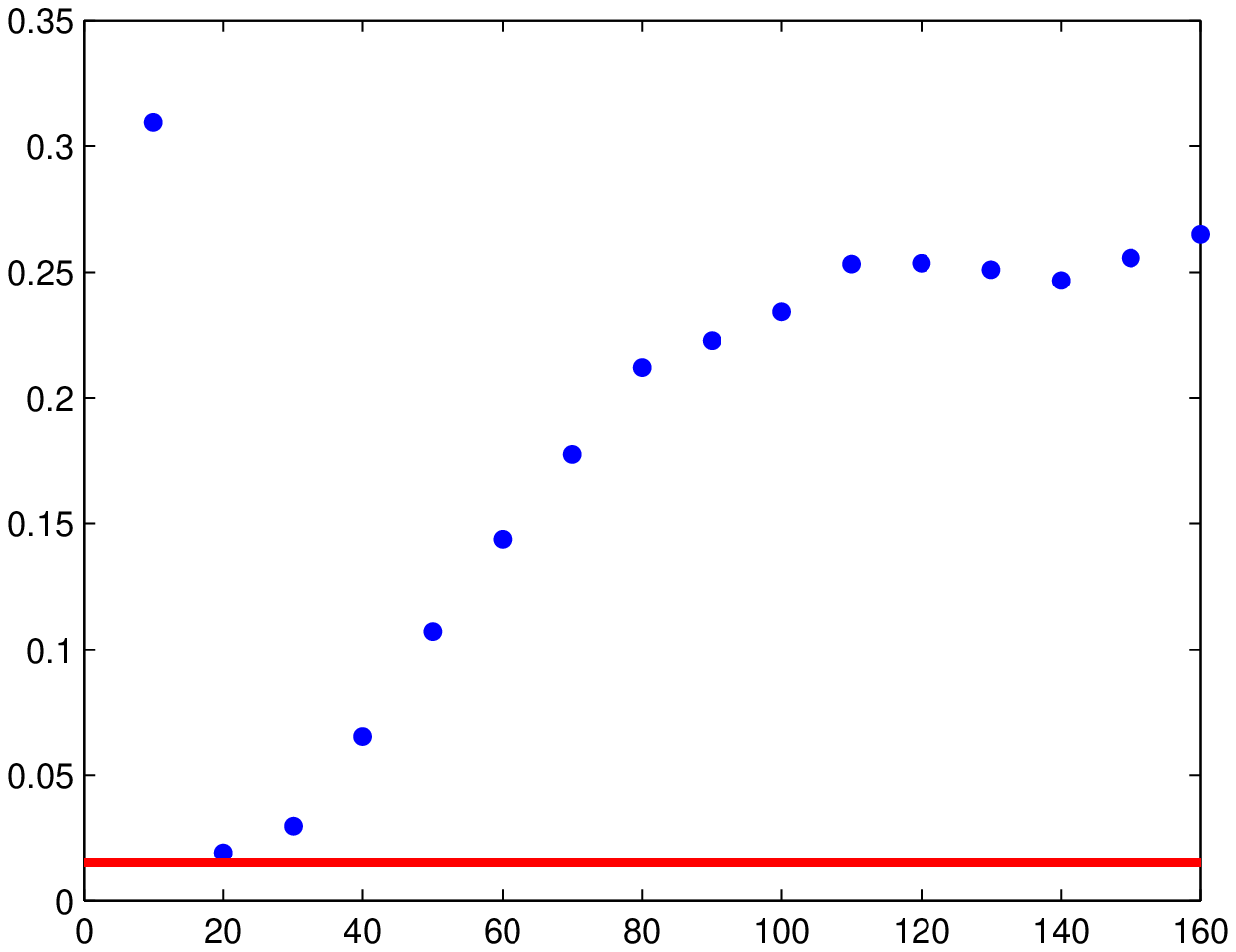}
\end{minipage}
\begin{minipage}{0.31\textwidth} \centering
\includegraphics[width=1\textwidth,height=.8\textwidth]{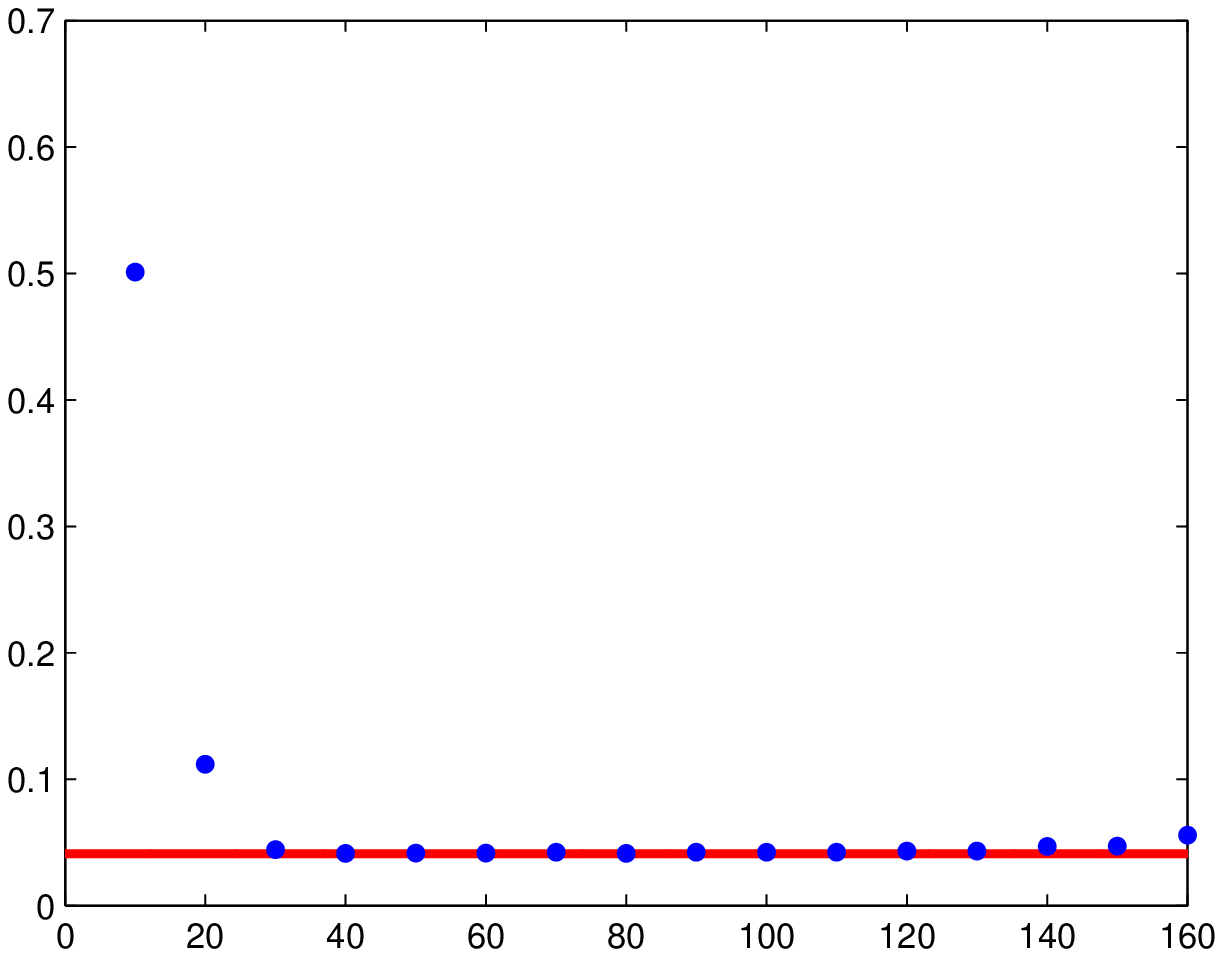}
\end{minipage}
\begin{minipage}{0.31\textwidth} \centering
\includegraphics[width=1\textwidth,height=.8\textwidth]{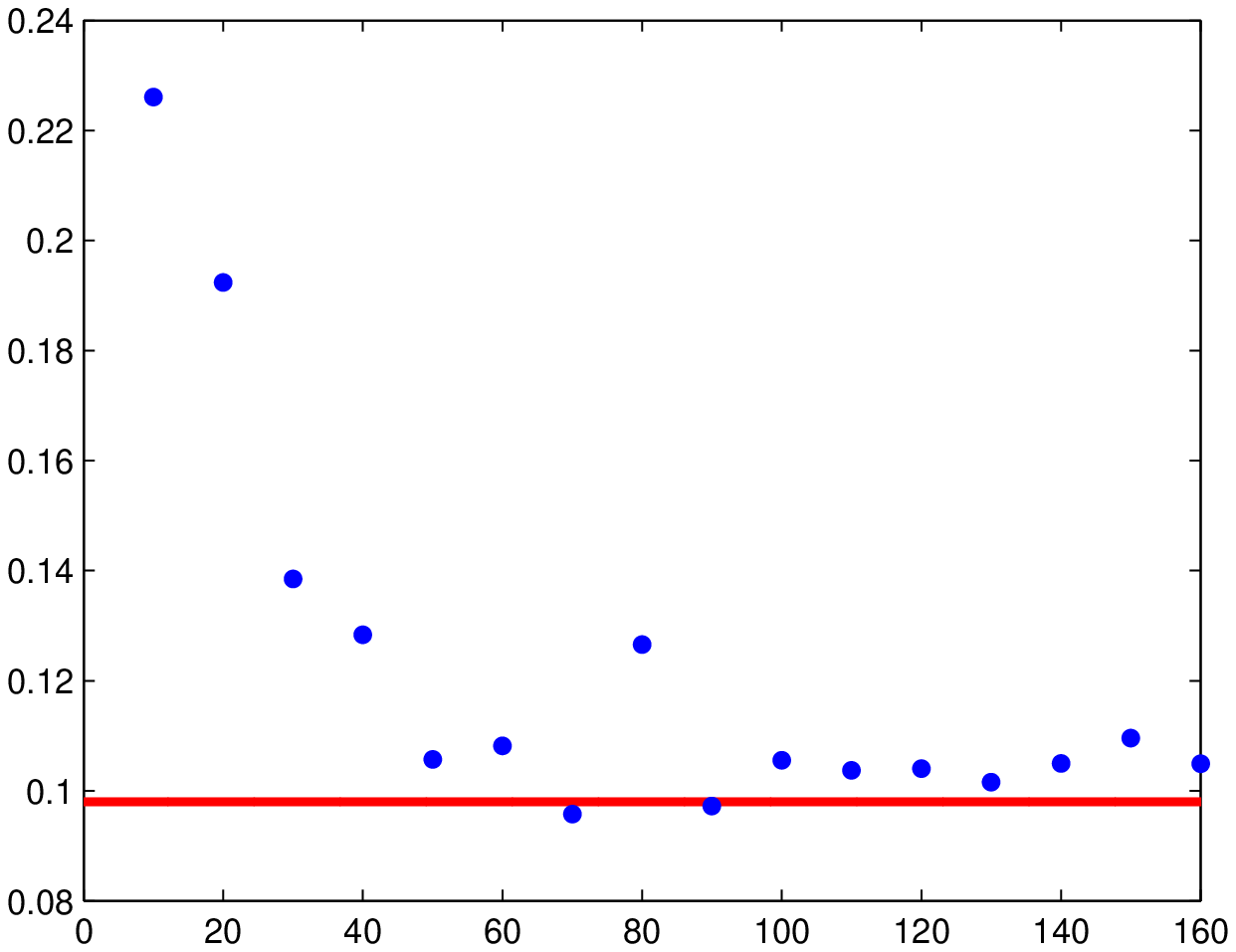}
\end{minipage}
\caption{\it Using our neighborhood choice to improve initialization
of $k$-flats: the first row is the visualization of three data sets, and the seconds row shows the corresponding figures such that the vertical axis is accuracy, and the horizontal axis
is fixed neighborhood size in geometric farthest insertion for
initialization of $K$ flats.  The red line is the result of using
adapted neighborhoods. The data sets are \#1,\#2, and \#3 as
described in Section~\ref{sec:ikf}.  Random initialization leads to
misclassification rates of .4 or greater for all three data sets.
\label{fig:kmeansit} }
\end{center}
\end{figure*}

We work on three data sets.  Data set \#1 consists of $1500$
points on three parallel $2$-planes in $\R^3$. $500$ points are
drawn from the unit square in $x,y$ plane, and then $500$ more from
the $x,y,z+.2$ plane, and then $500$ more from the $x,y,z+.4$ plane.
This data set is designed to favor the use of small neighborhoods.
The next data set is three random flats with 15\% Gaussian
noise and 5\% outliers, generated using the Matlab code from GPCA,
as in Section~\ref{sec:sim}. This data set is designed to favor
large neighborhood choices.  Finally, we work on a data set with
1500 points sampled from 3 planes in $\R^2$ as in
Figure~\ref{fig:nonei}. The error rates of $K$-flats with farthest
insertion initialization with fixed neighborhoods of size $10$,
$20$, $...$, $160$ are plotted against the error rates for farthest
insertion with adapted neighborhoods (searched over the same range),
averaged over 400 runs in Figure~\ref{fig:kmeansit}. Although our
method did not always beat the best fixed neighborhood, it was quite
close; and it always significantly better than the wrong fixed
neighborhood size. Both methods did significantly better than a
random initialization.

In Figure~\ref{fig:nonei} we plot the number of neighbors picked by
our algorithm for each point of a realization of data set \#3.

\section{Conclusions and future work}
\label{sec:conclusions}
 We presented a very simple geometric method for
HLM based on selecting a set of local best-fit
flats.  The size of
the local neighborhoods is determined automatically using the
$\ell_2$ $\beta$ numbers; it is proven under certain geometric
conditions that our method approximately finds the optimal local neighborhoods. We give extensive experimental evidence \\demonstrating the
state of the art accuracy and speed of the algorithm on synthetic
and real hybrid linear data.

We believe that one promising next step is to adapt the method for
multi-manifold clustering.  As it is, our method, while quite good
at unions of flats, cannot successfully handle unions of
curved manifolds.  We expect that by gluing together groups of
local best-fit flats related by some smoothness conditions, we will
be able to approach the problem of clustering data which lies on
unions of smooth manifolds.

We also believe that it will be possible to provide a theoretical framework for performance guarantees with noise for LBF and SLBF.  Specifically, we hope to prove a quantitative form of the following alternative:
suppose the data lies on the union of $d$-dimensional affine sets, perhaps with additive noise and outliers.  Then either
\begin{enumerate}
\item Most points are roughly as close to an affine set they don't belong to as they are to their nearest $O(d)$ neighbors;
\item A large fraction of the points have optimal neighborhoods contained in only one of the affine clusters, the principal components of these neighborhoods are good approximations to the clusters; and LBF and SLBF recover good approximations to the two affine clusters,
or
\item The data looks locally lower than $d$-dimensional, even though each cluster is globally $d$-dimensional, and has high curvature; in this case, there are pure optimal neighborhoods, but the local estimation does not accurately represent the affine clusters.
\end{enumerate}

%{\bf add about faces}
\appendix
\section{Proof of Theorem~\ref{thm:main}}\label{sec:proof_of_thm}
%\begin{proof}

%In order to demonstrate the basic ideas about the proof, we also use an
%$L_\infty$ version of the continuous $\beta_2$, which is formed as follows:
%$$\beta_\infty(\bx,r) = \min_{d-\text{flats } L} \max_{\by \in \Omega \cap B(\bx,r)}
%\frac{\dist(\bx,L)}{r}\,.$$
%
%
%Clearly, the function $\beta_\infty$ is not robust to noise, unlike $\beta_2$.
%However, direct computations are easier with $\beta_\infty$ and there is no need in the various comparability constants,
%which are $1$ in this case.

Assume without loss of generality that $i^* = 1$. Note that when $r\leq r_0$, $B(\bx^*,r) \cap T(L_1,w)=B(\bx^*,r) \cap \supp(\mu)$ and that $L_1$ is the minimizer of the RHS of~\eqref{eq:beta_2}. Combining these observations with~\eqref{eq:beta_2} and the fact that $\beta_2(\bx^*,r)$ is invariant to scaling of $r$ and $w$, we immediately obtain that for $r<r_0$:
\begin{equation}
\beta_2^2(\bx^*,r)=\frac{\int_{T(L_1,\frac{w}{\max(r,w)})\cap B(\bx^*,1)}\dist(\bx^*,L_1)^2\di \mu_1}{\mu_1\left({T(L_1,\frac{w}{\max(r,w)})\cap B(\bx^*,1)}\right)}.
%\int_{T(L_1,\frac{w}{\max(r,w)})\cap B(\bx^*,1)}\frac{\dist(\bx^*,L_1)^2\di x}{\vol(T(L_1,\frac{w}{\max(r,w)})\cap B(\bx^*,1))},
\label{eq:beta_2_scaled}
\end{equation}
In particular, $\beta_2(\bx^*,r)$ is constant for all $0\leq r\leq w$.

To show that $\beta_2(\bx^*,r)$ is strictly decreasing whenever $w \leq r \leq r_0$, we first note that
for any $r_1$ and $r_2$ satisfying $w \leq r_1 \leq r_2 \leq r_0$:
\be
T(L_1,\frac{w}{\max(r_2,w)})\cap B(\bx^*,1)\subset
T(L_1,\frac{w}{\max(r_1,w)})\cap B(\bx^*,1).
\ee
Moreover, any point in $T(L_1,\frac{w}{\max(r_1,w)})\setminus
T(L_1,\frac{w}{\max(r_2,w)})$ has a larger distance to $L_1$ than any point in $T(L_1,\frac{w}{\max(r_2,w)})$. Combining these observations with~\eqref{eq:beta_2_scaled}, we conclude that
$\beta_2(\bx^*,r_1)>\beta_2(\bx^*,r_2)$, i.e., $\beta_2(\bx^*,r)$ is strictly decreasing on $[w , r_0]$.

Next, we will prove~\eqref{eq:beta_local_min} with a weaker requirement on $r^*$.
More precisely, we define $r^*=\max(r^*_1,r^*_2)$, where
\be
r_1^*=\begin{cases}
\frac{r_0+2w}{\sqrt{1-{\frac{3\sqrt{2}\,(D-1)Kw^2}{(D+1)r_0(r_0+2w)}}}},& \text{when $d=1$};\\
\frac{r_0+2w}{\sqrt{1-(\frac{6(D-d)Kw^2}{(D-d+2)(r_0+2w)^2})^\frac{2}{d}}},&\text{when $d>1$}\end{cases}
\ee
and
\be
r_2^*=\frac{1}{\sqrt{\frac{2}{(r_0+2w)^2}-\frac{1}{r_0^2}}}.
\ee
We further assume that $w< r_0$ and
\be\label{eq:condition}
\frac{(r^{*2}-r_0^2)^\frac{d}{2}\cdot(r_0+w)}{(r^{*2}-r_0^2)^\frac{d}{2}+(r^{*2}-w^2)^\frac{d}{2}}+\frac{r^*}{\sqrt{d+1}}\leq r_0.
\ee
We will later show that \eqref{eq:condition0} implies \eqref{eq:condition} and we will also verify that $r_0 \leq r^* < 1.09\,r_0$.

Without loss of generality we assume that $\argmin_{i>1}\\\dist(\bx^*,L_i)=2$, and let $\bx_0$ be the center of mass of $\mu_1+\mu_2$ in $B(0,r)$. Then for any $r>r_0$
\begin{align}
&\min_{L} \!\!\!\!\!\! \int\limits_{B(\bx^*,r)}\!\!\!\!\!\left(\frac{\dist(\bx,L)}{r}\right)^2 \!\!\di\mu
\geq
\min_{L} \!\!\!\!\!\! \int\limits_{B(\bx^*,r)}\!\!\!\!\!\left(\frac{\dist(\bx,L)}{r}\right)^2 \!\!\di(\mu_1+\mu_2)\nonumber\\
&=\min_{L: \bx_0\in L} \!\!\!\!\!\! \int\limits_{B(\bx^*,r)}\!\!\!\!\!\left(\frac{\dist(\bx,L)}{r}\right)^2 \!\!\di(\mu_1+\mu_2)
\nonumber\\&\geq
\min_{L} \!\!\!\!\!\! \int\limits_{B(\bx^*,r)}\!\!\!\!\!\!\left(\frac{\dist(\bx,L)}{r}\right)^2 \di\mu_1
+
\min_{L: \bx_0\in L} \!\!\!\!\!\! \int\limits_{B(\bx^*,r)}\!\!\!\!\!\!\left(\frac{\dist(\bx,L)}{r}\right)^2 \di\mu_2
\nonumber\\&= \int\limits_{B(\bx^*,r)}\!\!\!\!\!\!\left(\frac{\dist(\bx,L_1)}{r}\right)^2 \di\mu_1
+
\min_{L: \bx_0\in L} \!\!\!\!\!\! \int\limits_{B(\bx^*,r)}\!\!\!\!\!\!\left(\frac{\dist(\bx,L)}{r}\right)^2 \di\mu_2.
\label{eq:least_square_line}
\end{align}
We claim that when $r=r^*$, the minimizer in the second expression in the RHS of~\eqref{eq:least_square_line} (denoted by $L_0$) satisfies that $\dim(L_0\cap L_2)=d-1$ and $(L_0\cap L_2^\perp)\perp L_2$. We denote the orthonormal vector passes through $\bx^*$ and $\bx_0$  by $\bu_1$, one of the $d$ orthonormal vectors that span $L_2$ by $\bu_2$, and one of  the $D-d-1$ orthonormal vectors that span $(\Sp(L_2))^\perp$ by $\bu_2$. We will prove that $\bu_1$ is the top eigenvector of $\int_{B(\bx^*,r^*)}(\bx-\bx_0)(\bx-\bx_0)^T\di \mu(\bx)$, and $\bu_2$ is the second top eigenvector, by proving
\begin{align}
&\int_{B(\bx^*,r^*)\cap T(L_2,w)}\!\!\!(\bu_1^T(\bx-\bx_0))^2\di \mu_2(\bx)\nonumber\\
>&\! \int_{B(\bx^*,r^*)\cap T(L_2,w)}\!\!\!(\bu_2^T(\bx-\bx_0))^2\di \mu_2(\bx)\nonumber\\
> &\int_{B(\bx^*,r^*)\cap T(L_2,w)}(\bu_3^T(\bx-\bx_0))^2\di \mu_2(\bx).\label{eq:principal_components}
\end{align}

We note that
\begin{align}
&\left(B(\bx^*,w)\cap L_1^\perp\right)\times\left(B(\bx^*,\sqrt{r^{*2}-w^2})\cap L_1\right)
\subset T(L_1,w)\cap\nonumber\\& B(\bx^*,r^*)
\subset
\left(B(\bx^*,w)\cap L_1^\perp\right)\times\left(B(\bx^*,r^*)\cap L_1\right).\label{eq:volume_L_1}
\end{align}
Defining $\by$ as nearest point to $\bx^*$ on $L_2$, then for $r^*>r_0+2w$,
we have that
\begin{align}
&\!\left(B(\by,w)\cap L_2^\perp\!\right)\!\times\!\left(\!B(\by,\sqrt{r^{*2}-(r_0+2w)^2})\cap L_2\!\right)\!
\subset T(L_2,w)\cap\nonumber\\& B(\bx^*,r^*)
\subset
\left(B(\by,w)\cap L_2^\perp\right)\!\times\!\left(B(\by,\sqrt{r^{*2}-r_0^2})\cap L_2\right).\label{eq:volume_L_2}
\end{align}

 Moreover
 \begin{align}\text{$\vol\left(B(\bx^*,r_1)\cap L^\perp\right)\!\times\!\left(B(\bx^*,r_2)\cap L\right)=C_0(d,D-d)\,r_1^{D-d}r_2^d$.}\label{eq:volume_general}\end{align} Denote the center of mass of $B(\bx^*,r^*)\cap T(L_2,w)$ by $\bx_1$, notice that  $\|\bx_0-\bx^*\|<r_0+w$,  the center of mass of $B(\bx^*,r^*)\cap T(L_1,w)$ is $\bx^*$, and $x^*$, $\bx_0$ and $\bx_1$ satisfies
 \be\label{eq:center_of_mass}
 \bx_0\!=\!\frac{\vol({B(\bx^*,r^*\!)\cap T(L_1,w)})\,\bx^*\!\!+\!\vol(B(\bx^*,r^*\!)\cap T(L_2,w))\,\bx_1}{\vol({B(\bx^*,r^*)\cap T(L_1,w)})+\vol(B(\bx^*,r^*)\cap T(L_2,w))}.
 \ee
Combining \eqref{eq:volume_L_1}, \eqref{eq:volume_L_2}, \eqref{eq:volume_general} and \eqref{eq:center_of_mass} we have the estimation
\begin{align}
&\!\|\bx_0\!-\!\bx^*\!\|\!\!\leq \!\!\frac{\vol(B(\bx^*,r^*)\cap T(L_2,w))\,(r_0+w)}
{\vol({B(\bx^*,r^*\!)\cap T(L_1,w)})\!\!+\!\!\vol({B(\bx^*,r^*\!)\cap T(L_2,w)})}\nonumber\\\leq & \frac{(r^{*2}-r_0^2)^\frac{d}{2}}{(r^{*2}-w^2)^\frac{d}{2}+(r^{*2}-r_0^2)^\frac{d}{2}}\cdot(r_0+w).
\label{eq:norm_x0}\end{align}

Therefore for any point $\bx_1$ in $B(\bx^*,r^*)\cap T(L_2,w)$, using \eqref{eq:condition} and \eqref{eq:norm_x0} we have
\[\left|\bu_1^T(\bx_1-\bx_0)\right|\geq r_0-\|\bx_0-\bx^*\|\geq \frac{r^*}{\sqrt{d+1}}
\]
and
\be
\int_{B(\bx^*,r^*)\cap T(L_2,w)}\!\!\!\!\!\!\!\!\!\!\!\!\!\!\!\!\!\!\!\!(\bu_1^T(\bx-\bx_0))^2\di \mu_2(\bx)
\geq \frac{r^{*2}}{{d+1}}
\mu_2(B(\bx^*,r^*)\cap T(L_2,w)).\label{eq:bu_1}
\ee
Since any points in $B(\bx^*,r^*)\cap T(L_2,w)$ has a distance to $\bx_0$ smaller than $r^*$, we have
\begin{align}
&\int_{B(\bx^*,r^*)\cap T(L_2,w)}\!\!\!\!\!\!\!\!\!\!(\bu_1^T(\bx-\bx_0))^2+d\int_{B(\bx^*,r^*)\cap T(L_2,w)}\!\!\!\!\!\!\!\!\!\!(\bu_2^T(\bx-\bx_0))^2\nonumber\\
&\int_{B(\bx^*,r^*)\cap T(L_2,w)}\!\!\!\!\!\!\!\!\!\!\|\bx-\bx_0\|^2\di \mu_2(\bx)<r^{*2}\mu_2(B(\bx^*,r^*)\cap T(L_2,w)).\label{eq:bu_2}
\end{align}
Combining \eqref{eq:bu_1} and \eqref{eq:bu_2}, the first inequality in \eqref{eq:principal_components} is proved.

By direct integration one obtains that the average of $(\bu_2^T\bx_1)^2$ for $\bx_1$ in \[
\left(B(\by,w)\cap L_2^\perp\right)\times\left(B(\by,\sqrt{r^{*2}-(r_0+2w)^2})\right),
\]
is $\frac{d}{d+2}(r^{*2}-(r_0+2w)^2)$, and the average of $(\bu_2^T\bx_1)^2$ for $\bx_1$ in \[T(L_2,w)\setminus \left( \left(B(\by,w)\cap L_2^\perp\right)\times\left(B(\by,\sqrt{r^{*2}-(r_0+2w)^2})\right)\right)\]
is larger than that of the set
\[
\left(B(\by,w)\cap L_2^\perp\right)\times\left(B(\by,\sqrt{r^{*2}-(r_0+2w)^2})\right).
\]
Applying these two facts, we obtain the estimate
\begin{align}
&\int_{B(\bx^*,r^*)\cap T(L_2,w)}(\bu_2^T(\bx^*-\bx_0))^2\di \mu_2
\nonumber\\\geq & \frac{d}{d+2}(r^{*2}-(r_0+2w)^2) \mu_2(B(\bx^*,r^*)\cap T(L_2,w)).\label{eq:principal_components1}
\end{align}

We also have
\begin{align}\label{eq:principal_components2}
\int_{B(\bx^*,r^*)\cap T(L_2,w)}\!\!\!\!\!\!\!\!\!\!\!\!\!\!\!\!\!\!(\bu_3^T(\bx^*-\bx_0))^2\di \mu_2
\leq w^2 \mu_2(B(\bx^*,r^*)\cap T(L_2,w)).
\end{align}

Using the fact that $r^*\geq r^*_2$, we have
\begin{align}\nonumber
&r^{*2}-(r_0+2w)^2\geq r^{*2}_2-(r_0+2w)^2= (r_0+2w)^2\cdot \\
&\left(\frac{r_0^2}{r_0-4r_0w-4w^2}-1\right)=\frac{(r_0+2w)^2}{r_0-4r_0w-4w^2}\cdot (r_0w+4w^2)
\nonumber\\&> 4w^2.\label{eq:principal_components3}
\end{align}

Combining \eqref{eq:principal_components1}, \eqref{eq:principal_components2} and \eqref{eq:principal_components3}, the second inequality in \eqref{eq:principal_components} is also proved.

To estimate $\beta_2(\bx^*,r^*)$ and $\beta_2(\bx^*,r_0)$, using integration the points in $\left(B(\bx^*,r_1)\cap L^\perp\right)\times\left(B(\bx^*,r_2)\cap L\right)$ has an average squared distance $\frac{D-d}{D-d+2}r_2^2$ to $L$. Besides, the points in $\left(B(\by,w)\cap L_2^\perp\right)\times\\\left(B(\by,\sqrt{r^{*2}-(r_0+2w)^2})\cap L_2\right)$ has an average squared distance at least $(r^{*2}-(r_0+2w)^2)/3$ to the minimizer $L$  in~\eqref{eq:least_square_line}.
  Combining these facts with~\eqref{eq:least_square_line}, ~\eqref{eq:volume_L_1}, ~\eqref{eq:volume_L_2}, and \eqref{eq:volume_general},  we have \begin{align}\label{eq:beta_2_r}
  \beta_2^2(x^*,r^*\!)\!>\!\frac{\frac{D-d}{D-d+2}w^{D-d+2}r^{*d}\!+\!
   w^{D-d}(r^{*2}\!-(r_0+2w)^2)^{\frac{d+2}{2}}\!/3
  }{r^{*2}\left(w^{D-d}r^{*d}+(K-1)w^{D-d}(r^{*2}-r_0^2)^{\frac{d}{2}}\right)}
  \end{align}
  and
    \begin{align}\label{eq:beta_2_r0}
    \beta_2^2(x^*,r_0)<\frac{D-d}{D-d+2}\frac{w^2}{r_0^2}.
    \end{align}

To prove~\eqref{eq:beta_local_min}, we only need to prove that the RHS of \eqref{eq:beta_2_r} is larger than the RHS of \eqref{eq:beta_2_r0}, which has a following simplified form:
\begin{align}
&\frac{D-d+2}{3(D-d)}\left(1-\frac{(r_0+2w)^2}{r^{*2}}\right)^\frac{d+2}{2}
\nonumber\\\geq& (K-1)\frac{w^2}{r_0^2}(1-\frac{r_0^2}{r^{*2}})^\frac{d}{2}+\frac{w^2}{r_0^2}(1-\frac{r_0^2}{r^{*2}}).
\label{eq:case0}\end{align}

When $d=1$, ~\eqref{eq:case0} follows from
%  \begin{align}
%\frac{D-d+2}{3(D-d)}\left(1-\frac{(r_0+2w)^2}{r^{*2}}\right)^\frac{d+2}{2}
%\geq K\frac{w^2}{r_0^2}(1-\frac{r_0^2}{r^{*2}})^\frac{1}{2},
%\end{align}
%that is
\begin{align}\label{eq:case1}
\left(\frac{D+1}{3(D-1)}\right)^2\left(1-\frac{(r_0+2w)^2}{r^{*2}}\right)^3
\geq K^2 \frac{w^4}{r_0^4}(1-\frac{r_0^2}{r^{*2}}).
\end{align}
From $r^*\geq r^*_1$, we have
\begin{align}\label{eq:case1_r1}
\left(1-\frac{(r_0+2w)^2}{r^{*2}}\right)^2\geq
\frac{18(D-d)^2K^2}{(D-d+2)^2}\frac{w^4}{r_0^2(r_0+2w)^2},
\end{align}
and from $r^*\geq r^*_2$ we have
\begin{align}\label{eq:case1_r2}
2(\frac{1}{(r_0+2w)^2}-\frac{1}{r^{*2}})\geq\frac{1}{r_0^2}-\frac{1}{r^{*2}}.
\end{align}

Then \eqref{eq:case1} follows from \eqref{eq:case1_r1} and \eqref{eq:case1_r2}, and therefore \eqref{eq:beta_local_min} is proved for the case $d=1$.

For the case $d\geq 2$, the proof of~\eqref{eq:case0} follows a similar strategy. Combing \eqref{eq:case1_r2} and
\begin{align}\label{eq:case1_r1_case2}
\left(1-\frac{(r_0+2w)^2}{r^{*2}}\right)^\frac{d}{2}\geq
\frac{6(D-d)K}{(D-d+2)}\frac{w^2}{(r_0+2w)^2},
\end{align}
we obtain
  \begin{align}\label{eq:case2}
\frac{D-d+2}{3(D-d)}\left(1-\frac{(r_0+2w)^2}{r^{*2}}\right)^\frac{d+2}{2}
\geq K\frac{w^2}{r_0^2}(1-\frac{r_0^2}{r^{*2}}),
\end{align}
and~\eqref{eq:case0} follows from~\eqref{eq:case2}.

Now we will prove that \eqref{eq:condition0} satisfies \eqref{eq:condition}. Notice that $r^{*2}-w^2>(r_0+2w)^2-w^2>r_0^2$,  it is sufficient to prove
\begin{align*}
\frac{r_0+w}{1+(\frac{r_0^2}{r^{*2}-r_0^2})^\frac{d}{2}}+\frac{r^*}{\sqrt{d+1}}\leq r_0.
\end{align*}
and since $r^*>r_1^*>(r_0+2w)/\sqrt{1-c}>(1+2c)r_0/\sqrt{1-c}$ and $w<c\,r_0$, where $c=0.02$, we only need to prove
\begin{align}\label{eq:condition1}
\frac{1+c}{1+\left(\frac{1}{\frac{(1+2c)^2}{1-c}-1}\right)^{\frac{d}{2}}}+\frac{1+2c}
{\sqrt{(d+1)(1-c)}}\leq 1.
\end{align}
It holds for $c=0.02$ and $d=1$. Since that when $c$ is fixed, $d=1$ maximizes the LHS of \eqref{eq:condition1}, \eqref{eq:condition1} holds for any $d$ with $c=0.02$. Therefore \eqref{eq:condition0} satisfies \eqref{eq:condition} and Theorem~\ref{thm:main} is proved.

At last we will show that $r^*<1.09\,r_0$. Indeed, $r^*_1<r_0(1+2\cdot 0.02)/\sqrt{1-0.02}<1.09\,r_0$, and $r^*_2<\frac{1}{\sqrt{\frac{2}{1.04^2}-1}}r_0<1.09 r_0$, therefore $r^*=\max(r^*_1,r^*_2)<1.09r_0$.
\begin{remark}
% missing example:
%, where $ note that 1.$r^*$ is not much bigger than $r_0$ when $\frac{w}{r_0}$ is small. In particular, $r^*<1.06\,r_0$ when \eqref{eq:condition0} holds. 2.
The function $\beta_2(x,r)$ often does not have a  local minimum  at exactly $r_0$. We demonstrate it for a particular case, but it is evident that this is rather typical. Assume that $K=2$, $d=1$, $D=2$ and $L_1\perp L_2$, then for sufficiently small $\eta$, $\{B(\bx^*,r)\cap T(L_2,w_2)\}\subset T(L_1,\beta_2(\bx^*,r_0))$. Following the same argument for the interval $[w_{i^*}, r_0]$, $\beta_2(\bx^*,r)$ is decreasing in the interval $[r_0,r_0+\eta]$.
\end{remark}

\bibliographystyle{spmpsci}
\bibliography{refs_2_19_12}

%\end{small}
\end{document}